\documentclass{article}

\usepackage{microtype}
\usepackage{graphicx}
\usepackage{subfigure}
\usepackage{booktabs} %

\usepackage{hyperref}

\usepackage[accepted]{icml2025}

\usepackage[utf8]{inputenc} %
\usepackage[T1]{fontenc}    %
\usepackage{hyperref}       %
\usepackage{url}            %
\usepackage{booktabs}       %
\usepackage{amsfonts}       %
\usepackage{nicefrac}       %
\usepackage{microtype}      %
\usepackage{xcolor}         %

\usepackage{orcidlink}
\usepackage{multirow}
\usepackage{adjustbox}
\usepackage{amsmath}
\usepackage{amssymb}
\usepackage{mathtools}
\usepackage{bm}
\usepackage{tabu}
\usepackage{amsthm}
\usepackage{xcolor}
\usepackage{colortbl}
\usepackage{multirow}
\usepackage{wrapfig}
\usepackage{subcaption}
\usepackage{lipsum}
\usepackage{soul}
\usepackage{amsmath}
\usepackage{setspace}
\usepackage{tabu}
\usepackage{makecell}
\usepackage{pifont}
\usepackage{wrapfig}
\usepackage{cleveref}
\usepackage[textsize=tiny]{todonotes}

\RequirePackage{xspace}

\makeatletter
\DeclareRobustCommand\onedot{\futurelet\@let@token\@onedot}
\def\@onedot{\ifx\@let@token.\else.\null\fi\xspace}

\def\eg{\emph{e.g}\onedot} 
\def\ie{\emph{i.e}\onedot}

\def\wrt{w.r.t\onedot} 
 
\def\etal{\emph{et al}\onedot}
\makeatother

\newcommand{\bfeps}{\bm{\epsilon}}
\newcommand{\bfx}{\mathbf{x}}
\newcommand{\bfv}{\mathbf{v}}

\newcommand{\dif}{\mathrm d}

\definecolor{myred}{HTML}{df5b3f}  %
\definecolor{mygreen}{HTML}{4caf50} %
\definecolor{myblue}{HTML}{354e97} %

\newcommand{\percent}[1]{{\color{blue} (#1\%)}}

\newcommand{\blue}[1]{{\color{blue} (#1)}}

\newcommand{\PreserveBackslash}[1]{\let\temp=\\#1\let\\=\temp}
\newcolumntype{C}[1]{>{\PreserveBackslash\centering}p{#1}}
\newcolumntype{L}[1]{>{\PreserveBackslash\raggedright}p{#1}}

\definecolor{Gray}{gray}{0.9}
\definecolor{Gray1}{gray}{0.5}

\theoremstyle{plain}

\theoremstyle{definition}

\theoremstyle{remark}

\icmltitlerunning{DiffMoE: Dynamic Token Selection for Scalable Diffusion Transformers}

\begin{document}

\twocolumn[
\icmltitle{DiffMoE: Dynamic Token Selection for Scalable Diffusion Transformers}

\icmlsetsymbol{equal}{*}
\icmlsetsymbol{corr}{$\dagger$}

\begin{icmlauthorlist}
\icmlauthor{Minglei Shi}{tsinghua,equal}
\icmlauthor{Ziyang Yuan}{tsinghua,equal}
\icmlauthor{Haotian Yang}{kuaishou}
\icmlauthor{Xintao Wang}{kuaishou,corr}
\icmlauthor{Mingwu Zheng}{kuaishou}
\icmlauthor{Xin Tao}{kuaishou}
\icmlauthor{Wenliang Zhao}{tsinghua}
\icmlauthor{Wenzhao Zheng}{tsinghua}
\icmlauthor{Jie Zhou}{tsinghua}
\icmlauthor{Jiwen Lu}{tsinghua,corr}
\icmlauthor{Pengfei Wan}{kuaishou}
\icmlauthor{Di Zhang}{kuaishou}
\icmlauthor{Kun Gai}{kuaishou}
\end{icmlauthorlist}

\icmlaffiliation{tsinghua}{Tsinghua University, Beijing, China}
\icmlaffiliation{kuaishou}{Kuaishou Technology, Beijing, China}

\icmlcorrespondingauthor{Xintao Wang}{xintao.wang@kuaishou.com}
\icmlcorrespondingauthor{Jiwen Lu}{lujiwen@tsinghua.edu.cn}

\begin{center}
     \textbf{Project Page:} \url{https://shiml20.github.io/DiffMoE/}
\end{center}
]

\printAffiliationsAndNotice{\icmlEqualContribution} %

\begin{abstract}

Diffusion Transformers (DiTs) have emerged as the dominant architecture for visual generation tasks, yet their uniform processing of inputs across varying conditions and noise levels fails to leverage the inherent heterogeneity of the diffusion process. While recent mixture-of-experts (MoE) approaches attempt to address this limitation, they struggle to achieve significant improvements due to their restricted token accessibility and fixed computational patterns. We present \textbf{DiffMoE}, a novel MoE-based architecture that enables experts to access global token distributions through a \textbf{batch-level global token pool} during training, promoting specialized expert behavior. To unleash the full potential of inherent heterogeneity, DiffMoE incorporates a \textbf{capacity predictor} that dynamically allocates computational resources based on noise levels and sample complexity. Through comprehensive evaluation, DiffMoE achieves state-of-the-art performance among diffusion models on ImageNet benchmark, substantially outperforming both dense architectures with ${3\times}$ activated parameters and existing MoE approaches while maintaining ${1\times}$ activated parameters. The effectiveness of our approach extends beyond class-conditional generation to more challenging tasks such as text-to-image generation, demonstrating its broad applicability across different diffusion model applications.

\end{abstract}

\begin{figure}[t]
\begin{center}
\centerline{\includegraphics[width=\columnwidth]{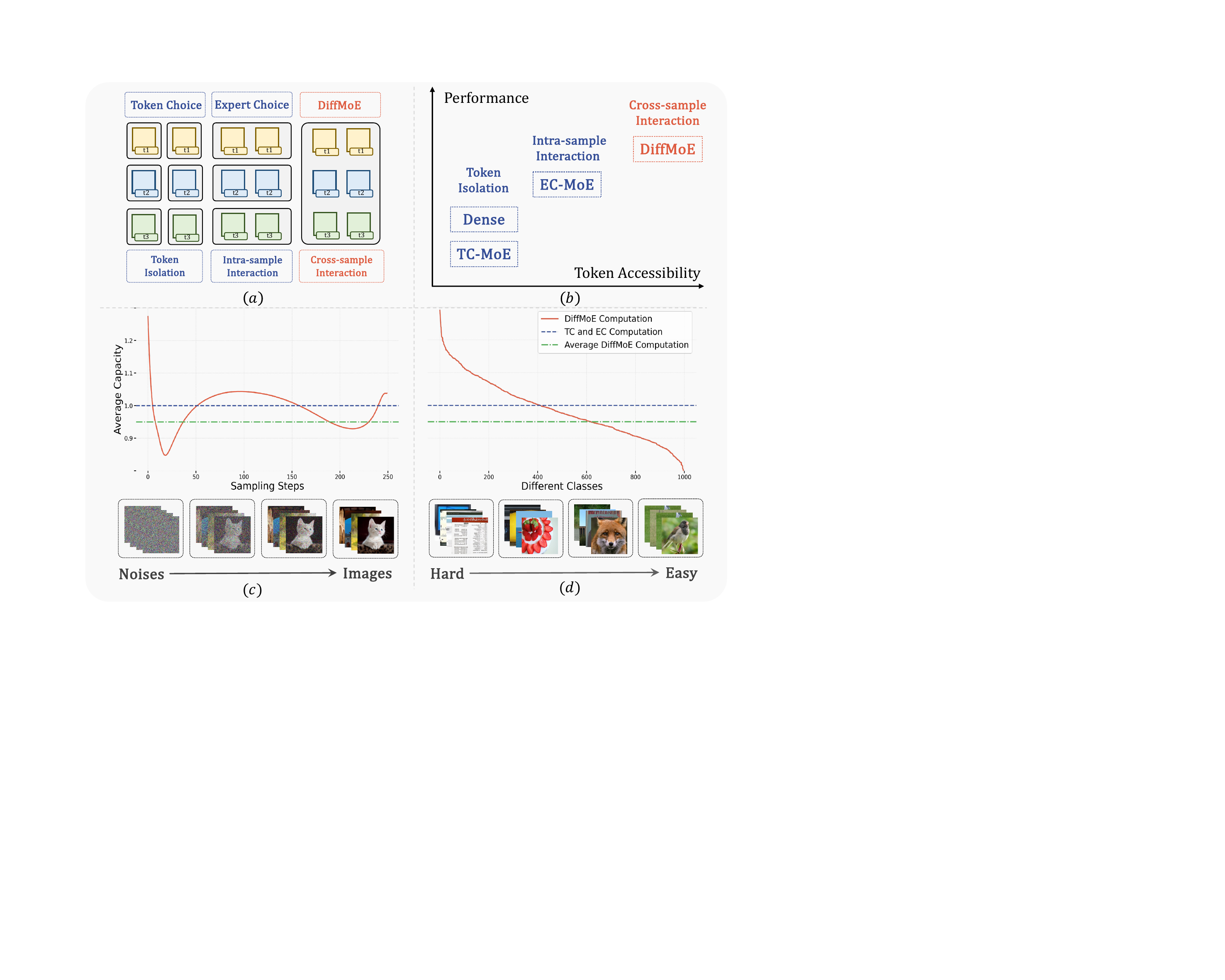}}

\caption{\textbf{Token Accessibility and Dynamic Computation.} \textbf{(a)} Token accessibility levels from token isolation to cross-sample interaction. Colors represent tokens in different samples, $t_i$ indicates noise levels. \textbf{(b)} Performance-accessibility analysis across architectures. \textbf{(c)} Computational dynamics during diffusion sampling, showing adaptive computation from noise to image. \textbf{(d)} Class-wise computation allocation from hard (technical diagrams) to easy (natural photos) tasks. Results from DiffMoE-L-E16-Flow (700K).}

\label{fig:teaser}
\end{center}
\vskip -0.4in
\end{figure}

\section{Introduction}
\label{submission}
The Mixture-of-Experts (MoE) framework \cite{shazeer2017outrageouslylargeneuralnetworks, lepikhin2020gshardscalinggiantmodels} has emerged as a powerful paradigm for enhancing overall multi-task performance while maintaining computational efficiency. This is achieved by combining multiple expert networks, each focusing on a distinct task, with their outputs integrated through a gating mechanism. In language modeling, MoE has achieved performance comparable to dense models of ${2\times}-{3\times}$ activated parameters ~\cite{deepseekai2024deepseekv3technicalreport,minimax2025minimax01scalingfoundationmodels, muennighoff2024olmoeopenmixtureofexpertslanguage}. The current MoE primarily follows two gating paradigms: Token-Choice (TC), where each token independently selects a subset of experts for processing; and Expert-Choice (EC), where each expert selects a subset of tokens from the sequence for processing.

Diffusion ~\cite{ho2020denoising, rombach2022high, podell2023sdxl, song2021score} and flow-based ~\cite{ma2024sit,esser2024scalingrectifiedflowtransformers,liu2023instaflow} models inherently represent multi-task learning frameworks, as they process varying token distributions across different noise levels and conditional inputs.
While this heterogeneity characteristic naturally aligns with the MoE framework's ability for multi-task handling, existing attempts~\cite{FeiDiTMoE2024,sun2024ecditscalingdiffusiontransformers,segmoe,sehwag2024stretchingdollardiffusiontraining} to integrate MoE with diffusion models have yielded suboptimal results, failing to achieve the remarkable improvements observed in language models. 
Specifically, Token-choice MoE (TC-MoE) ~\cite{FeiDiTMoE2024} often underperforms compared to conventional dense architectures under the same number of activations; Expert-choice MoE (EC-MoE) ~\cite{sehwag2024stretchingdollardiffusiontraining,sun2024ecditscalingdiffusiontransformers} shows marginal improvements over dense models, but only when trained for much longer.

We are curious about what fundamentally limits MoE's effectiveness in diffusion models. Our key finding reveals that \textbf{global token distribution accessibility is crucial for MoE success in diffusion models, necessitating the model learn and dynamically process the tokens from different noise levels and conditions},  as illustrated in Figure~\ref{fig:teaser}. Previous approaches have neglected this crucial component, resulting in compromised performance. Specifically, Dense models and TC-MoE \textbf{isolates tokens}, preventing them from interacting with others during expert selection, while EC-DiT restricts \textbf{intra-sample token interaction }, which fails to access other samples with different noise levels and conditions. These limitations hinder the model’s ability to capture the full spectrum of the heterogeneity inherent in diffusion processes.

\begin{figure*}[ht]
\vskip -0.02in
\begin{center}
\centerline{\includegraphics[width=0.95\linewidth]{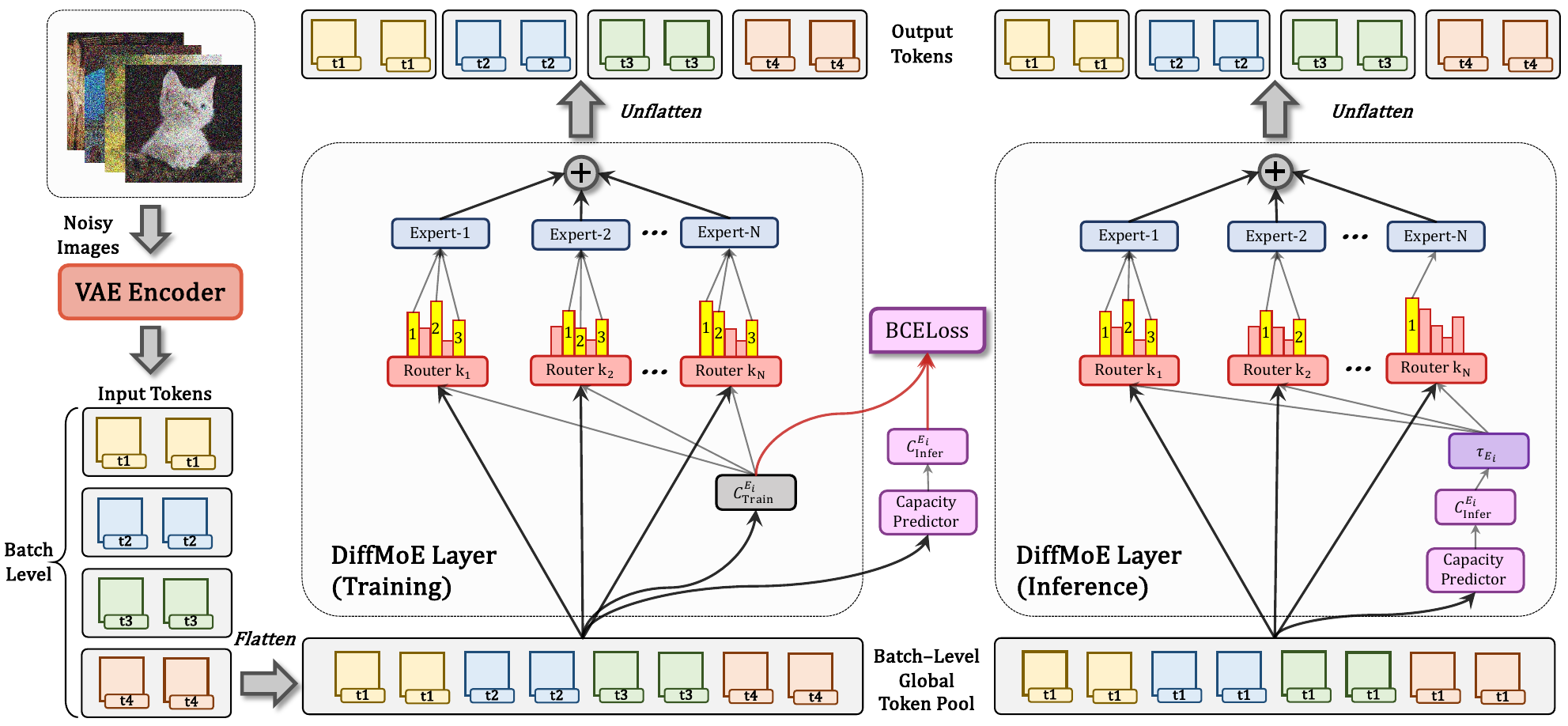}}
\caption{\textbf{DiffMoE Architecture Overview.} DiffMoE flattens tokens into a batch-level global token pool, where each expert maintains a fixed training capacity of $C_{\text{train}}^{E_i}=1$. During inference, a dynamic capacity predictor adaptively routes tokens across different sampling steps and conditions. Different colors denote tokens from distinct samples, while $t_i$ represents corresponding noise levels.}

\label{fig:method}
\end{center}
\vskip -0.28in
\end{figure*}

To address these limitations, we introduce \textbf{DiffMoE}, a novel architecture that features a \textbf{batch-level global token pool} for enhanced \textbf{cross-sample token interaction} during training, as illustrated in Figure~\ref{fig:method}. This approach approximates the complete token distribution across different noise levels and samples, facilitating more specialized expert learning through comprehensive global token information access. Our empirical analysis demonstrates that the global token pool accelerates loss convergence, surpassing dense models with equivalent activation parameters. Though some concurrent works in large language models~\cite{deepseekai2024deepseekv3technicalreport,qiu2025demons} show similar principles in global batch, we argue that these principles are particularly crucial for diffusion transformers due to their inherently more complex heterogeneity nature.

However, conventional MoE inference strategies, which maintain fixed computational resource allocation across different noise levels and conditions, fail to fully leverage the potential of DiffMoE's batch-level global token pool. To optimize token selection during inference, we propose a \textbf{capacity predictor} that dynamically adjusts resource allocation. This adaptive mechanism learns from training-time token routing patterns, efficiently distributing computational resources between complex and simple cases. Furthermore, we implement a \textbf{dynamic threshold} at inference time to achieve flexible performance-computation trade-offs.

By integrating the global token pool and capacity predictor, \textbf{DiffMoE achieves superior performance over dense models with ${3\times}$ activated parameters} while maintaining efficient scaling properties (See Table~\ref{tab:sota_supp}). Our approach offers extra several advantages over existing methods: it eliminates the potentially detrimental load balancing losses present in TC-MoE and overcomes the intra-sample token selection constraints of EC-MoE, resulting in enhanced flexibility and scalability. Extensive empirical evaluations demonstrate DiffMoE's superior scaling efficiency and performance improvements across diverse diffusion applications.

Our contributions can be summarized as follows: (1) We identify the fundamental importance of global token distribution accessibility in facilitating dynamic token selection for MoE-based diffusion models; (2) We introduce DiffMoE, a novel framework incorporating a global token pool and capacity predictor to enable efficient model scaling; (3) We achieve state-of-the-art performance on ImageNet benchmark among diffusion models. through dynamic computation allocation while preserving computational efficiency; and (4) We conduct comprehensive experiments that validate our approach's effectiveness across diverse diffusion applications.

\section{Method}
\subsection{Preliminaries} 
\textbf{Diffusion Models.} Diffusion models~\cite{ho2020denoising, rombach2022high, sohl2015deep, song2021score} are a powerful family of generative models, which can transform the noise distribution $p_1(\bfx)$ to the data distribution $p_0(\bfx)$. The diffusion process can be represented as: $\bfx_t = \alpha_t \bfx_0 + \sigma_t \bfeps, \quad t\in [0, 1],\quad \bfeps\sim \mathcal{N}(0, \mathbf{I}),\label{equ:diffusion_forward}$ Where \(\alpha_t\) and \(\sigma_t\) are monotonically decreasing and increasing functions of \(t\), respectively. The marginal distribution \(p_1(\bfx)\) converges to \(\mathcal{N}(0, \mathbf{I})\), when $\alpha_1=\sigma_0=0, \alpha_0=\sigma_1=1$.

To train a diffusion model, we can use the denoising score matching method ~\cite{song2021score} which constructs a score prediction model \(\bfeps_\theta(\bfx_t, t)\) to estimate the scaled score function \( -\sigma_t \nabla_\bfx \log p_t(\bfx_t)\) with training objective formulated in Eq. ~\ref{loss:eps}. Sampling from a diffusion model can be achieved by solving the reverse-time SDE or the corresponding diffusion ODE~\cite{song2021score} in an iterative manner. Recently, flow-based models~\cite{liu2022flow, lipman2022flow,esser2024scaling} have shown superior performance through alternative training objective formulated in Eq. ~\ref{loss:v} while maintaining the same architecture as DiT~\cite{peebles2023scalable}. Sampling from a flow-based model can be achieved by solving the probability flow ODE.

\textbf{Mixture of Experts.} Mixture of Experts (MoE) ~\cite{shazeer2017outrageouslylargeneuralnetworks, cai2024surveymixtureexperts} is based on a fundamental insight: different parts of a model can specialize in handling distinct tasks. By selectively activating only relevant components, MoE enables efficient scaling of model capacity while maintaining computational efficiency.

MoE layers generally consist of \(N\) experts, each implemented as a Feed-Forward Network (FFN) with identical architecture, denoted by $E_1(\bfx),\dots, E_{N}(\bfx)$ with input $\bfx$. A routing matrix $ \mathbf{W}_r \in \mathbb{R}^{D\times N}$ is used to calculate token-expert affinity matrix:
\begin{equation}
    \mathbf{M} = \texttt{softmax}_E(\bfx\mathbf{W}_r), \quad \bfx \in \mathbb{R}^{B\times S\times D},
\end{equation}
where $B$ is the batch size, $S$ is the token length of one sample, $D$ is the hidden dimension, $\texttt{softmax}_E$ denotes the $\texttt{softmax}$ operation along the expert axis.
There are two common gating paradigms: Token-Choice (TC)~\cite{shazeer2017outrageouslylargeneuralnetworks, FeiDiTMoE2024} and Expert-Choice (EC) ~\cite{zhou2022mixtureofexpertsexpertchoicerouting,sun2024ecditscalingdiffusiontransformers}. For TC, each token of each sample individually selects $\texttt{top}\text{-}K$ experts via a gating function, the gating function and output of TC-MoE layers are defined as follows:
\begin{align}
    \mathbf{G}^{TC}_{s,i} &= 
    \begin{cases} 
    \mathbf{M}_{s,i}, & \mathbf{M}_{s,i} \in \texttt{top}\text{-}K(\{\mathbf{M}_{s,i}\}_{i=1}^N) \\
    0, & \text{otherwise}
    \end{cases}
    \label{eq:gating_tc}
    \\
    \mathbf{y}_s &= \sum_{i=1}^{K} \mathbf{G}^{TC}_{s,i} E(\mathbf{x}_s), \mathbf{x}_s \in \mathbb{R}^{1\times D}, s \in \{1,\dots,S\}.
    \label{eq:y_tc}
\end{align}

 Different from TC, EC makes every expert selects $K^\prime$ tokens from each sample of the input \( \bfx \in \mathbb{R}^{B\times S\times D} \). The gating function $G_{s,i}^{EC}$ can be implemented analogously to Eq.~\ref{eq:gating_tc}, with the modification that the \texttt{top} operation selects $K^\prime$ tokens along the token length dimension, \ie $S$. Similar to Eq. ~\ref{eq:y_tc}, the output for a token $\bfx_s$ of EC-MoE layers can be calculated as: $\mathbf{y}_s = \sum\limits_{i=1}^{N} \mathbf{G}^{EC}_{s,i} E(\bfx_s), \mathbf{x}_s \in \mathbb{R}^{1\times D}$. Both TC and EC struggle to achieve significant improvements comparing with dense models due to their restricted token accessibility and fixed computational patterns.

\subsection{DiffMoE: Dynamic Token Selection}
\label{sec:Method}

\textbf{Batch-level Global Token Pool.} Since MoE architectures replace FFN layers, both TC and EC paradigms in diffusion models  are inherently limited to processing tokens within individual samples, where gating mechanisms operate exclusively on tokens sharing identical conditions and noise levels. This architectural constraint not only prevents experts from learning crucial contrastive patterns but, more fundamentally, restricts their access to the global token distribution that characterizes the full spectrum of the diffusion process. To capture this essential global context, we introduce a \textit{Batch-level Global Token Pool} for DiffMoE by flattening batch and token dimensions, enabling experts to access a comprehensive token distribution spanning different noise levels and conditions. This design, which simulate the true token distribution of the entire dataset during training, can be formulated as follows:
\begin{equation}
    \bfx \in \mathbb{R}^{B\times S\times
    D} \rightarrow \bfx_{\mathrm{pool}} \in \mathbb{R}^{BS\times D}.
\end{equation}
 During the training phase, we push expert $E$ to select $K^{E}_{\text{train}}$ tokens, forcing each expert to capture the characteristics of tokens from different conditional information and noise levels, while keeping expert load balance during training. The corresponding batch-level global token-expert affinity matrix will be calculated as follows:
\begin{equation}
    \mathbf{M}^{Dy} = \bfx_{\mathrm{pool}}\mathbf{W}_r, \bfx_{\mathrm{pool}} \in \mathbb{R}^{BS\times D}, \mathbf{W}_r \in \mathbb{R}^{D\times N}.
\end{equation}
Then, using $\mathbf{M}^{Dy} \in \mathbb{R}^{BS\times N}$, the gating value of MoE and the output of DiffMoE layers can be computed as follows:
\begin{align}
\mathbf{G}^{Dy}_{s,i} & = 
\begin{cases} 
\mathbf{M}^{Dy}_{s,i}, & \mathbf{M}^{Dy}_{s,i} \in \texttt{top}\text{-} K^{E}_{\text{train}} (\{\mathbf{M}^{Dy}_{s,i}\}_{s=1}^{BS}) \\
0, & \text{otherwise}
\end{cases} 
\\ 
\mathbf{y}_s & = \sum\limits_{i=1}^{N} \mathbf{G}^{Dy}_{s,i} E(\mathbf{x}_s).
\end{align}

\begin{figure}[t]
\begin{center}
\vskip 0.05in
\centerline{\includegraphics[width=\columnwidth]{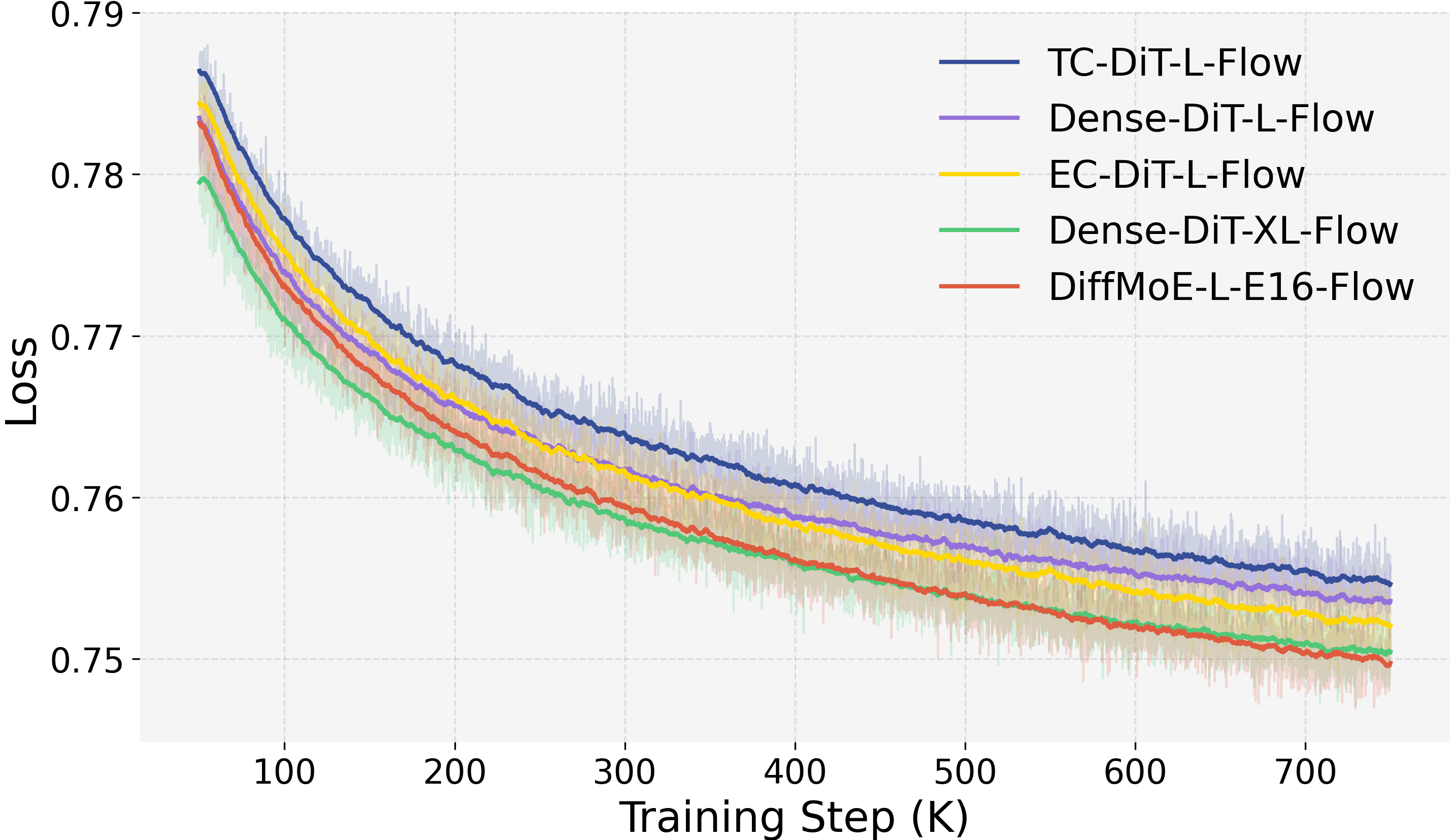}}
\caption{\textbf{Training Loss Curves of Different Flow-based Models.} DiffMoE with batch-level global token pool achieves consistently lower diffusion losses than baselines without batch-level global token pool.}

\label{fig:loss_comparison_flow}
\end{center}
\vskip -0.2in
\end{figure}

As shown in Figure~\ref{fig:loss_comparison_flow}, DiffMoE consistently achieves lower diffusion losses than all baselines.

\textbf{Capacity and Computational Cost.} To establish a rigorous and fair comparison framework, we define the capacity $C^{E}$ for a single expert $E$, which serves as a standardized metric for quantifying computational costs. This capacity metric enables fair comparisons between DiffMoE and baseline models by accurately measuring the computational resources utilized by each expert:
\begin{equation}
    C^{E} = \frac{N\times\text{\# tokens processed by } E}{\text{\# all input tokens}} = \frac{ N K^{E}}{BS},
\end{equation}

where $K^{E}$ denotes the number of tokens assigned to expert $E$, $N$ is the number of experts, and $BS$ represents the size of global token pool. Here, we define the capacity $C$ for one forward process for both training and inference phases:
\begin{equation}
 C = \frac{1}{LN}\sum\limits_{l=1}^{L} \sum\limits_{i=1}^N C^{E^l_i},
\end{equation}

where $E^l_i$ denotes $i_{\rm th}$ expert in the $l_{\rm th}$ MoE layer. 

During training phase, $C_{\rm train}$ is fixed to $1$ across all the MoE models, indicating that they keep the same computational cost as dense models. Specifically, DiffMoE keeps $K^E_{\rm train}=BS/N$ for $C_{\rm train} = 1$. TC-DiT selects the \texttt{top}-$1$ expert, while EC-DiT selects \texttt{top}-$(S/N)$ tokens per sample in a batch to ensure the same computation. During the inference phase, we compute the global average inference capacity $C_{\rm infer}^{\mathrm{avg}}$ by averaging over all timesteps:
$C_{\rm infer}^{\mathrm{avg}} = \frac{1}{T} \sum_{t=1}^{T} C^{t}_{\rm infer}$, where $C_{\rm infer}^{t}$ represents the inference capacity at sampling step $t$.

\textbf{Capacity Predictor.} Although batch-level global token routing enables efficient model training, conventional MoE inference strategies with fixed computational resource allocation fail to fully leverage its potential. This limitation stems from the static resource distribution across different noise levels and conditional information during inference. To optimize token selection, we propose a \textit{capacity predictor}---a lightweight structure that dynamically determines token selection per expert through a two-layer $\mathrm{MLP}$ with $\mathrm{SiLU}$ activations. This adaptive mechanism learns from training-time token routing patterns, efficiently distributing computational resources between complex and simple cases. Formally, let $\mathrm{CP}(\mathbf{x}_{\text{pool}})\in \mathbb{R}^{BS\times N}$ denote the predictor's output for input $\mathbf{x}_{\text{pool}}\in \mathbb{R}^{BS\times D}$:
\begin{equation}
     \mathrm{CP}(\bfx_{\text{pool}}) = \mathbf{W}_2\sigma_{\mathrm{SiLU}}(\mathbf{W}_1\bfx_{\text{pool}}).
\end{equation}

We can optimize the capacity predictor by minimizing the object function:
\begin{align}
    \mathcal{L}_{\mathrm{CP}} &=  \textit{BCELoss}(\mathbf{O}, \mathrm{CP}(\texttt{sg}[\bfx_\text{pool}]) \\
    &= -\frac{1}{LBSN} \sum_{l=1}^L\sum_{i=1}^{BS}\sum_{j=1}^N \{ \mathbf{O}^l_{i,j}\log(\mathrm{CP}(\texttt{sg}[\bfx_\text{pool}])_{i,j}) \notag \\ 
    &\quad + (1-\mathbf{O}^l_{i,j})\log(1- \mathrm{CP}(\texttt{sg}[\bfx_\text{pool}])_{i,j})
    \},
\end{align}
where \texttt{sg} denotes the \texttt{stop-gradient} operation, and $\mathbf{O} \in \mathbb{R}^{L\times BS\times N}$ is defined as follows:
\begin{equation}
    \mathbf{O}^l_{s,i} = 
    \begin{cases} 
        1, & \text{if } \bfx_{\text{pool},s} \text{ is processed by } E^l_i 
        \\
        0, & \text{otherwise}
    \end{cases}.
\end{equation}

We employ the \texttt{stop-gradient} technique to train the capacity predictor, ensuring it focuses solely on the input features at the current layer while preventing it from interfering with the training of the main diffusion transformers. Therefore, $\mathcal{L}_{\rm CP}$ will not affect actual diffusion loss. During inference, the capacity predictor determines the inference capacity $C_{\rm infer}^{E^l_i, t}$ for each expert \(E^l_i\) at timestep $t$ based on a threshold. Let $\tau_{E^l_i}$ denote the threshold of $E^l_i$. Using $\mathcal{T} = \{\tau_{E^l_i} \mid i \in \{1, \dots, N\},\ l \in \{1, \dots, L\}\}$, the model achieves an adaptive $C_{\rm infer}^{E^l_i,t}$ allocation at sampling step $t$ tailored to different input tokens as follows:
\begin{align}
    &   C_{\rm infer}^{E^l_i,t}(\tau_{E^l_i}) = \sum\limits_{s=1}^S \delta^l_{s,i}  \notag \\ 
    & \text{where } 
        \delta^l_{s,i}=  \begin{cases} 
            1, & \text{if } \mathrm{CP}(\bfx_{\rm pool})_{s,i} >  \tau_{E^l_i}
            \\
            0, & \text{otherwise}
    \end{cases}.
\end{align}
Then we can calculate $C^{\rm avg}_{\rm infer}$ \wrt $\mathcal{T}$ 
\begin{equation}
   C^{\rm avg}_{\rm infer}(\mathcal{T}) = \frac{1}{TNL} \sum_{t=1}^T \sum_{i=1}^N \sum_{l=1}^L C_{\rm infer}^{E^l_i,t}(\tau_{E^l_i}). 
\end{equation}

\textbf{Dynamic Threshold.} We can set the threshold $\mathcal{T}$ to control the number of tokens processed by each expert during inference. It is evident that \(\# \text{tokens processed during inference}\), decreases as $\tau_{E^l_i}$ increases for all expert. We can adjust $\mathcal{T}$  flexibly to achieve a better trade-off between computational complexity and generation quality. We employ two distinct approaches for threshold $\mathcal{T}$ determination: \textit{Interval Search} and \textit{Dynamic Threshold}. The interval search method addresses an optimization problem formulated as follows:
\begin{equation}
    \min_{\mathcal{T}} \text{FID}(\mathcal{T}) \quad \text{subject to } C^{\rm avg}_{\rm infer}(\mathcal{T}) \le 1.
    \label{eq:interval_search}
\end{equation}

To simplify the optimization problem, we assume that $\tau = \gamma < 1, \forall \tau \in \mathcal{T}$, where $\gamma$ is a constant in our experiments.

However, interval search method is labor-intensive and time-consuming, making it impractical for real-world applications. To address this limitation, we propose a dynamic threshold method that automatically maintains thresholds (denoted as $\mathcal{T}^{Dy} = \{\tau^{Dy}_{E^l_i} \mid i \in \{1, \dots, N\},\ l \in \{1, \dots, L\}\}$) for all experts during the training phase. To ensure the inference computational cost approximates the training cost (\ie $ C^{\rm avg}_{\rm infer} \approx 1$), we employ the Exponential Moving Average (EMA) technique as follows:
\begin{align}
    & \mathrm{Quantile}_{E^l_i} \leftarrow \mathrm{CP}(\bfx_{\rm pool})_{s_k, i}, \notag \\
    & \tau^{Dy}_{E^l_i} \leftarrow \alpha \cdot \tau^{Dy}_{E^l_i} + (1 - \alpha) \cdot \mathrm{Quantile}_{E^l_i},
\end{align}
where $s_k$ denotes the $k_{\text{th}}$ value in descending order, $\alpha$ is a constant which is equals to 0.95 in our experiments. \\

\section{Experiments}
\label{sec:exp} We evaluate DiffMoE on class-conditional image generation across three key aspects: (1) \textbf{Training and Inference Performance}, (2) \textbf{Dynamic Computation}, and (3) \textbf{Scalability}. Our experiments demonstrate DiffMoE's effectiveness through extensive analysis. Additionally, we verify its adaptability on text-to-image generation tasks.

\begin{table*}[t]
  \centering
\caption{\textbf{DiffMoE Model Configurations}. Hyperparameter settings and computational specifications for class-conditional models. See Appendix~\ref{appendix:sec_cap} for activated parameter calculations.}

  \adjustbox{width=\linewidth}
  {

    \begin{tabular}{lccccccc}\toprule
    \multirow{1}{*}{Model Config} 
    & \multirow{1}{*}{\#Avg. Activated Params ($C^{\rm avg}_{\rm infer}=1$).} 
    & \multirow{1}{*}{\#Total Params.}
    & \multirow{1}{*}{\#Blocks L} 
    & \multirow{1}{*}{\#Hidden dim. D}
    & \multirow{1}{*}{\#Head n}
    & \multirow{1}{*}{\#Experts}
    & \multirow{1}{*}{$C_{\rm train}$}
    \\\midrule
    DiffMoE-S-E16 & 32M   & 139M    & 12  & 384  & 6  & 16 & 1 \\
    DiffMoE-B-E16 & 130M  & 555M    & 12  & 768  & 12 & 16 & 1 \\
    DiffMoE-L-E8  & 458M  & 1.176B  & 24  & 1024 & 16 & 8 & 1 \\
    DiffMoE-L-E16 & 458M  & 1.982B  & 24  & 1024 & 16 & 16 & 1 \\\bottomrule
    \end{tabular}%
    }
  \label{tab:model_arch}%
  \vspace{-10pt}
\end{table*}%

\subsection{Experiment Setup}

\textbf{Baseline and Model architecture.} We compare with Dense-DiT trained by denoising score matching~\cite{peebles2023scalable} and flow matching~\cite{ma2024sit}, TC-DiT~\cite{FeiDiTMoE2024}, and EC-DiT~\cite{sun2024ecditscalingdiffusiontransformers}. For fair comparison, we reimplement TC-DiT and EC-DiT based on their public repositories and pseudo-codes while maintaining identical activated computation. Models are named as: \textbf{[Model]-[Size]-[\# Experts]-[Training Type]}.
For class-conditional generation, we replace even FFN layers with MoE layers containing $N$ identical FFN components~\cite{lepikhin2020gshardscalinggiantmodels}, while maintaining the original DiT architecture~\cite{peebles2023scalable}, details of architecture are shown in Table~\ref{tab:model_arch}. For text-to-image generation, we add cross-attention modules~\cite{rombach2022high}, where DiffMoE-E16-T2I-Flow activates 1.2B parameters (matching Dense-DiT-T2I-Flow) from total 4.6B parameters. Full details are in Appendix~\ref{appendix:sec_implementation_details}.

\textbf{Evaluation.} We evaluated DiffMoE through both quantitative and qualitative metrics. Quantitatively, we used FID50K~\cite{fid} with 250 DDPM/Euler steps for class-conditional generation, and compared with SiT~\cite{ma2024sit} using Heun sampler at 125 steps. For text-to-image generation, we employed GenEval metrics~\cite{ghosh2023genevalobjectfocusedframeworkevaluating}. Training losses were analyzed to validate the model's convergence behavior. Additionally, we assessed the model's performance through visual inspection of samples generated from diverse prompts.

\begin{table*}[t]
  \centering
    \caption{\textbf{State-of-the-art Comparison}. Evaluation on ImageNet $256\times 256$ class-conditional generation. DiffMoE achieves better FID with fewer parameters. -G/-U denotes with/without guidance~\cite{ho2022classifierfreediffusionguidance}. $^\dagger$: results from~\cite{ma2024sit} (DDPM) and~\cite{peebles2023scalable} (Flow). $^*$: our reproduction.  \textbf{Bold} indicates best performance in each cell.}
  \adjustbox{width=\linewidth}
  {

    \begin{tabular}{lccccc}\toprule
    \multirow{1}{*}{Diffusion Models (7000K)} & \multirow{1}{*}{\# Avg. Activated Params.} & \multirow{1}{*}{FID$\downarrow$} & \multirow{1}{*}{IS$\uparrow$} & \multirow{1}{*}{Precision$\uparrow$} & \multirow{1}{*}{Recall$\uparrow$} \\\midrule

    \textcolor{gray}{Dense-DiT-XL-Flow-U$^\dagger$~\cite{peebles2023scalable}} & \textcolor{gray}{675M}  & \textcolor{gray}{9.35}  & \textcolor{gray}{126.06}  & \textcolor{gray}{0.67}  & \textcolor{gray}{0.68}  \\
    Dense-DiT-XL-Flow-U$^*$~\cite{peebles2023scalable} & 675M  & \textbf{9.47}  & 115.58  &  0.67  & 0.67  \\
    \rowcolor{Gray} DiffMoE-L-E8-Flow-U & 458M   & {9.60}  & {131.46}  & 0.67  & 0.67  \\\midrule

    \textcolor{gray}{Dense-DiT-XL-Flow-G$^\dagger$ (cfg=1.5, ODE)~\cite{ma2024sit}} & \textcolor{gray}{675M}  & \textcolor{gray}{2.15}    & \textcolor{gray}{254.9}  & \textcolor{gray}{0.81}  & \textcolor{gray}{0.60}  \\
    Dense-DiT-XL-Flow-G$^*$ (cfg=1.5, ODE)~\cite{ma2024sit} & 675M  & 2.19    & 272.30  &  0.83  & 0.58 \\
    \rowcolor{Gray} DiffMoE-L-E8-Flow-G (cfg=1.5, ODE) & 458M   & \textbf{2.13}  & 274.39  & 0.81  & {0.60}  \\\midrule
     
    \textcolor{gray}{Dense-DiT-XL-DDPM-U$^\dagger$~\cite{peebles2023scalable}} & \textcolor{gray}{675M}  & \textcolor{gray}{9.62}  & \textcolor{gray}{121.50}  & \textcolor{gray}{0.67}  & \textcolor{gray}{0.67}  \\
    Dense-DiT-XL-DDPM-U$^*$~\cite{peebles2023scalable} & 675M  & 9.62  & 123.19  &  0.66  & {0.68}  \\
    \rowcolor{Gray} DiffMoE-L-E8-DDPM-U & 458M   & \textbf{9.17}  & {131.10}  & {0.67}  & 0.67  \\\midrule

    \textcolor{gray}{Dense-DiT-XL-DDPM-G$^\dagger$ (cfg=1.5)~\cite{peebles2023scalable}} & \textcolor{gray}{675M}  & \textcolor{gray}{2.27}  & \textcolor{gray}{278.2}  & \textcolor{gray}{0.83}  & \textcolor{gray}{0.57}  \\
    Dense-DiT-XL-DDPM-G$^*$ (cfg=1.5)~\cite{peebles2023scalable} & 675M  & 2.32  & 279.18  & {0.83}  & 0.57  \\
    \rowcolor{Gray} DiffMoE-L-E8-DDPM-G (cfg=1.5) & 458M   &     \textbf{2.30}  & {284.78} & 0.82  & {0.59}  \\
 
     \bottomrule
    \end{tabular}%
    }
  \label{tab:sota}%
  \vskip -0.2in
\end{table*}%

\begin{table}[t]
  \centering
\caption{\textbf{Baseline Model Comparisons.} DiffMoE-L-E16-Flow achieves best FID50K (14.41 w/o CFG) among TC, EC, and Dense variants. DDPM results in Appendix~\ref{tab:compare_w_baseline_ddpm}.}
  \adjustbox{width=\linewidth}
  {

    \begin{tabular}{lcccc}\toprule
    \multirow{1}{*}{Model  (700K)} 
    & \multirow{1}{*}{\# Avg. Activated Params.} 
    & \multirow{1}{*}{$C^{\rm avg}_{\rm infer}$} 
    & \multirow{1}{*}{FID50K $\downarrow$}
  
    \\\midrule
    TC-DiT-L-E16-Flow & 458M & 1 & 19.06  \\
    EC-DiT-L-E16-Flow & 458M & 1 & 16.12 \\
    Dense-DiT-L-Flow & 458M & 1 & 17.01 \\
    Dense-DiT-XL-Flow & 675M & 1 & \underline{14.77} \\
     \rowcolor{Gray} DiffMoE-L-E16-Flow & 454M & 0.95 & \textbf{14.41} \\\bottomrule
    \end{tabular}%
    }
  \label{tab:compare_w_baseline}%
  \vskip -0.19in
\end{table}%

\begin{table}[t]
  \vskip 0.05in
  \centering
  \caption{ \textbf{Ablation of Capacity Predictor.}  
    We employ the capacity predictor to perform dynamic token selection, comparing it with the fixed TopK token selection method.}
      \adjustbox{width=\linewidth}
      {
        \begin{tabular}{lcccc}\toprule
        \multirow{1}{*}{Model  (700K)} 
        & \multirow{1}{*}{Capacity Predictor} 
        & \multirow{1}{*}{$C^{\rm avg}_{\rm infer}$} 
        & \multirow{1}{*}{FID50K $\downarrow$}
        \\\midrule
        DiffMoE-L-E16-Flow & w/o  & 0.9 & 16.63 \\
        DiffMoE-L-E16-Flow & w/o  & 0.95 & 15.99 \\
        DiffMoE-L-E16-Flow & w/o & 1 & \underline{15.25} \\
         \rowcolor{Gray} DiffMoE-L-E16-Flow &  w & 0.95 & \textbf{14.41} \\
        \bottomrule
        \end{tabular}%
        }
    \label{tab:dy_vs_topk_token_selection}%
    \vskip -0.05in
\end{table}%

\begin{figure}[t]
\begin{center}
\centerline{\includegraphics[width=\columnwidth]{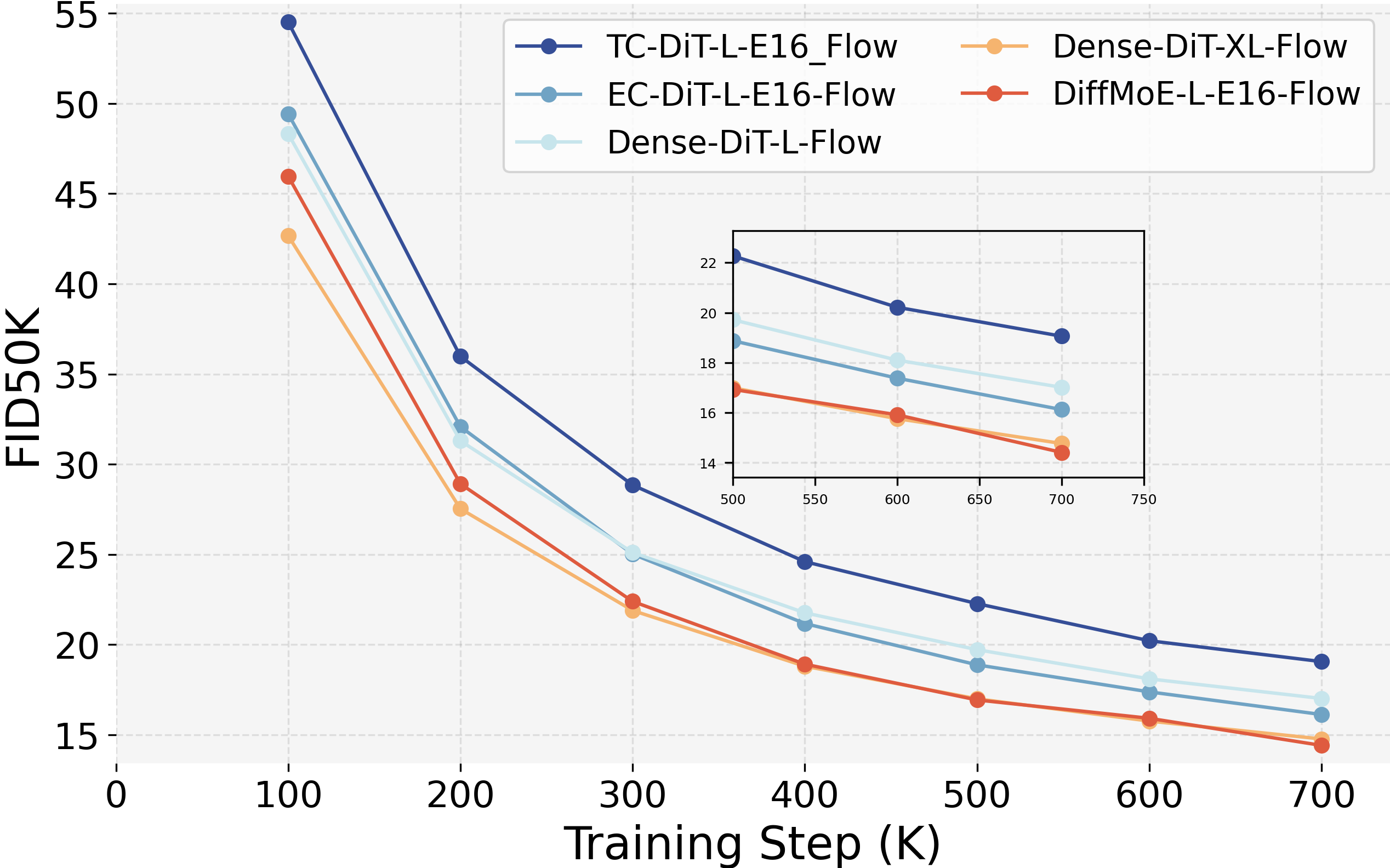}}
\caption{\textbf{Comparisons with the Baseline Models.} We compare TC, EC, and Dense Models. DiffMoE-L-E16-Flow even surpasses the DenseDiT-XL-Flow (1.5x params) by achieving the best quality (14.41 FID50K w/o CFG at 700K). The results of the DDPM method remain consistent with those provided in the Appendix~\ref{appendix:sec_ddpm_reults}. }
\label{fig:comparing_w_baseline}
\end{center}
\vskip -0.2in
\end{figure}

\subsection{Main Results: Class-conditional Image Generation}

Class-conditional image generation is a task of synthesizing images based on specified class labels.

\textbf{Comparison with Baseline.} DiffMoE-L-E16 demonstrates superior efficiency by outperforming Dense-DiT-XL (with $1.5\times$ parameters) after 700K steps, as shown in Table~\ref{tab:compare_w_baseline}. It consistently achieves lower training loss compared to variants TC-DiT-L-E16, EC-DiT-L-E16 and Dense-DiT-L (Figure~\ref{fig:comparing_w_baseline},~\ref{fig:ddpm_compare_with_baseline},~\ref{fig:loss-comparison}). These improvements hold across both DDPM and Flow Matching paradigms while maintaining equivalent activated parameters. With more training time computation, DiffMoE-L-E16 can outperform  Dense-DiT-XXXL  (with $3\times$ parameters) as shown in Table~\ref{tab:sota_supp}.

\begin{figure}[t]
    \centering
    \begin{minipage}{\columnwidth} %
        \includegraphics[width=\linewidth]{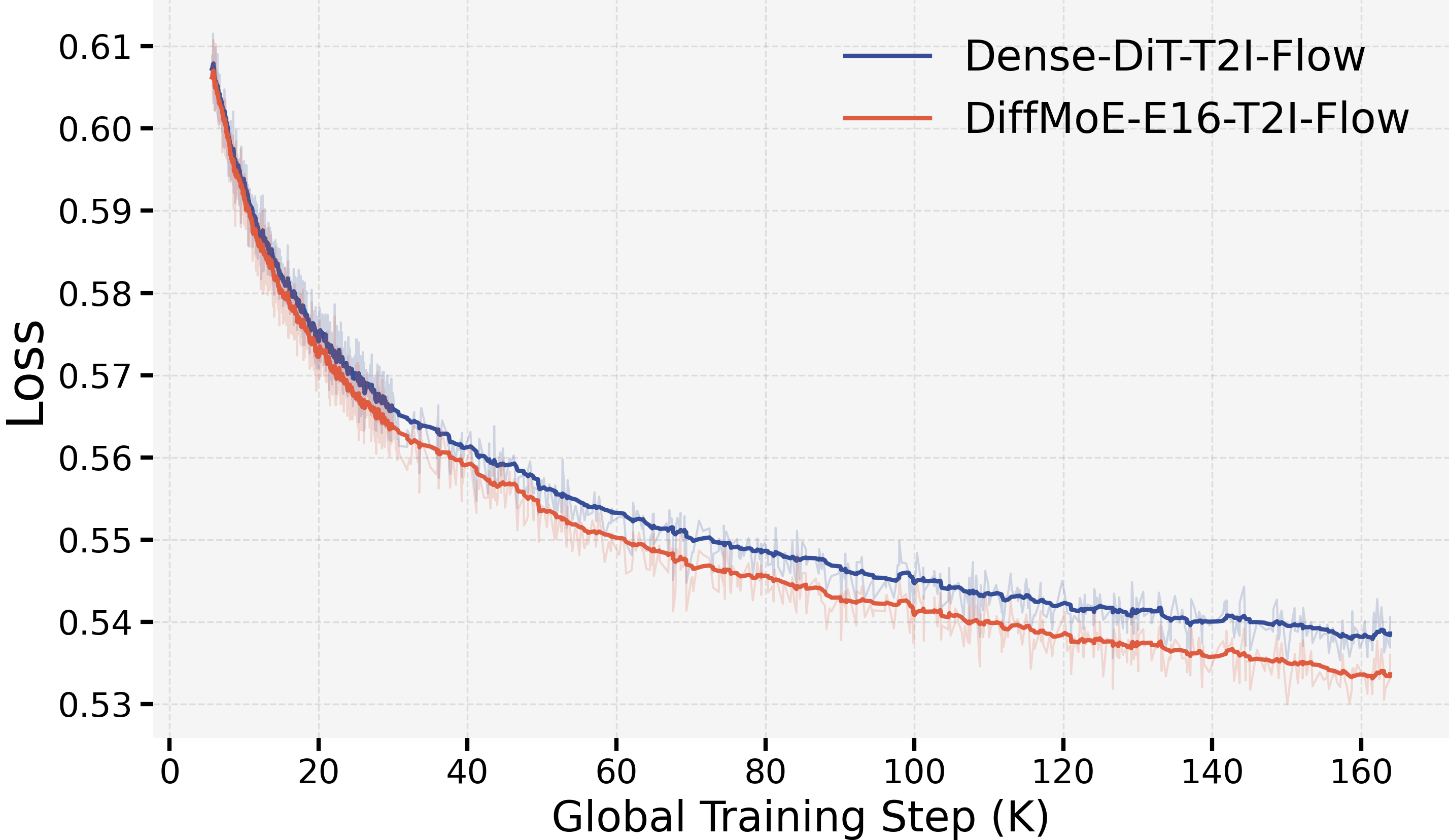}
        \centering %
    \end{minipage}
    \caption{\textbf{Text-to-Image Generation Loss Curves.} Training loss comparison between DiffMoE-E16-T2I-Flow and Dense-DiT-T2I-Flow models over 160K steps. DiffMoE consistently achieves lower loss values, demonstrating superior convergence efficiency compared to the dense baseline.}
    \label{fig:loss-t2i}
    \vskip -0.2in %
\end{figure}

\setlength{\parskip}{0.5em} 

\textbf{Comparison with SOTA.} After 7000K steps, DiffMoE-L-E8 achieves state-of-the-art FID50K scores of DDPM/Flow (2.30/2.13)  with cfg=1.5, surpassing Dense-DiT-XL (2.32/2.19) as shown in Table~\ref{tab:sota}. All evaluations follow DiT's~\cite{peebles2023scalable} protocol. Generated C2I samples are shown in Figure~\ref{fig:flow_c2i_viz} and~\ref{fig:ddpm_c2i_viz}.

\subsection{Main Results: Text-to-Image Generation}
In Text-to-Image (T2I) generation, DiffMoE demonstrates superior performance over the Dense Model across multiple metrics. Without/With supervised fine-tuning (SFT), DiffMoE achieves GenEval scores of 0.44/0.51, outperforming the Dense Model's 0.38/0.49, with improvements across nearly all sub-metrics (Appendix Table~\ref{tab:geneval_t2i}). And as shown in Figure~\ref{fig:loss-t2i}, DiffMoE maintains lower training losses while using the same activated parameters. Qualitative results in Figure~\ref{fig:flow_t2i_compare_viz} validate DiffMoE's ability to generate higher-quality images without SFT. Additional samples from SFT-enhanced models are shown in Figure~\ref{fig:flow_t2i_viz}.

\subsection{Dynamic Computation Analysis}
For the convenience of elaboration, we use the flow matching training method to do the following analysis while DDPM results are also provided in the Appendix~\ref{appendix:sec_ddpm_reults}.

\textbf{Analysis of Inference Capacity.} DiffMoE-L-E16-Flow demonstrates superior parameter efficiency with its inference capacity ($C^{\rm avg}_{\rm infer}$) being 1 less than TC-DiT and EC-DiT, while achieving better performance, as shown in Table~\ref{tab:compare_w_baseline}. Notably, with only 454M average activated parameters, our model outperforms Dense-DiT-XL-Flow (675M parameters), highlighting the effectiveness of dynamic expert allocation. Detailed analysis of average activated parameters is provided in the Appendix ~\ref{appendix:sec_cap}.

\begin{table*}[t]
  \centering
    \caption{\textbf{Parameter Scaling Behavior} of Diffusion Models on ImageNet 256×256 Class-Conditional Generation. This table evaluates the impact of parameter scaling on model performance. DiffMoE demonstrates superior FID scores with fewer parameters after 3000K training steps, highlighting its efficiency. Models with -G employ classifier-free guidance~\cite{ho2022classifierfreediffusionguidance}. The base parameter configuration is fixed at 458 million (458M). \textbf{Bold} indicates the best performance in each metric, while \underline{underline} denotes the second-best performance.}
  \adjustbox{width=\linewidth}
  {

    \begin{tabular}{llcccc}\toprule
    \multirow{1}{*}{Diffusion Models (3000K)} & \multirow{1}{*}{\# Avg. Activated Params.} & \multirow{1}{*}{FID$\downarrow$} & \multirow{1}{*}{IS$\uparrow$} & \multirow{1}{*}{Precision$\uparrow$} & \multirow{1}{*}{Recall$\uparrow$} \\\midrule

    Dense-DiT-XL-FlowG (cfg=1.5, ODE) & 675M \blue{1.5x}  & 2.52  &  273.78  &  0.84  & 0.56  \\
    Dense-DiT-XXL-Flow-G (cfg=1.5, ODE) & 951M \blue{2x}  & 2.41  &  281.96  &  0.84  & 0.57  \\
    Dense-DiT-XXXL-Flow-G (cfg=1.5, ODE) & 1353M \blue{3x}  & 2.37  & 291.29  &  0.84  & 0.57  \\
    
    \rowcolor{Gray} DiffMoE-L-E8-Flow-G (cfg=1.5, ODE) & 458M \blue{1x}   & 2.40  & 280.30  &  0.83  & 0.57  \\
    \rowcolor{Gray} DiffMoE-L-E16-Flow-G (cfg=1.5, ODE) & 458M \blue{1x}   & \underline{2.36}  & 287.26  &  0.83  & 0.58  \\
    \rowcolor{Gray} DiffMoE-XL-E16-Flow-G (cfg=1.5, ODE) & 675M \blue{1.5x}   & \textbf{2.30}  & 291.23  &  0.83  & 0.58·  \\\bottomrule
    \end{tabular}%
    }
  \label{tab:sota_supp}%
  \vskip -0.2in
\end{table*}%

\begin{figure}[t]
\begin{center}
\centerline{\includegraphics[width=\columnwidth]{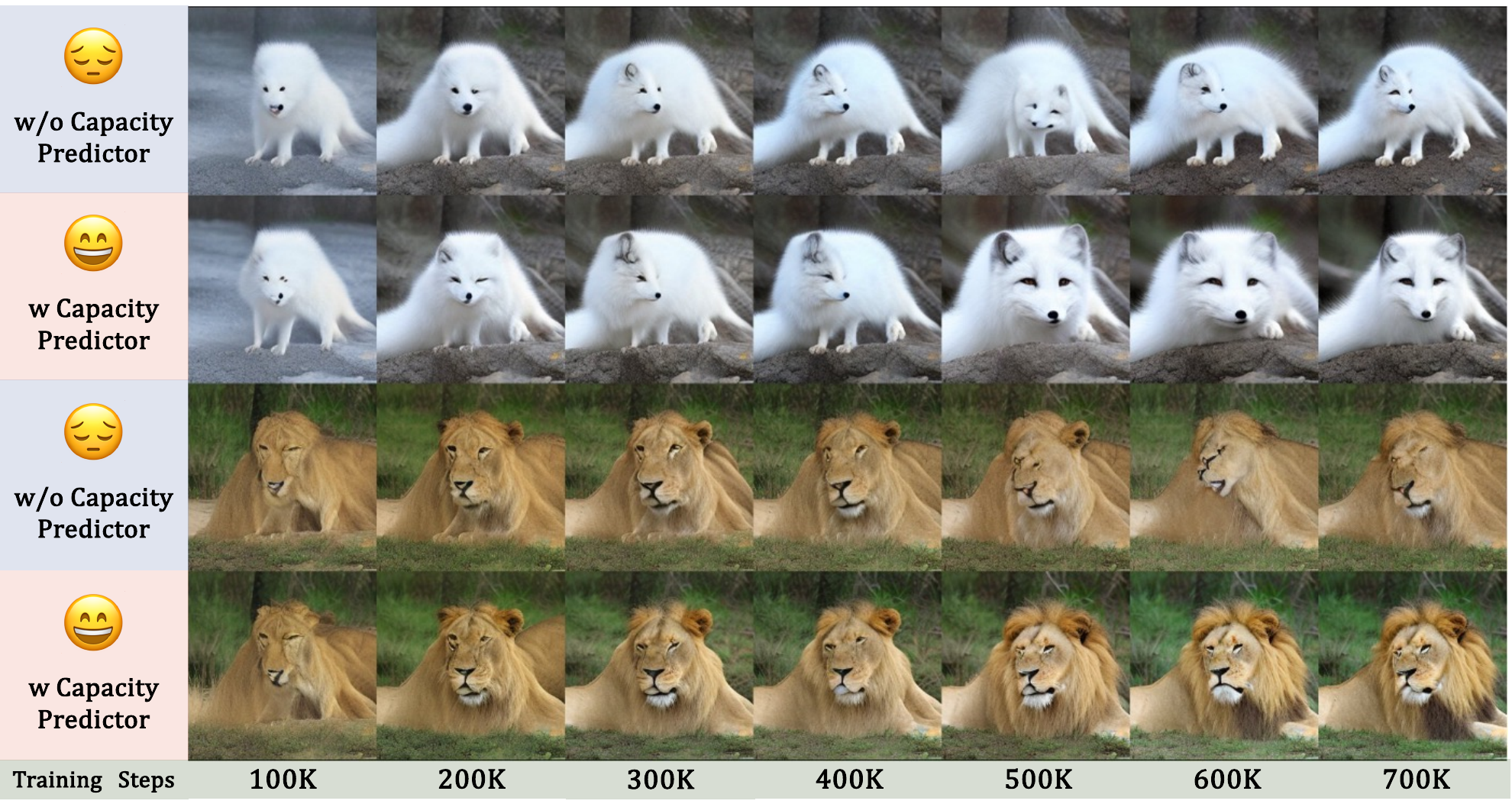}}
\caption{\textbf{Training and Inference Gap.} Comparison of sampling strategies with DiffMoE-L-E16-Flow (Batch Size = 1). For each group,   
\textbf{Top:} Sampling w/o Capacity Predictor (with Fixed TopK Method.)
\textbf{Bottom:} Sampling with Capacity Predictor.}
\label{fig:train_infer_gap}
\end{center}
\vskip -0.4in
\end{figure}

\textbf{Ablation of Capacity Predictor.}
Dynamic token selection through our capacity predictor demonstrates superior performance over traditional static topK token selection, as shown in Table~\ref{tab:dy_vs_topk_token_selection}. This improvement stems from the predictor's ability to intelligently allocate more computational resources to challenging tasks. The capacity predictor plays a crucial role in unleashing DiffMoE's full potential by dynamically adjusting resource allocation, which is particularly important for optimizing inference efficiency (Section~\ref{sec:Method}). Without such adaptive mechanism, DiffMoE suffers from severe quality degradation due to sub-optimal resource utilization, as illustrated in Figure~\ref{fig:train_infer_gap}.

\begin{figure}[t]
\begin{center}
\centerline{\includegraphics[width=\columnwidth]{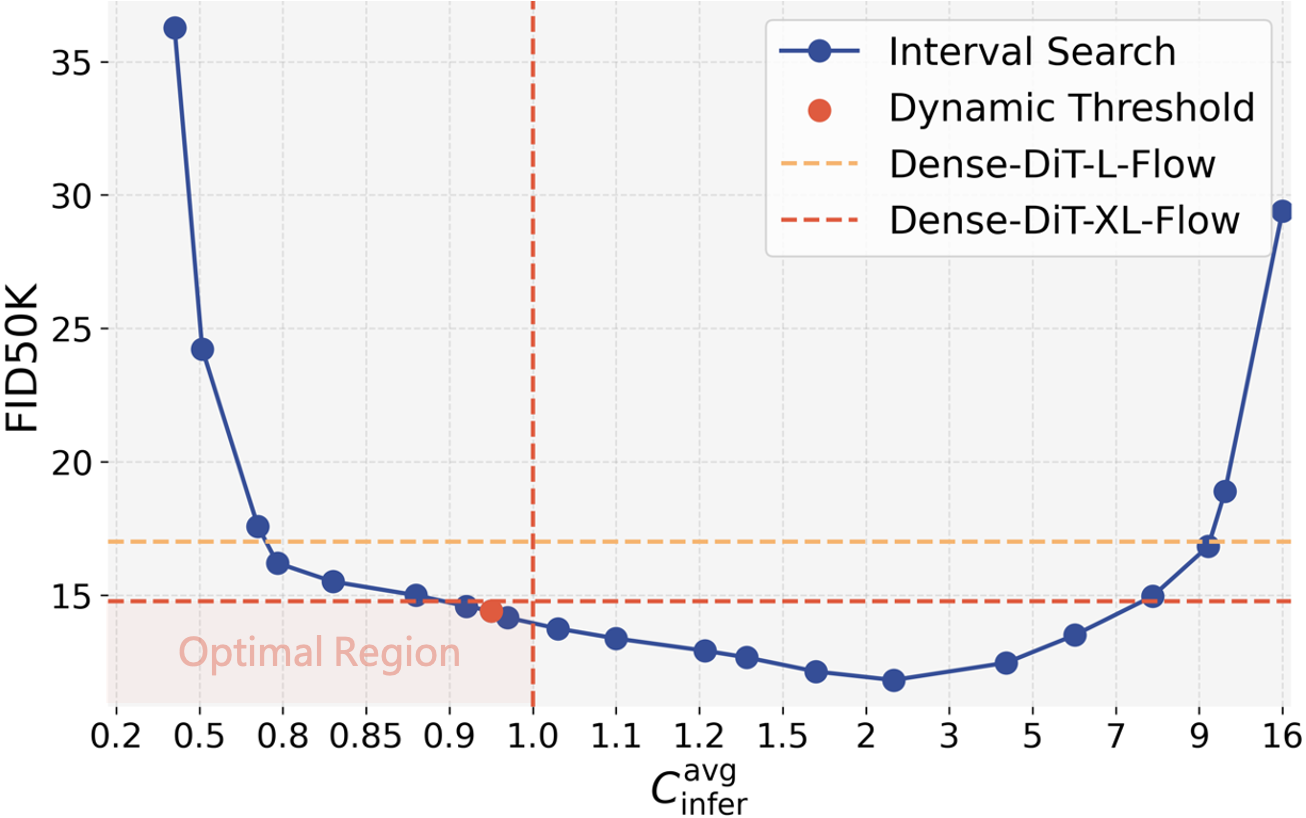}}
\caption{\textbf{Different Threshold Methods.} We employ two distinct approaches for threshold determination: \textcolor{myred}{dynamic threshold (red point)} and \textcolor{myblue}{interval search  (blue points)}. Visualization using DiffMoE-
L-E16-Flow (700K).  }
\label{fig:diff_trd_method}
\end{center}
\vskip -0.4in
\end{figure}

\textbf{Interval Search vs. Dynamic Threshold.}
Both interval search and dynamic threshold methods achieve optimal performance in DiffMoE-L-E16-Flow, with the dynamic threshold ($\mathcal{T}^{Dy}$) emerging as our preferred approach due to its elegance and efficiency. Through interval search from $0.0$ to $0.999$, we identify an optimal threshold ($\gamma \approx 0.4$) that minimizes FID while maintaining $C^{\rm avg}_{\rm infer}\le 1$. Meanwhile, the dynamic threshold automatically maintains $C^{\rm avg}_{\rm infer} \approx 1$ during inference, achieving comparable FID scores within the optimal region, as shown in Figure~\ref{fig:diff_trd_method} and Table~\ref{tab:diff_trd_method}. Our experiments reveal a U-shaped relationship between FID and $C^{\rm avg}_{\rm infer}$, indicating that both over-activation and under-activation of parameters degrade performance. Both methods successfully identify thresholds within the optimal region, but the dynamic threshold's straightforward implementation and computational efficiency make it our default choice throughout this paper.

\textbf{Harder Work Needs More Computation.} Figure~\ref{fig:teaser} demonstrates that different classes require varying computational resources during generation. By analyzing 1K class labels and ranking their $C^{\rm avg}_{\rm infer}$, we observe distinct patterns in computational demands. The most challenging cases typically involve objects with precise details, complex materials, structural accuracy, and specific viewing angles (e.g., technical instruments, detailed artifacts). In contrast, natural subjects like common animals (birds, dogs, cats) generally require less computation. Figures~\ref{fig:top10_hardest} and~\ref{fig:top10_easiest} display the top-10 most and least computationally intensive classes for both flow-based and DDPM models, respectively.

\begin{figure}[t]
\begin{center}
\centerline{\includegraphics[width=\columnwidth]{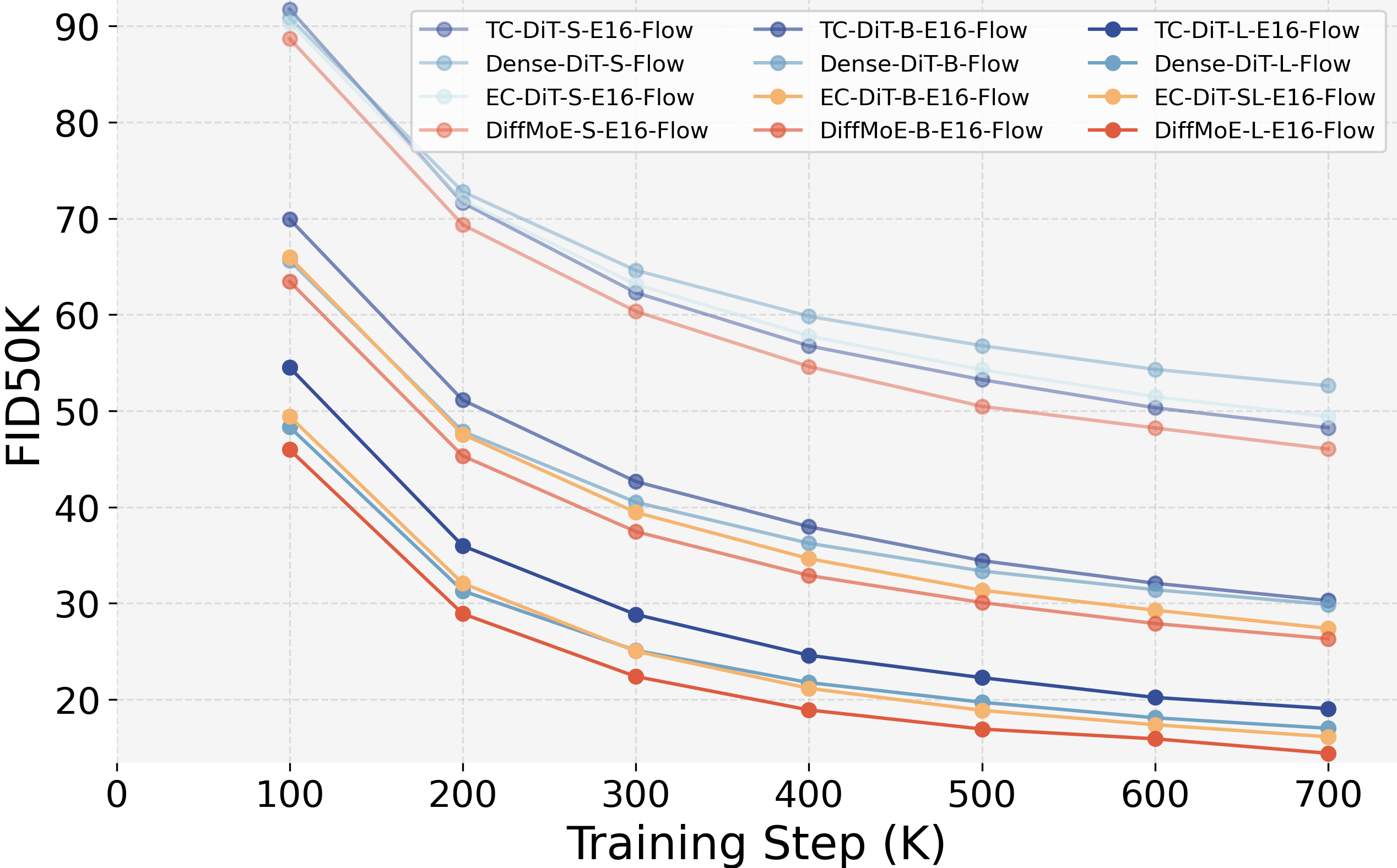}}
\caption{\textbf{Scaling Model Size.} We analyze the impact of model size scaling by plotting FID50K scores across training steps. DiffMoE consistently outperforms the corresponding baseline models across all scales (S/B/L).}
\label{fig:scale_model_size}
\end{center}
\vskip -0.3in
\end{figure}

\subsection{Scaling Behavior}
\textbf{Scaling the Model Size.} DiffMoE demonstrates consistent performance improvements across small (S), base (B), and large (L) configurations, with activated parameters of 32M, 130M, and 458M respectively (Figure~\ref{fig:scale_model_size}).

\begin{figure}[t]
\begin{center}
\centerline{\includegraphics[width=\columnwidth]{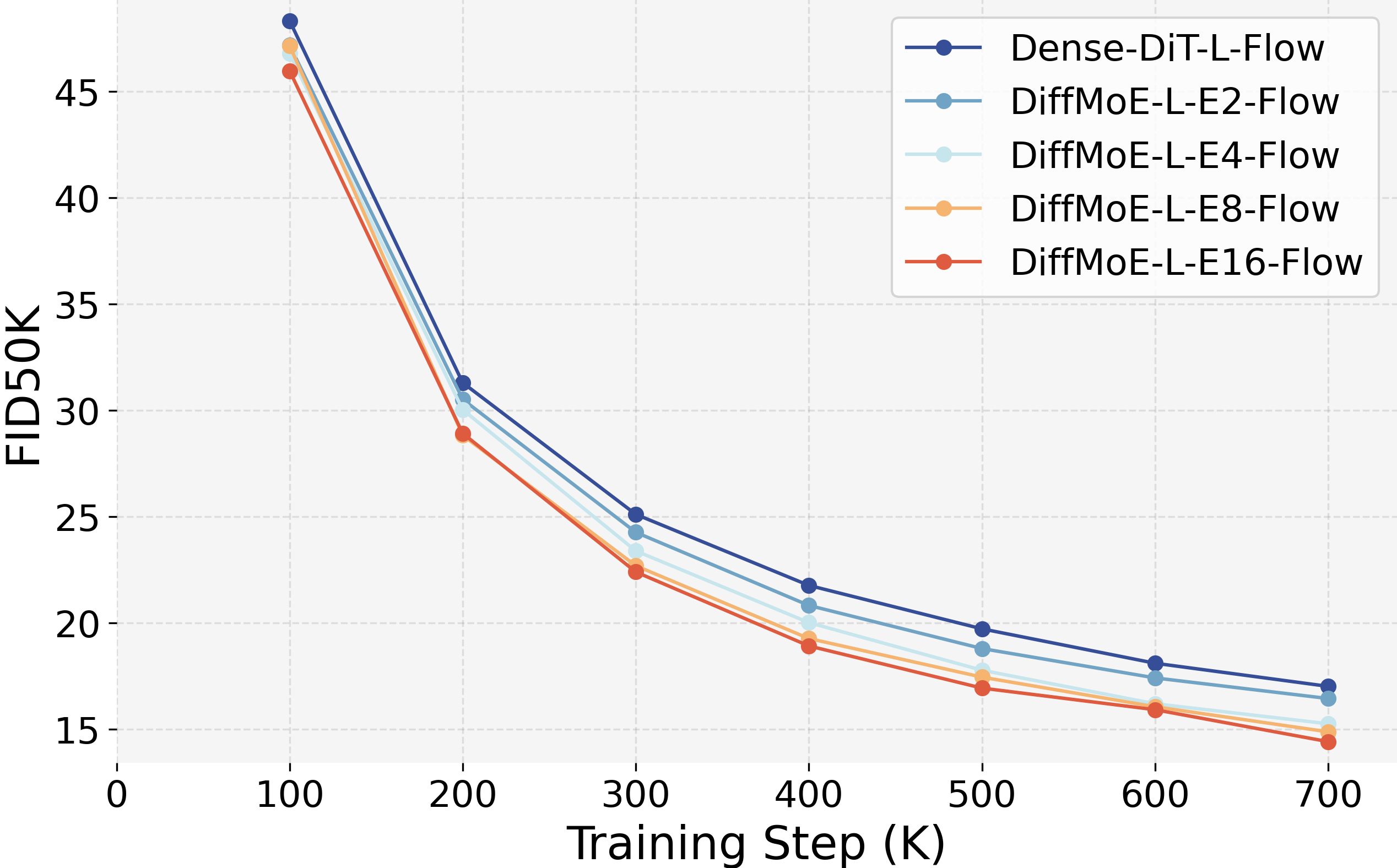}}

\caption{\textbf{Scaling Number of Experts.} Comparison of FID50K scores during training between Dense-DiT-L-Flow (E1) and models with increasing expert counts (E2, E4, E8, E16).}

\label{fig:scale_num_experts}
\end{center}
\vskip -0.4in
\end{figure}

\textbf{Scaling Number of Experts.} As shown in Figure~\ref{fig:scale_num_experts}, model performance improves consistently when scaling experts from 2 to 16, with diminishing returns between E8 and E16. Based on this analysis, we trained DiffMoE-L-E8 for 7000K iterations, achieving optimal performance-efficiency trade-off and state-of-the-art results.

\textbf{Scaling Parameter Behavior.} To explore the upper limits of DiffMoE and quantify its performance efficiency, we scaled the model to larger sizes and trained them for 3000K steps. As illustrated in Table~\ref{tab:sota_supp}, DiffMoE-L-E16-Flow achieves the best performance among the evaluated models. Notably, DiffMoE-L-E16 surpasses the performance of Dense-DiT-XXXL-Flow, which uses 3x the parameters, while operating with only 1x the parameters. This highlights the exceptional parameter efficiency and scalability of DiffMoE.

\section{Related Works}
\textbf{Diffusion Models.} Diffusion models~\cite{ho2020denoising, podell2023sdxl, peebles2023scalable, esser2024scalingrectifiedflowtransformers} have emerged as the dominant paradigm in visual generation in recent years. These models transform gaussian distribution into target data distribution through iterative processes, with two primary training paradigms: Denoising Diffusion Probabilistic Models (DDPM) trained via score-matching~\cite{ho2020denoising, song2021score}, which learns the inverse of a diffusion process and Rectified Flow approaches optimized through flow-matching~\cite{lipman2022flow, ma2024sit, esser2024scalingrectifiedflowtransformers}, which is a more generic modeling techinique and can construct a straight probaility path connecting data and noise. We implement DiffMoE using both paradigms, demonstrating its versatility across these complementary training methodologies.

\textbf{Mixture of Experts.}
Mixture of Experts (MoE)~\cite{shazeer2017outrageouslylargeneuralnetworks, lepikhin2020gshardscalinggiantmodels} enables efficient model scaling through conditional computation by selectively activating expert subsets. This approach has demonstrated remarkable success in Large Language Models (LLMs), as evidenced by cutting-edge implementations like DeepSeek-V3~\cite{deepseekai2024deepseekv3technicalreport}, minimax-01~\cite{minimax2025minimax01scalingfoundationmodels}, and OLMOE~\cite{muennighoff2024olmoeopenmixtureofexpertslanguage}.Recent works have explored incorporating MoE architectures into diffusion models, but face several limitations. MEME~\cite{lee2023multiarchitecturemultiexpertdiffusionmodels}, eDiff-I~\cite{balaji2022eDiff-I}, and ERNIE-ViLG 2.0~\cite{feng2023ernievilg20improvingtexttoimage} restrict experts to specific timestep ranges. SegMoE~\cite{segmoe} and DiT-MoE~\cite{FeiDiTMoE2024} suffer from expert utilization imbalance due to isolated token processing. While EC-DiT~\cite{sehwag2024stretchingdollardiffusiontraining,sun2024ecditscalingdiffusiontransformers} recognizes complex tokens' need for additional computation, it constrains token selection within individual samples and requires longer training for marginal improvements. These approaches, by limiting global token distribution across noise levels and conditions, fail to capture diffusion processes' inherent heterogeneity. DiffMoE addresses these challenges through batch-level global token pool for training, and dynamically adapting computation to both noise levels and sample complexity for inference.

\section{Conclusion}
In this work, we introduce DiffMoE, a simple yet powerful approach to scaling diffusion models efficiently through dynamic token selection and global token accessibility. Our method effectively addresses the uniform processing limitation in diffusion transformers by leveraging specialized expert behavior and dynamic resource allocation. Extensive experimental results demonstrate that DiffMoE substantially outperforms existing TC-MoE and EC-MoE methods, as well as dense models with 1.5× parameters, while maintaining comparable computational costs. The effectiveness of our approach not only validates its utility in class-conditional generation but also positions DiffMoE as a key enabler for advancing large-scale text-to-image and text-to-video generation tasks. While we excluded modern MoE enhancements for fair comparisons, integrating advanced techniques like Fine-Grained Expert~\cite{yang2024xmoesparsemodelsfinegrained} and Shared Expert~\cite{dai2024deepseekmoeultimateexpertspecialization} presents compelling opportunities for future work. These architectural improvements, combined with DiffMoE's demonstrated scalability and efficiency, could significantly advance the development of more powerful world simulators in the AI landscape.

\section*{Impact Statement}

DiffMoE represents a significant advancement in visual content generation, demonstrating exceptional scalability to billion-parameter architectures while maintaining computational efficiency. 
However, like other powerful AI models, it raises important ethical considerations that warrant careful attention. These include potential misuse for generating misleading content, privacy concerns regarding training data and generated content, environmental impact of large-scale model training, and broader societal implications of automated content generation.
We are committed to addressing these challenges through robust safety measures, transparent guidelines, and continuous engagement with stakeholders to ensure responsible development and deployment of this technology.

\section*{Acknowledgements}

This work was supported in part by the National Natural Science Foundation of China under Grant 624B1026, Grant 62125603, Grant 62336004, and Grant 62321005.

\nocite{langley00}

\bibliography{reference}

\begin{thebibliography}{47}
\providecommand{\natexlab}[1]{#1}
\providecommand{\url}[1]{\texttt{#1}}
\expandafter\ifx\csname urlstyle\endcsname\relax
  \providecommand{\doi}[1]{doi: #1}\else
  \providecommand{\doi}{doi: \begingroup \urlstyle{rm}\Url}\fi

\bibitem[Albergo \& Vanden-Eijnden(2023)Albergo and Vanden-Eijnden]{albergo2023buildingnormalizingflowsstochastic}
Albergo, M.~S. and Vanden-Eijnden, E.
\newblock Building normalizing flows with stochastic interpolants, 2023.
\newblock URL \url{https://arxiv.org/abs/2209.15571}.

\bibitem[Balaji et~al.(2022)Balaji, Nah, Huang, Vahdat, Song, Zhang, Kreis, Aittala, Aila, Laine, Catanzaro, Karras, and Liu]{balaji2022eDiff-I}
Balaji, Y., Nah, S., Huang, X., Vahdat, A., Song, J., Zhang, Q., Kreis, K., Aittala, M., Aila, T., Laine, S., Catanzaro, B., Karras, T., and Liu, M.-Y.
\newblock ediff-i: Text-to-image diffusion models with ensemble of expert denoisers.
\newblock \emph{arXiv preprint arXiv:2211.01324}, 2022.

\bibitem[Bao et~al.(2023)Bao, Nie, Xue, Cao, Li, Su, and Zhu]{bao2023worthwordsvitbackbone}
Bao, F., Nie, S., Xue, K., Cao, Y., Li, C., Su, H., and Zhu, J.
\newblock All are worth words: A vit backbone for diffusion models, 2023.
\newblock URL \url{https://arxiv.org/abs/2209.12152}.

\bibitem[Cai et~al.(2024)Cai, Jiang, Wang, Tang, Kim, and Huang]{cai2024surveymixtureexperts}
Cai, W., Jiang, J., Wang, F., Tang, J., Kim, S., and Huang, J.
\newblock A survey on mixture of experts, 2024.
\newblock URL \url{https://arxiv.org/abs/2407.06204}.

\bibitem[Chang et~al.(2022)Chang, Zhang, Jiang, Liu, and Freeman]{chang2022maskgit}
Chang, H., Zhang, H., Jiang, L., Liu, C., and Freeman, W.~T.
\newblock Maskgit: Masked generative image transformer.
\newblock In \emph{Proceedings of the IEEE/CVF Conference on Computer Vision and Pattern Recognition}, pp.\  11315--11325, 2022.

\bibitem[Chen et~al.(2018)Chen, Rubanova, Bettencourt, and Duvenaud]{chen2018neural}
Chen, R.~T., Rubanova, Y., Bettencourt, J., and Duvenaud, D.~K.
\newblock Neural ordinary differential equations.
\newblock \emph{Advances in neural information processing systems}, 31, 2018.

\bibitem[Dai et~al.(2024)Dai, Deng, Zhao, Xu, Gao, Chen, Li, Zeng, Yu, Wu, Xie, Li, Huang, Luo, Ruan, Sui, and Liang]{dai2024deepseekmoeultimateexpertspecialization}
Dai, D., Deng, C., Zhao, C., Xu, R.~X., Gao, H., Chen, D., Li, J., Zeng, W., Yu, X., Wu, Y., Xie, Z., Li, Y.~K., Huang, P., Luo, F., Ruan, C., Sui, Z., and Liang, W.
\newblock Deepseekmoe: Towards ultimate expert specialization in mixture-of-experts language models, 2024.
\newblock URL \url{https://arxiv.org/abs/2401.06066}.

\bibitem[DeepSeek-AI et~al.(2024)DeepSeek-AI, Liu, Feng, Xue, Wang, Wu, Lu, Zhao, Deng, Zhang, Ruan, Dai, Guo, et~al.]{deepseekai2024deepseekv3technicalreport}
DeepSeek-AI, Liu, A., Feng, B., Xue, B., Wang, B., Wu, B., Lu, C., Zhao, C., Deng, C., Zhang, C., Ruan, C., Dai, D., Guo, D., et~al.
\newblock Deepseek-v3 technical report, 2024.
\newblock URL \url{https://arxiv.org/abs/2412.19437}.

\bibitem[Dhariwal \& Nichol(2021)Dhariwal and Nichol]{dhariwal2021diffusion}
Dhariwal, P. and Nichol, A.
\newblock Diffusion models beat gans on image synthesis.
\newblock \emph{NeurIPS}, 34:\penalty0 8780--8794, 2021.

\bibitem[Esser et~al.(2024{\natexlab{a}})Esser, Kulal, Blattmann, Entezari, M{\"u}ller, Saini, Levi, Lorenz, Sauer, Boesel, et~al.]{esser2024scaling}
Esser, P., Kulal, S., Blattmann, A., Entezari, R., M{\"u}ller, J., Saini, H., Levi, Y., Lorenz, D., Sauer, A., Boesel, F., et~al.
\newblock Scaling rectified flow transformers for high-resolution image synthesis.
\newblock \emph{arXiv preprint arXiv:2403.03206}, 2024{\natexlab{a}}.

\bibitem[Esser et~al.(2024{\natexlab{b}})Esser, Kulal, Blattmann, Entezari, Müller, Saini, Levi, Lorenz, Sauer, Boesel, Podell, Dockhorn, English, Lacey, Goodwin, Marek, and Rombach]{esser2024scalingrectifiedflowtransformers}
Esser, P., Kulal, S., Blattmann, A., Entezari, R., Müller, J., Saini, H., Levi, Y., Lorenz, D., Sauer, A., Boesel, F., Podell, D., Dockhorn, T., English, Z., Lacey, K., Goodwin, A., Marek, Y., and Rombach, R.
\newblock Scaling rectified flow transformers for high-resolution image synthesis, 2024{\natexlab{b}}.
\newblock URL \url{https://arxiv.org/abs/2403.03206}.

\bibitem[Fei et~al.(2024)Fei, Fan, Yu, Li, and Huang]{FeiDiTMoE2024}
Fei, Z., Fan, M., Yu, C., Li, D., and Huang, J.
\newblock Scaling diffusion transformers to 16 billion parameters.
\newblock \emph{arXiv preprint}, 2024.

\bibitem[Feng et~al.(2023)Feng, Zhang, Yu, Fang, Li, Chen, Lu, Liu, Yin, Feng, Sun, Chen, Tian, Wu, and Wang]{feng2023ernievilg20improvingtexttoimage}
Feng, Z., Zhang, Z., Yu, X., Fang, Y., Li, L., Chen, X., Lu, Y., Liu, J., Yin, W., Feng, S., Sun, Y., Chen, L., Tian, H., Wu, H., and Wang, H.
\newblock Ernie-vilg 2.0: Improving text-to-image diffusion model with knowledge-enhanced mixture-of-denoising-experts, 2023.
\newblock URL \url{https://arxiv.org/abs/2210.15257}.

\bibitem[Ghosh et~al.(2023)Ghosh, Hajishirzi, and Schmidt]{ghosh2023genevalobjectfocusedframeworkevaluating}
Ghosh, D., Hajishirzi, H., and Schmidt, L.
\newblock Geneval: An object-focused framework for evaluating text-to-image alignment, 2023.
\newblock URL \url{https://arxiv.org/abs/2310.11513}.

\bibitem[Heusel et~al.(2017)Heusel, Ramsauer, Unterthiner, Nessler, and Hochreiter]{fid}
Heusel, M., Ramsauer, H., Unterthiner, T., Nessler, B., and Hochreiter, S.
\newblock Gans trained by a two time-scale update rule converge to a local nash equilibrium.
\newblock In Guyon, I., Luxburg, U.~V., Bengio, S., Wallach, H., Fergus, R., Vishwanathan, S., and Garnett, R. (eds.), \emph{Advances in Neural Information Processing Systems}, volume~30. Curran Associates, Inc., 2017.
\newblock URL \url{https://proceedings.neurips.cc/paper_files/paper/2017/file/8a1d694707eb0fefe65871369074926d-Paper.pdf}.

\bibitem[Ho \& Salimans(2022)Ho and Salimans]{ho2022classifierfreediffusionguidance}
Ho, J. and Salimans, T.
\newblock Classifier-free diffusion guidance, 2022.
\newblock URL \url{https://arxiv.org/abs/2207.12598}.

\bibitem[Ho et~al.(2020)Ho, Jain, and Abbeel]{ho2020denoising}
Ho, J., Jain, A., and Abbeel, P.
\newblock Denoising diffusion probabilistic models.
\newblock \emph{NeurIPS}, 33:\penalty0 6840--6851, 2020.

\bibitem[Kingma et~al.(2021)Kingma, Salimans, Poole, and Ho]{kingma2021variational}
Kingma, D., Salimans, T., Poole, B., and Ho, J.
\newblock Variational diffusion models.
\newblock \emph{NeurIPS}, 34:\penalty0 21696--21707, 2021.

\bibitem[Lee et~al.(2023)Lee, Kim, Go, Jeong, Oh, and Choi]{lee2023multiarchitecturemultiexpertdiffusionmodels}
Lee, Y., Kim, J.-Y., Go, H., Jeong, M., Oh, S., and Choi, S.
\newblock Multi-architecture multi-expert diffusion models, 2023.
\newblock URL \url{https://arxiv.org/abs/2306.04990}.

\bibitem[Lepikhin et~al.(2020)Lepikhin, Lee, Xu, Chen, Firat, Huang, Krikun, Shazeer, and Chen]{lepikhin2020gshardscalinggiantmodels}
Lepikhin, D., Lee, H., Xu, Y., Chen, D., Firat, O., Huang, Y., Krikun, M., Shazeer, N., and Chen, Z.
\newblock Gshard: Scaling giant models with conditional computation and automatic sharding, 2020.
\newblock URL \url{https://arxiv.org/abs/2006.16668}.

\bibitem[Lipman et~al.(2022)Lipman, Chen, Ben-Hamu, Nickel, and Le]{lipman2022flow}
Lipman, Y., Chen, R.~T., Ben-Hamu, H., Nickel, M., and Le, M.
\newblock Flow matching for generative modeling.
\newblock \emph{arXiv preprint arXiv:2210.02747}, 2022.

\bibitem[Lipman et~al.(2023)Lipman, Chen, Ben-Hamu, Nickel, and Le]{lipman2023flowmatchinggenerativemodeling}
Lipman, Y., Chen, R. T.~Q., Ben-Hamu, H., Nickel, M., and Le, M.
\newblock Flow matching for generative modeling, 2023.
\newblock URL \url{https://arxiv.org/abs/2210.02747}.

\bibitem[Liu et~al.(2022{\natexlab{a}})Liu, Gong, and Liu]{liu2022flow}
Liu, X., Gong, C., and Liu, Q.
\newblock Flow straight and fast: Learning to generate and transfer data with rectified flow.
\newblock \emph{arXiv preprint arXiv:2209.03003}, 2022{\natexlab{a}}.

\bibitem[Liu et~al.(2022{\natexlab{b}})Liu, Gong, and Liu]{liu2022flowstraightfastlearning}
Liu, X., Gong, C., and Liu, Q.
\newblock Flow straight and fast: Learning to generate and transfer data with rectified flow, 2022{\natexlab{b}}.
\newblock URL \url{https://arxiv.org/abs/2209.03003}.

\bibitem[Liu et~al.(2023)Liu, Zhang, Ma, Peng, et~al.]{liu2023instaflow}
Liu, X., Zhang, X., Ma, J., Peng, J., et~al.
\newblock Instaflow: One step is enough for high-quality diffusion-based text-to-image generation.
\newblock In \emph{The Twelfth International Conference on Learning Representations}, 2023.

\bibitem[Loshchilov \& Hutter(2017)Loshchilov and Hutter]{adamw}
Loshchilov, I. and Hutter, F.
\newblock Decoupled weight decay regularization.
\newblock \emph{arXiv preprint arXiv:1711.05101}, 2017.

\bibitem[Ma et~al.(2024)Ma, Goldstein, Albergo, Boffi, Vanden-Eijnden, and Xie]{ma2024sit}
Ma, N., Goldstein, M., Albergo, M.~S., Boffi, N.~M., Vanden-Eijnden, E., and Xie, S.
\newblock Sit: Exploring flow and diffusion-based generative models with scalable interpolant transformers.
\newblock \emph{arXiv preprint arXiv:2401.08740}, 2024.

\bibitem[MiniMax et~al.(2025)MiniMax, Li, Gong, Yang, Shan, Liu, Zhu, Zhang, Guo, Chen, Li, Jiao, Li, Zhang, Sun, Dong, Zhu, Zhuang, Song, Zhu, Han, Li, Xie, Xu, Yan, Zhang, Xiao, Kang, et~al.]{minimax2025minimax01scalingfoundationmodels}
MiniMax, Li, A., Gong, B., Yang, B., Shan, B., Liu, C., Zhu, C., Zhang, C., Guo, C., Chen, D., Li, D., Jiao, E., Li, G., Zhang, G., Sun, H., Dong, H., Zhu, J., Zhuang, J., Song, J., Zhu, J., Han, J., Li, J., Xie, J., Xu, J., Yan, J., Zhang, K., Xiao, K., Kang, K., et~al.
\newblock Minimax-01: Scaling foundation models with lightning attention, 2025.
\newblock URL \url{https://arxiv.org/abs/2501.08313}.

\bibitem[Muennighoff et~al.(2024)Muennighoff, Soldaini, Groeneveld, Lo, Morrison, Min, Shi, Walsh, Tafjord, Lambert, Gu, Arora, Bhagia, Schwenk, Wadden, Wettig, Hui, Dettmers, Kiela, Farhadi, Smith, Koh, Singh, and Hajishirzi]{muennighoff2024olmoeopenmixtureofexpertslanguage}
Muennighoff, N., Soldaini, L., Groeneveld, D., Lo, K., Morrison, J., Min, S., Shi, W., Walsh, P., Tafjord, O., Lambert, N., Gu, Y., Arora, S., Bhagia, A., Schwenk, D., Wadden, D., Wettig, A., Hui, B., Dettmers, T., Kiela, D., Farhadi, A., Smith, N.~A., Koh, P.~W., Singh, A., and Hajishirzi, H.
\newblock Olmoe: Open mixture-of-experts language models, 2024.
\newblock URL \url{https://arxiv.org/abs/2409.02060}.

\bibitem[Pan et~al.(2023)Pan, Sun, Ge, Li, Duan, Wu, Zhang, Zhou, Qin, Wang, Dai, Qiao, and Li]{pan2023journeydb}
Pan, J., Sun, K., Ge, Y., Li, H., Duan, H., Wu, X., Zhang, R., Zhou, A., Qin, Z., Wang, Y., Dai, J., Qiao, Y., and Li, H.
\newblock Journeydb: A benchmark for generative image understanding, 2023.

\bibitem[Peebles \& Xie(2023{\natexlab{a}})Peebles and Xie]{peebles2023scalable}
Peebles, W. and Xie, S.
\newblock Scalable diffusion models with transformers.
\newblock In \emph{Proceedings of the IEEE/CVF International Conference on Computer Vision}, pp.\  4195--4205, 2023{\natexlab{a}}.

\bibitem[Peebles \& Xie(2023{\natexlab{b}})Peebles and Xie]{peebles2023scalablediffusionmodelstransformers}
Peebles, W. and Xie, S.
\newblock Scalable diffusion models with transformers, 2023{\natexlab{b}}.
\newblock URL \url{https://arxiv.org/abs/2212.09748}.

\bibitem[Podell et~al.(2023)Podell, English, Lacey, Blattmann, Dockhorn, M{\"u}ller, Penna, and Rombach]{podell2023sdxl}
Podell, D., English, Z., Lacey, K., Blattmann, A., Dockhorn, T., M{\"u}ller, J., Penna, J., and Rombach, R.
\newblock Sdxl: Improving latent diffusion models for high-resolution image synthesis.
\newblock \emph{arXiv preprint arXiv:2307.01952}, 2023.

\bibitem[Qiu et~al.(2025)Qiu, Huang, Zheng, Wen, Wang, Men, Titov, Liu, Zhou, and Lin]{qiu2025demons}
Qiu, Z., Huang, Z., Zheng, B., Wen, K., Wang, Z., Men, R., Titov, I., Liu, D., Zhou, J., and Lin, J.
\newblock Demons in the detail: On implementing load balancing loss for training specialized mixture-of-expert models.
\newblock \emph{arXiv preprint arXiv:2501.11873}, 2025.

\bibitem[Rombach et~al.(2022)Rombach, Blattmann, Lorenz, Esser, and Ommer]{rombach2022high}
Rombach, R., Blattmann, A., Lorenz, D., Esser, P., and Ommer, B.
\newblock High-resolution image synthesis with latent diffusion models.
\newblock In \emph{CVPR}, pp.\  10684--10695, 2022.

\bibitem[Russakovsky et~al.(2015)Russakovsky, Deng, Su, Krause, Satheesh, Ma, Huang, Karpathy, Khosla, Bernstein, Berg, and Fei-Fei]{ILSVRC15}
Russakovsky, O., Deng, J., Su, H., Krause, J., Satheesh, S., Ma, S., Huang, Z., Karpathy, A., Khosla, A., Bernstein, M., Berg, A.~C., and Fei-Fei, L.
\newblock {ImageNet Large Scale Visual Recognition Challenge}.
\newblock \emph{International Journal of Computer Vision (IJCV)}, 115\penalty0 (3):\penalty0 211--252, 2015.
\newblock \doi{10.1007/s11263-015-0816-y}.

\bibitem[Sauer et~al.(2022)Sauer, Schwarz, and Geiger]{sauer2022stylegan}
Sauer, A., Schwarz, K., and Geiger, A.
\newblock Stylegan-xl: Scaling stylegan to large diverse datasets.
\newblock In \emph{ACM SIGGRAPH 2022 conference proceedings}, pp.\  1--10, 2022.

\bibitem[Sehwag et~al.(2024)Sehwag, Kong, Li, Spranger, and Lyu]{sehwag2024stretchingdollardiffusiontraining}
Sehwag, V., Kong, X., Li, J., Spranger, M., and Lyu, L.
\newblock Stretching each dollar: Diffusion training from scratch on a micro-budget, 2024.
\newblock URL \url{https://arxiv.org/abs/2407.15811}.

\bibitem[Shazeer et~al.(2017)Shazeer, Mirhoseini, Maziarz, Davis, Le, Hinton, and Dean]{shazeer2017outrageouslylargeneuralnetworks}
Shazeer, N., Mirhoseini, A., Maziarz, K., Davis, A., Le, Q., Hinton, G., and Dean, J.
\newblock Outrageously large neural networks: The sparsely-gated mixture-of-experts layer, 2017.
\newblock URL \url{https://arxiv.org/abs/1701.06538}.

\bibitem[Sohl-Dickstein et~al.(2015)Sohl-Dickstein, Weiss, Maheswaranathan, and Ganguli]{sohl2015deep}
Sohl-Dickstein, J., Weiss, E., Maheswaranathan, N., and Ganguli, S.
\newblock Deep unsupervised learning using nonequilibrium thermodynamics.
\newblock In \emph{ICML}, pp.\  2256--2265. PMLR, 2015.

\bibitem[Song et~al.(2021)Song, Sohl-Dickstein, Kingma, Kumar, Ermon, and Poole]{song2021score}
Song, Y., Sohl-Dickstein, J., Kingma, D.~P., Kumar, A., Ermon, S., and Poole, B.
\newblock Score-based generative modeling through stochastic differential equations.
\newblock In \emph{ICLR}, 2021.

\bibitem[Sun et~al.(2024{\natexlab{a}})Sun, Lei, Zhang, Li, Huang, Pang, Dai, and Du]{sun2024ecditscalingdiffusiontransformers}
Sun, H., Lei, T., Zhang, B., Li, Y., Huang, H., Pang, R., Dai, B., and Du, N.
\newblock Ec-dit: Scaling diffusion transformers with adaptive expert-choice routing, 2024{\natexlab{a}}.
\newblock URL \url{https://arxiv.org/abs/2410.02098}.

\bibitem[Sun et~al.(2024{\natexlab{b}})Sun, Jiang, Chen, Zhang, Peng, Luo, and Yuan]{sun2024autoregressivemodelbeatsdiffusion}
Sun, P., Jiang, Y., Chen, S., Zhang, S., Peng, B., Luo, P., and Yuan, Z.
\newblock Autoregressive model beats diffusion: Llama for scalable image generation, 2024{\natexlab{b}}.
\newblock URL \url{https://arxiv.org/abs/2406.06525}.

\bibitem[Tian et~al.(2024)Tian, Jiang, Yuan, Peng, and Wang]{VAR}
Tian, K., Jiang, Y., Yuan, Z., Peng, B., and Wang, L.
\newblock Visual autoregressive modeling: Scalable image generation via next-scale prediction.
\newblock 2024.

\bibitem[Yang et~al.(2024)Yang, Qi, Gu, Wang, Gao, and Xu]{yang2024xmoesparsemodelsfinegrained}
Yang, Y., Qi, S., Gu, W., Wang, C., Gao, C., and Xu, Z.
\newblock Xmoe: Sparse models with fine-grained and adaptive expert selection, 2024.
\newblock URL \url{https://arxiv.org/abs/2403.18926}.

\bibitem[Yatharth~Gupta(2024)]{segmoe}
Yatharth~Gupta, Vishnu V~Jaddipal, H.~P.
\newblock Segmoe: Segmind mixture of diffusion experts.
\newblock \url{https://github.com/segmind/segmoe}, 2024.

\bibitem[Zhou et~al.(2022)Zhou, Lei, Liu, Du, Huang, Zhao, Dai, Chen, Le, and Laudon]{zhou2022mixtureofexpertsexpertchoicerouting}
Zhou, Y., Lei, T., Liu, H., Du, N., Huang, Y., Zhao, V., Dai, A., Chen, Z., Le, Q., and Laudon, J.
\newblock Mixture-of-experts with expert choice routing, 2022.
\newblock URL \url{https://arxiv.org/abs/2202.09368}.

\end{thebibliography}
\bibliographystyle{icml2025}

\newpage
\appendix
\onecolumn

\section{Generative Modeling.}
In this section, we will provide a detailed bachground of generative modeling of both DDPM~\cite{ho2020denoising,song2021score,rombach2022high} and Rectified Flow~\cite{lipman2023flowmatchinggenerativemodeling,ma2024sit,esser2024scalingrectifiedflowtransformers} which is helpful to understand the difference and relationship between them.

Generative modeling essentially defines a mapping between $\bfx_1$ from a noise distribution $p_1(\bfx)$ to $\bfx_0$ from a data distribution $p_0(\bfx)$ leads to time-dependent processes represented as below

\begin{equation}
    \bfx_t = \alpha_t \bfx_0 + \sigma_t\bfeps, t\in [0,1]
    \label{eq:xt},
\end{equation}

where $\alpha_t$ is a decreasing function of $t$ and $\sigma_t$ is an increasing function of $t$. We set $\alpha_0=1,\sigma_0=0$ and  $\alpha_1=0,\sigma_0=1$ to make the marginals $p_t(\bfx_t)=\mathbb{E}_{\bfeps\sim\mathcal{N}(0,\rm I)}p_t(\bfx_t | \bfeps)$ are consistent with data $p_0(\bfx)$ and noise $p_1(\bfx)$ distirbution.  $p_1(\bfx)$ usually be chosen as gaussion distribution $\mathcal{N}(0,1)$.

Different forward path from data to noise leads to different training object which significantly affect the performance of the model. Next we will introduce DDPM and Rectified Flow.

\subsection{Denosing Diffusion Probabilistic Models (DDPM)}
In DDPM the choice for $\alpha_t$ and $\sigma_t$ is referred to as the noise schedule and the signal-to-noise-ratio (SNR) $\alpha_t^2/\sigma_t^2$ is strictly decreasing w.r.t $t$ ~\cite{kingma2021variational}. Moreover, ~\cite{kingma2021variational}
 prove that the following stochastic differential Eq. (SDE) has same transition distribution as $p_t(\bfx_t|\bfx_0)$ for any $t\in [0,1]$:
 \begin{equation}
     \mathrm{d} \bfx_t = f(t) \bfx_t \mathrm{d} t + g(t) \mathrm{d} \textbf{w}_t, t\in[0,1], \bfx_0\sim p_0(\bfx_0),
\label{eq:sde_diffusion}
 \end{equation}
where $\textbf{w}_t\in \mathbb{E}^D$ is the standard Wiener process, and

\begin{equation}
    f(t)=\frac{\mathrm{d}\log{\alpha_t}}{\mathrm{d}t},\quad g^2(t)=\frac{\mathrm{d}\sigma^2_t}{\mathrm{d}t}-2\frac{\mathrm{d}\log{\alpha_t}}{\mathrm{d}t} \sigma_t^2.
\end{equation}

~\cite{song2021score} proved that the forward path in Eq. ~\ref{eq:sde_diffusion} has an equivalent reverse process from time $1$ to $0$ under some regularity conditions, starting with $p_T(\bfx_T)$

\begin{equation}
     \mathrm{d} \bfx_t = [f(t) \bfx_t - g(t)^2\nabla_\bfx \log{p_t(\bfx_t)}]\mathrm{d} t + g(t) \mathrm{d}\bar{\textbf{w}}_t, \quad \bfx_T \sim p_T(\bfx_T),
     \label{eq:reverse_sde}
\end{equation}
where $\bar{\textbf{w}}_t\in \mathbb{E}^D$ is the standard Wiener process. We can esitimate the score term $\nabla_\bfx \log{p_t(\bfx_t)}$ as each time $t$ to iterativtely solve the reverse process, then get the gernerated target. DPMs train a neural network $\bfeps_\theta(\bfx, t)$ parameterized by $\theta$ to esitimated the scaled score function $-\sigma\nabla_\bfx \log{p_t(\bfx_t)}$. To optimize $\bfeps_\theta$, we minimize the following objective~\cite{ho2020denoising, song2021score, ma2024sit}

\begin{align}
\mathcal{L}_{\rm DDPM}(\theta) &=
    \mathbb{E}_{t,p_0(\bfx_0),p(\bfx_t|\bfx_0)}
    \left[
    \lambda(t)
    \lVert
    \bfeps_\theta(\bfx_t, t) + \sigma_t\nabla_\bfx \log p_t(\bfx_t)
    \rVert_2^2
    \right],%
    \label{loss:eps}
\end{align}

where $\lambda_t$ is a time-dependent coefficent. $\bfeps_\theta(\bfx_t,t)$ can be interpreted as predicting the Gaussian noise added to $\bfx_t$, thus it is commonly referred to as a noise prediction model. Consequently, the diffusion model is known as a denoising diffusion probabilistic model. Substitute score term in \eqref{eq:reverse_sde} with $-\bfeps_\theta(\bfx_t,t)/\sigma_t$, we can solve the reverse process and generate samples from DPMs with numerical solvers. To further accelerate the sampling process, Song \etal ~\cite{song2021score} proved that the equvivalent probability flow ODE is 

\begin{equation}
    \frac{\dif \bfx_t}{\dif t} = \bfv_\theta(\bfx_t, t) :=
    f(t)\bfx_t + \frac{g^2(t)}{2\sigma_t}\bfeps_\theta(\bfx_t, t), \quad
    \bfx_1 \sim \mathcal{N}(0, \mathbf{I}).
    \label{equ:diffusion_ode}
\end{equation}

Thus samples can be also generated by solving the ODE from time $1$ to $0$.

\subsection{Rectified Flow Models (Flow)}
Recified flow models~\cite{liu2022flowstraightfastlearning,albergo2023buildingnormalizingflowsstochastic,lipman2023flowmatchinggenerativemodeling} connects data $\bfx_0$ and noise $\bfeps$ on a straight line as follows

\begin{equation}
    \bfx_t = (1-t)\bfx_0 + t\bfeps,\quad t\in[0,1].
    \label{eq:flow_path}
\end{equation}
To precisely express the relationship between $\bfx_t$, $\bfx_0$, and $\bfeps$, we first construct a time-dependent vector field $u: [0,1] \times \mathbb{R}^D \rightarrow \mathbb{R}^D$. This vector field $u_t$ can be used to construct a time-dependent diffeomorphic map, known as a flow $\phi: [0,1] \times \mathbb{R}^D \rightarrow \mathbb{R}^D$, through the following ODE:

\begin{align}
    \frac{\dif}{\dif t}\phi_t(\bfx_0) &= u_t(\phi_t(\bfx_0)) \\
    \phi_0(\bfx_0) &= \bfx_0.
\end{align}
The vector field $u_t$ can be modeled as a neural network $\bfv_\theta$ ~\cite{chen2018neural} which leads to a deep parametric model of the flow $\phi_t$, called a \textit{Continuous Normalizing Flow} (CNF). We can using \textit{conditional flow matching} (CFM) technique~\cite{lipman2022flow} to training a CNF. 
Now we can define the flow we need as follows:

\begin{equation}
    \psi_t(\cdot |\bfeps): \bfx_0 \mapsto \alpha_t\bfx_0 + \sigma_t \bfeps .
    \label{eq:generative_flow}
\end{equation}
The corresponding velocity vector field of the flow $\psi_t$ can be represented as:

\begin{equation}
    u_t(\psi_t(\cdot |\bfeps)|\bfeps) = \frac{\dif}{\dif t} \psi_t(\bfx_0 |\bfeps) =  \Dot{\alpha_t}\bfx_0 + \Dot{\sigma_t} \bfeps = \bfeps - \bfx_0 .
    \label{eq:v_field_of_flow}
\end{equation}

Using conditional flow matching technique, $\bfv(\bfx_t, t)$ in Eq.~\eqref{equ:diffusion_ode} can be modeled as a neural network $\bfv_\theta(\bfx_t, t)$ by minimziing the following objective

\begin{align}
\mathcal{L}_{\rm Flow}(\theta) &=
    \mathbb{E}_{t,p_0(\bfx_0),p_1(\bfeps)}
    \lVert
    \bfv_\theta(\bfx_t, t) - \frac{\dif}{\dif t} \psi_t(\bfx_0 | \bfeps)
    \rVert_2^2 \\
   &= \mathbb{E}_{t,p_0(\bfx_0),p(\bfeps)}
    \lVert
    \bfv_\theta(\bfx_t, t) - (\Dot{\alpha_t}\bfx_0 + \Dot{\sigma_t}\bfeps)
    \rVert_2^2 \\
   &= \mathbb{E}_{t,p_0(\bfx_0),p(\bfeps)}
    \lVert
    \bfv_\theta(\bfx_t, t) - (\bfeps - \bfx_0)
    \rVert_2^2.
    \label{loss:v}
\end{align}

Samples can be generated by sovling the probability flow ODE below with learned velocity using numerical sovler like Euler, Heun, Runge-Kutta method. 
\begin{equation}
    \dif \bfx_t = \bfv_\theta(\bfx_t, t)\dif t,\quad \bfx_1 =\bfeps \sim \mathcal{N}(0,1).
\end{equation}

\subsection{Relationship DDPM and Flow}
There exists a straightforward connection between $\bfv_\theta(\bfx_t,t)$ and the score term $-\sigma_t\nabla_\bfx\log{p_t(\bfx_t)}$ can be derived as follows

\begin{align}
    \bfv_\theta(\bfx_t, t) &= f(t)\bfx_t + \frac{g^2(t)}{2\sigma_t}\bfeps_\theta(\bfx_t, t) \\
    &\approx \frac{\Dot{\alpha}_t}{\alpha_t}\bfx_t + \left(\Dot{\sigma}_t - \frac{\Dot{\alpha}_t}{\alpha_t}\right)(-\sigma_t\nabla_\bfx \log p_t(\bfx_t)).
    \label{eq:connection_v_eps}
\end{align}
Let $\zeta_t = \Dot{\sigma}_t - \frac{\Dot{\alpha}_t}{\alpha_t}$, and we have $\bfeps_\theta(\bfx_t, t) \approx -\sigma_t\nabla_\bfx \log p_t(\bfx_t$, then we can get $ \bfv_\theta(\bfx_t,t)=\frac{\Dot{\alpha}_t}{\alpha_t}\bfx_t + \zeta_t\bfeps_\theta(\bfx_t, t)$

By plugging ~\eqref{eq:connection_v_eps} into the loss $\mathcal{L}_{\rm Flow}$ in Eq. ~\eqref{loss:v} we have:

\begin{align}
\mathcal{L}_{\rm Flow}(\theta) &=
    \mathbb{E}_{t,p_0(\bfx_0),p_1(\bfeps)}
    \lVert
    \bfv_\theta(\bfx_t, t) - (\Dot{\alpha_t}\bfx_0 + \Dot{\sigma_t}\bfeps)
    \rVert_2^2 \\
   &= \mathbb{E}_{t,p_0(\bfx_0),p(\bfeps)}
    \lVert
   \frac{\Dot{\alpha}_t}{\alpha_t}\bfx_t + \zeta_t\bfeps_\theta(\bfx_t, t) - \Dot{\sigma_t}\bfeps
    \rVert_2^2 \\
    &= \mathbb{E}_{t,p_0(\bfx_0),p(\bfeps)}
    \lVert
   \zeta_t\bfeps_\theta(\bfx_t, t) - \zeta_t\bfeps
    \rVert_2^2 \\
    &= \mathbb{E}_{t,p_0(\bfx_0),p(\bfeps)}
    \left[\zeta_t^2 \lVert
   \bfeps_\theta(\bfx_t, t) - \bfeps
    \rVert_2^2\right].
    \label{loss:v_reparameter}
\end{align}

Considering Eq. ~\eqref{eq:xt}, we have $\bfx_t \sim \mathcal{N}(\alpha_t \bfx_0, \sigma_t\mathbf{I})$ and 
$\nabla_\bfx \log p(\bfx_t) = \sigma_t^{-1} (\bfx_t - \alpha_t \bfx_0) = \sigma_t^{-1} (\sigma_t \bfeps)=\bfeps $. Then, we can get the equivilant loss function as below:

\begin{equation}
    \mathcal{L}_{\rm Flow}(\theta) =
    \mathbb{E}_{t,p_0(x_0),p(\bfx_t|\bfx_0)}
    \left[
    \zeta_t^2 \left\lVert
    \bfeps_\theta(\bfx_t, t) + \sigma_t \nabla_\bfx \log p_t(\bfx_t)
    \right\rVert_2^2
    \right].
\end{equation}

Recall that 
$
\mathcal{L}_{\rm DDPM}(\theta) =
    \mathbb{E}_{t,p_0(\bfx_0),p(\bfx_t|\bfx_0)}
    \left[
    \lambda(t)
    \lVert
    \bfeps_\theta(\bfx_t, t) + \sigma_t\nabla_\bfx \log p_t(\bfx_t)
    \rVert_2^2
    \right].
$

We can find that $\mathcal{L}_{\rm DDPM}$ and $\mathcal{L}_{\rm Flow}$ have the same form, with the only difference being their time-dependent weighting functions, which lead to different trajectories and properties.

\begin{table*}[t]
  \centering
    \caption{\textbf{Comprehensive State-of-the-art Comparison}. Evaluation on ImageNet $256\times 256$ class-conditional generation. DiffMoE achieves better FID with fewer parameters. -G/-U denotes with/without guidance~\cite{ho2022classifierfreediffusionguidance}. $^\dagger$: results from~\cite{ma2024sit} (DDPM) and~\cite{peebles2023scalable} (Flow). $^*$: our reproduction.  \textbf{Bold} indicates best performance in each cell.}
  \adjustbox{width=\linewidth}
  {

    \begin{tabular}{lccccc}\toprule
    \multirow{1}{*}{Diffusion Models} & \multirow{1}{*}{\# Avg. Activated Params.} & \multirow{1}{*}{FID$\downarrow$} & \multirow{1}{*}{IS$\uparrow$} & \multirow{1}{*}{Precision$\uparrow$} & \multirow{1}{*}{Recall$\uparrow$} \\\midrule
    \textcolor{gray}{\textit{GAN}} &     &   &    &    &    \\
    
    StyleGAN-XL~\cite{sauer2022stylegan} & 166M    & 2.30  & 265.1  & 0.78  & 0.53  \\\midrule

    \textcolor{gray}{\textit{Masked Modeling}} \\
    Mask-GIT~\cite{chang2022maskgit} & 227M     & 6.18  & 182.1  & 0.80  & 0.51  \\\midrule

    \textcolor{gray}{\textit{Autoregressive Model}} \\
    LlamaGen-3B~\cite{sun2024autoregressivemodelbeatsdiffusion}   & 3000M  & 10.94  & 101.0  & 0.69  & 0.63  \\
    VAR-$d20$~\cite{VAR}   & 600M  & 2.57  & 302.6  & 0.83  & 0.56  \\
    
    \midrule

    \textcolor{gray}{\textit{Diffusion Model}} \\

    ADM-U ~\cite{dhariwal2021diffusion}   & 554M  & 10.94  & 101.0  & 0.69  & 0.63  \\
    ADM-G~\cite{dhariwal2021diffusion} & 608M    & 4.59  & 186.7  & 0.83  & 0.53  \\
    LDM-4-G~\cite{rombach2022high}  & 400M    & 3.95  & 247.7  & 0.87  & 0.48  \\
    U-ViT-H/2-G~\cite{bao2023worthwordsvitbackbone}  & 501M    & 2.29  & 263.88  & 0.82  & 0.57  \\\midrule

    \textcolor{gray}{Dense-DiT-XL-Flow-U$^\dagger$~\cite{peebles2023scalable}} & \textcolor{gray}{675M}  & \textcolor{gray}{9.35}  & \textcolor{gray}{126.06}  & \textcolor{gray}{0.67}  & \textcolor{gray}{0.68}  \\
    \textcolor{gray}{Dense-DiT-XL-Flow-G$^\dagger$ (cfg=1.5, ODE)~\cite{ma2024sit}} & \textcolor{gray}{675M}  & \textcolor{gray}{2.15}    & \textcolor{gray}{254.9}  & \textcolor{gray}{0.81}  & \textcolor{gray}{0.60}  \\
    Dense-DiT-XL-Flow-U$^*$~\cite{peebles2023scalable} & 675M  & \textbf{9.47}  & 115.58  &  0.67  & 0.67  \\
    Dense-DiT-XL-Flow-G$^*$ (cfg=1.5, ODE)~\cite{ma2024sit} & 675M  & 2.19    & 272.30  &  0.83  & 0.58 \\

    \rowcolor{Gray} DiffMoE-L-E8-Flow-U & 458M   & \textbf{9.60}  & 131.46  & 0.67  & 0.67  \\
    \rowcolor{Gray} DiffMoE-L-E8-Flow-G (cfg=1.5, ODE) & 458M   & \textbf{2.13}  & 274.39  & 0.81  & 0.60  \\\midrule
     
    \textcolor{gray}{Dense-DiT-XL-DDPM-U$^\dagger$~\cite{peebles2023scalable}} & \textcolor{gray}{675M}  & \textcolor{gray}{9.62}  & \textcolor{gray}{121.50}  & \textcolor{gray}{0.67}  & \textcolor{gray}{0.67}  \\
    \textcolor{gray}{Dense-DiT-XL-DDPM-G$^\dagger$ (cfg=1.5)~\cite{peebles2023scalable}} & \textcolor{gray}{675M}  & \textcolor{gray}{2.27}  & \textcolor{gray}{278.2}  & \textcolor{gray}{0.83}  & \textcolor{gray}{0.57}  \\
    Dense-DiT-XL-DDPM-U$^*$~\cite{peebles2023scalable} & 675M  & 9.62  & 123.19  &  0.66  & 0.68  \\
    Dense-DiT-XL-DDPM-G$^*$ (cfg=1.5)~\cite{peebles2023scalable} & 675M  & 2.32  & 279.18  & 0.83  & 0.57  \\
    \rowcolor{Gray} DiffMoE-L-E8-DDPM-U & 458M   & \textbf{9.17}  & 131.10  & 0.67  & 0.67  \\
    \rowcolor{Gray} DiffMoE-L-E8-DDPM-G (cfg=1.5) & 458M   &     \textbf{2.30}  & 284.78 & 0.82  & 0.59  \\

     \bottomrule
    \end{tabular}%
    }
  \label{tab:comprehensive_sota}%
  \vspace{-10pt}
\end{table*}%

\begin{table*}[t]
  \centering
    \caption{\textbf{Performance Comparison on Text-to-Image Generation Tasks}. Evaluation results on GenEval Benchmark~\cite{ghosh2023genevalobjectfocusedframeworkevaluating} at $256\times 256$ resolution across six different categories: single object generation, two-object composition, object counting, color recognition, spatial positioning, and color attribute understanding. DiffMoE-E16-T2I-Flow demonstrates superior performance over Dense-DiT-T2I-Flow, particularly in object generation and spatial tasks. \textbf{Bold} indicates best overall performance in each cell. \# A.A.P. denotes \# Avg. Act. Params.  \# T.P. denotes \# Total Params.}
  \adjustbox{width=\linewidth}
  {
    \begin{tabular}{lccccccccc}\toprule
    \multirow{1}{*}{Diffusion Models} & 
    \multirow{1}{*}{\# A.A.P.} & 
    \multirow{1}{*}{\# T.P.} & 
    \multirow{1}{*}{Single Obj. $\uparrow$} & 
    \multirow{1}{*}{Two Obj.$\uparrow$} & 
    \multirow{1}{*}{Counting Obj.$\uparrow$} & 
    \multirow{1}{*}{Colors $\uparrow$} & 
    \multirow{1}{*}{Position $\uparrow$} & 
    \multirow{1}{*}{Color Attri. $\uparrow$} & 
    \multirow{1}{*}{Overall $\uparrow$} \\\midrule
    Dense-DiT-T2I-Flow (w/o SFT) & 1.2B & 1.2B & 0.80   & 0.33  & 0.24  & 0.53  & 0.10 & 0.26 & 0.38 \\
    \rowcolor{Gray} DiffMoE-E16-T2I-Flow (w/o SFT) & 1.2B & 4.6B & 0.90   & 0.42  & 0.24  & 0.59  & 0.19 & 0.28 & \textbf{0.44} \\\midrule
    Dense-DiT-T2I-Flow (w SFT) & 1.2B & 1.2B & 0.93   & 0.48  & 0.50  & 0.73  & 0.07 & 0.20 & 0.49 \\
    \rowcolor{Gray} DiffMoE-E16-T2I-Flow (w SFT) & 1.2B & 4.6B & 0.96   & 0.53  & 0.46  & 0.78  & 0.13 & 0.20 & \textbf{0.51} \\\midrule
    
    \bottomrule
    \end{tabular}%
    }
  \label{tab:geneval_t2i}%
  \vspace{-10pt}
\end{table*}%

\begin{table*}[t]
  \centering
     \caption{\textbf{Performance comparison of different diffusion models with varying token interaction strategies}. All models are trained with Flow Matching for 700K steps. The interaction levels (L1/L2/L3) represent: L1 for isolated token processing, L2 for local token routing within samples, and L3 for global token routing across samples. Our DiffMoE-L-Flow with Dynamic Global CP achieves the best FID score of 14.41 while maintaining parameter efficiency and reduced computational cost. \# A.A.P. denotes \# Avg. Act. Params.}
    \begin{tabular}{lclll}
    \toprule
    Model & \# A.A.P. & Training Strategy & Inference Strategy & FID50K $\downarrow$\\
    \midrule
    TC-DiT-L-E16-Flow & 458M & L1: Isolated & L1: Isolated & 19.06 \\
    Dense-DiT-L-Flow & 458M  & L1: Isolated & L1: Isolated & 17.01 \\
    EC-DiT-L-E16-Flow & 458M   & L2: Local & L2: Local Static TopK Routing & 16.12 \\
    EC-DiT-L-E16-Flow & 458M   & L2: Local & L2: Local Dynamic Intra-sample Routing & 23.74 \\
    DiffMoE-L-E16-Flow & 458M & L3: Global & L3: Global Static TopK Routing & 15.25 \\
    \midrule
    Dense-DiT-XL-Flow & 675M   & L1: Isolated & L1: Isolated &  \underline{14.77} \\
    \midrule
    \rowcolor{Gray} DiffMoE-L-E16-Flow & 454M & L3: Global & L3: Global Dynamic Cross-sample Routing & \textbf{14.41} \\
    \bottomrule
    \end{tabular}%
  \label{tab:flow_matching_comparison}%
\end{table*}%

\section{More Implementation Details}

\subsection{Training setup.}
\label{appendix:sec_implementation_details}
We train class-conditional DiffMoE and baseline models at 256x256 image resolution on the ImageNet dataset~\cite{ILSVRC15} a highly-competitive generative benchmark, which contains 1281167 training images. We use horizontal flips as the only data augmentation. We train all models with AdamW ~\cite{adamw}. We use a constant learning rate of $1\times10^{-4}$, no weight decay and a fixed global batch size of 256, following  ~\cite{peebles2023scalablediffusionmodelstransformers}. We also maintain an exponential moving average(EMA) of DiffMoE and baseline model weights over training with a decay of 0.9999. All results reported use the EMA mode. For most experiments, we utilized 4 NVIDIA H800 GPUs during training. To achieve state-of-the-art results, we extended the training process with 8 NVIDIA H800 GPUs for improved efficiency. For the text-to-image model pre-training, we employ 32 NVIDIA H800 GPUs with our internal datasets, and then conduct supervised fine-tuning (SFT) on the JourneyDB dataset~\cite{pan2023journeydb}.

\subsection{Implementation Algorithms.}
We provide a detailed illustration of the DiffMoE layer during training and inference in Algorithm ~\ref{algo:diff_moe_train} and ~\ref{algo:diff_moe_infer}, respectively. We also implemented the same EC-DiT layer in Algorithm ~\ref{algo:ecdit} as ~\cite{sun2024ecditscalingdiffusiontransformers}. We implemented TC-DiT in layer in Algorithm ~\ref{algo:tcdit} similar to ~\cite{dai2024deepseekmoeultimateexpertspecialization,FeiDiTMoE2024}

\section{Calculation of Average Activated Parameters and Average Capacity $C^{\rm avg}_{\rm infer}$}
\label{appendix:sec_cap}

\subsection{Computing $C^{\rm avg}_{\rm infer}$}
To compute the global average capacity ($C^{\rm avg}_{\rm infer}$), we analyze 50K samples across all experts and sampling steps. For a quick approximation, sampling 1K samples is sufficient due to DiffMoE's stable performance characteristics.

\subsection{Estimating Average Activated Parameters using $C^{\rm avg}_{\rm infer}$}

We will introduce the relationship between average activated parameters and average capacity in detail. Let $\text{\#Module}$ denote the number of parameters a certain Module, $N_E$ denote the number of experts. Under approximate conditions, for DiT~\cite{peebles2023scalable} models, we have

\begin{equation}
    \text{\# Average Activated Parameters} \approx  \left( \frac{1+C^{\rm avg}_{\rm infer}}{1+N_E} \right)\text{\# FFN} + \text{\# Attention} +  \text{\# AdaLN} +  \text{\# Other Modules}.
\end{equation}.

Table ~\ref{tab:componets_params} displays the parameters and their corresponding percentages of the main modules of large-size DiffMoE. 

\begin{table}[t]
  \centering
    \begin{minipage}{0.47\textwidth}
      \begin{minipage}{\textwidth}
        \centering
        \caption{\textbf{Module Parameters and Percentage.} We have counted the number of parameters (M) of various modules to facilitate our analysis. }
        \adjustbox{width=\linewidth}
        {
                \begin{tabular}{llllll}\toprule
                \multicolumn{1}{l}{Model}
                      & \multicolumn{1}{l}{FFN} & \multicolumn{1}{l}{Attention} & \multicolumn{1}{l}{AdaLN} & \multicolumn{1}{l}{Others} & \multicolumn{1}{l}{Total} \\\midrule
                Dense-DiT-L & 201.44\percent{44.0} & 100.7\percent{22.0} & 151\percent{33.0}   & 4.7\percent{1.0}   & 457.84 \\
                DiffMoE-L-E2 & 314.8\percent{55.1} & 100.7\percent{17.6} & 151\percent{26.4}   & 4.7\percent{0.8}   & 571.2 \\
                DiffMoE-L-E4 & 516.3\percent{66.8} & 100.7\percent{13.0} & 151\percent{19.5}   & 4.7(\percent{0.6}   & 772.7 \\
                DiffMoE-L-E8 & 919.3\percent{78.2} & 100.7\percent{8.6} & 151\percent{12.8}   & 4.7\percent{0.4}   & 1175.7 \\
                DiffMoE-L-E16 & 1725.3\percent{87.1} & 100.7\percent{5.1} & 151\percent{7.6})   & 4.7\percent{0.2}   & 1981.7 \\\bottomrule
                \end{tabular}%
        }
        \label{tab:componets_params}%
      \end{minipage}
        \vskip 0.1 in  
      \begin{minipage}{\textwidth}
            \caption{\textbf{Different Threshold Method.} We use both interval search and dynamic threshold method to find out the optimal $\tau_{E_i}$. We find that the dynamic threshold makes a good balance between $C^{\rm avg}_{\rm infer}$ and performance.}
            \begin{minipage}{0.47\textwidth}
            \centering
            \adjustbox{width=\linewidth}
            {
                \begin{tabular}{lccc}\toprule
                \multirow{1}{*}{$\mathcal{T}$} 
                & \multirow{1}{*}{$C^{\rm avg}_{\rm infer}$} 
                & \multirow{1}{*}{FID50K $\downarrow$}
                \\\midrule
                0.999 & 0.41 &  36.28 \\
                0.99 & 0.51 & 24.22 \\
                0.9 & 0.71 & 17.59 \\
                0.8 & 0.78 & 16.21 \\
                0.7 & 0.83 & 15.51 \\
                0.6 & 0.88 & 15.00 \\
                0.5 & 0.92 & 14.59 \\
                0.4 & 0.97 & \textbf{14.16} \\
                 \textcolor{gray}{0.3} & \textcolor{gray}{ 1.03} & \textcolor{gray}{ 13.75}\\
                 \textcolor{gray}{0.2} & \textcolor{gray}{ 1.10} & \textcolor{gray}{ 13.38}\\
                 \textcolor{gray}{0.1} & \textcolor{gray}{ 1.22} & \textcolor{gray}{ 12.92}\\
                 
                 \midrule
                 \rowcolor{Gray} Dynamic & 0.95 & \underline{14.41} \\
                \bottomrule
                \end{tabular}%
            }
            \label{tab:diff_trd_method}%
            \end{minipage}
            \hfill  %
            \begin{minipage}{0.46\textwidth}
                \centering
                \adjustbox{width=\linewidth}
                {
                    \begin{tabular}{lccc}\toprule
                    \multirow{1}{*}{$\mathcal{T}$} 
                    & \multirow{1}{*}{$C^{\rm avg}_{\rm infer}$} 
                    & \multirow{1}{*}{FID50K $\downarrow$}
                    \\\midrule
                    0.2 & 1.10 & 13.38 \\
                    0.1 & 1.22 & 12.92 \\
                    1E-2 & 1.71 & 12.14 \\
                    1E-3 & 2.33 & \textbf{11.82} \\
                    1E-4 & 3.17 & \underline{11.87} \\
                    1E-5 & 4.36 & 12.47 \\
                    1E-6 & 6.01 & 13.51 \\
                    1E-7 & 7.88 & 13.96 \\
                    1E-8 & 9.67 & 16.85 \\
                    1E-9 & 11.18 & 18.90 \\
                    0.0 & 16 & 29.39 \\
                    \bottomrule
                    \end{tabular}%
                }   
            \label{tab:diff_trd_method}%
            \end{minipage} 
        \end{minipage}
    \end{minipage}
    \hfill  %
    \begin{minipage}{0.49\textwidth}
        \begin{minipage}{\textwidth}
        \centering
        \caption{\textbf{Comparisons with the Baseline Models. (DDPM)} We compare TC, EC and Dense Model and show the average activated parameters of all the experts across all the sampling steps.}
        \adjustbox{width=\linewidth}
        {
          \begin{tabular}{lcccc}\toprule
          \multirow{1}{*}{Model  (700K)} 
          & \multirow{1}{*}{\# Avg. Activated Params.} 
          & \multirow{1}{*}{$C^{\rm avg}_{\rm infer}$} 
          & \multirow{1}{*}{FID50K $\downarrow$}
          \\\midrule
          TC-DiT-L-E16-DDPM & 458M & 1 & 20.81  \\
          EC-DiT-L-E16-DDPM & 458M & 1 & 17.65 \\
          Dense-DiT-L-DDPM & 458M & 1 & 17.87 \\
          Dense-DiT-XL-DDPM & 675M & 1 & \underline{15.28} \\
           \rowcolor{Gray} DiffMoE-L-E16-DDPM & 458M & 1 & \textbf{14.60} \\\bottomrule
          \end{tabular}%
        }
        \label{tab:compare_w_baseline_ddpm}%
        \end{minipage}
        \vskip 0.1in
        \begin{minipage}{\textwidth}
            \caption{\textbf{Decoder Ablation Study}. Evaluation of various pre-trained VAE decoder weights. $^\dagger$: results from~\cite{ma2024sit} (DDPM) and~\cite{peebles2023scalable} (Flow). $^*$: our reproduction. All other results are from our experiments. In general, with VAE decoder EMA version, the FID score is consistently lower than MSE version.}
            \centering
            \adjustbox{width=\linewidth}
            {
                \begin{tabular}{lcccccc}\toprule
                \multirow{1}{*}{Model} 
                & \multirow{1}{*}{Training Steps} 
                & \multirow{1}{*}{VAE-Decoder} 
                & \multirow{1}{*}{Sampler} 
                & \multirow{1}{*}{Batch Size} 
                & \multirow{1}{*}{FID50K $\downarrow$}
                \\\midrule
                Dense-DiT-XL-Flow & 400K & ft-MSE & Euler & 125 & 18.80 \\
                Dense-DiT-XL-Flow & 400K & ft-EMA & Euler & 125 & 18.74 &\\
                Dense-DiT-XL-Flow & 400K & ft-MSE & Dopri5 & 125 &  18.63 \\
                Dense-DiT-XL-Flow & 400K & ft-EMA & Dopri5 & 125 &  18.45 &\\\midrule
                
                Dense-DiT-XL-Flow$^*$ & 7000K & ft-MSE & Heun & 125 &  9.66 \\
                Dense-DiT-XL-Flow$^*$ & 7000K & ft-EMA & Heun & 125 &  9.63 \\
                Dense-DiT-XL-Flow$^*$ & 7000K & ft-MSE & Dopri5 & 125 &  9.51 \\
                Dense-DiT-XL-Flow$^*$ & 7000K & ft-EMA & Dopri5 & 125 &  9.48 \\\midrule
    
                Dense-DiT-XL-DDPM-G $^\dagger$ & 7000K & ft-MSE & DDPM & 125 &  2.30\\
                Dense-DiT-XL-DDPM-G $^\dagger$ & 7000K & ft-EMA &  DDPM & 125 &  2.27\\
                \bottomrule
                \label{tab:ablation_vae}
                \end{tabular}%
            }   
        \end{minipage}
    \end{minipage}
\end{table}

\begin{figure}[ht]
    \centering
    \begin{minipage}{0.49\columnwidth} %
        \includegraphics[width=\linewidth]{figs/Loss_Comparison_Ksteps_Smoothed_L-Flow.png}
        \par\vspace{0.1in} %
        \centering %
        \textbf{(a)} \textbf{Loss Comparison of L-Flow Series}
    \end{minipage}
    \hfill %
    \begin{minipage}{0.49\columnwidth} %
        \includegraphics[width=\linewidth]{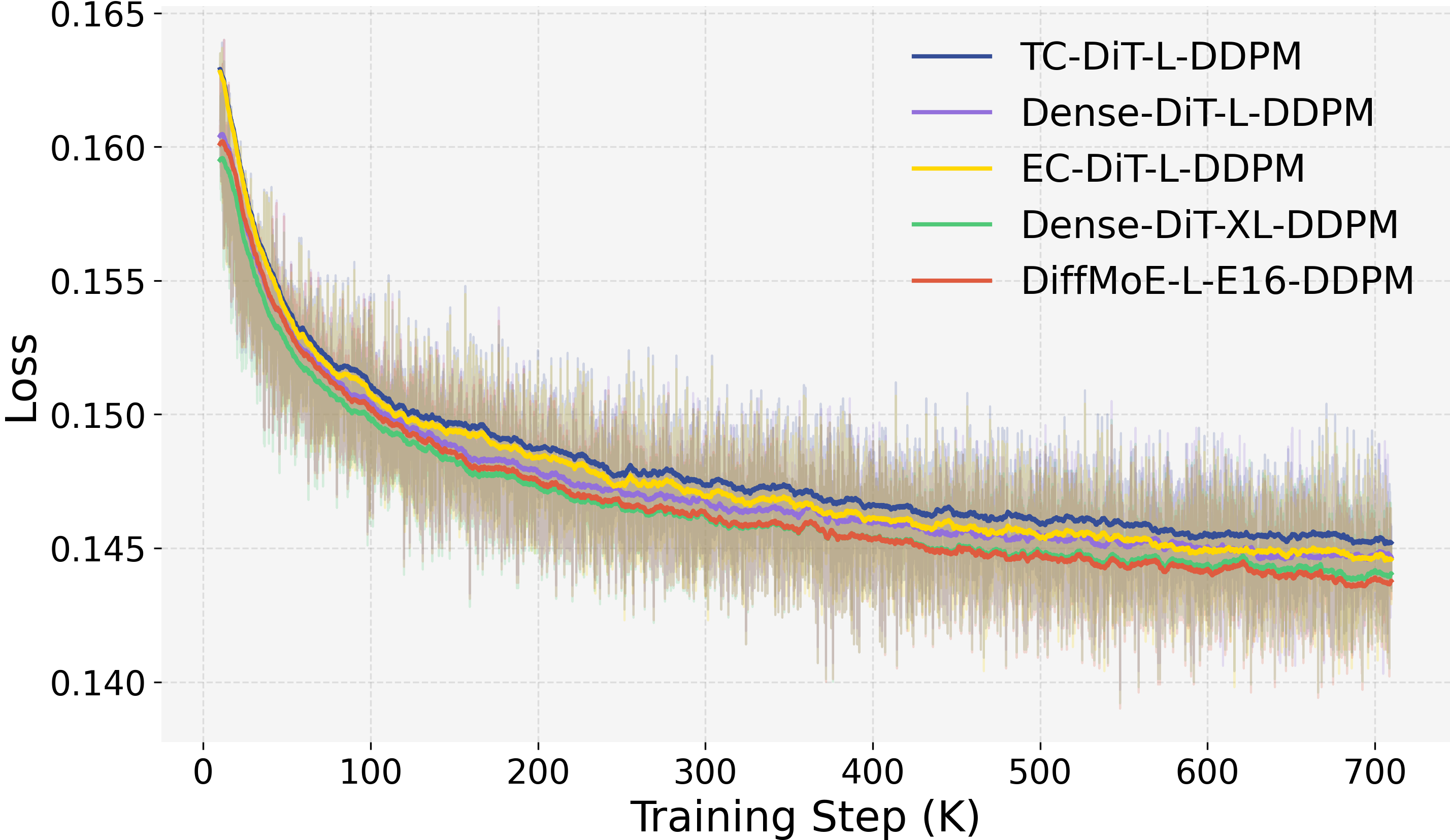}
        \par\vspace{0.1in} %
        \centering %
        \textbf{(b)} \textbf{Loss Comparison of L-DDPM Series}
    \end{minipage}
    \caption{\textbf{Loss Comparison of L-Flow and L-DDPM Series}. The relative losses illustrated in subfigures (a) and (b) demonstrate the exceptional training dynamics of DiffMoE, consistently outperforming all baseline models.}
    \label{fig:loss-comparison}
\end{figure}

\begin{figure}[ht]
    \centering
    \begin{minipage}{0.49\columnwidth} %
        \includegraphics[width=\linewidth]{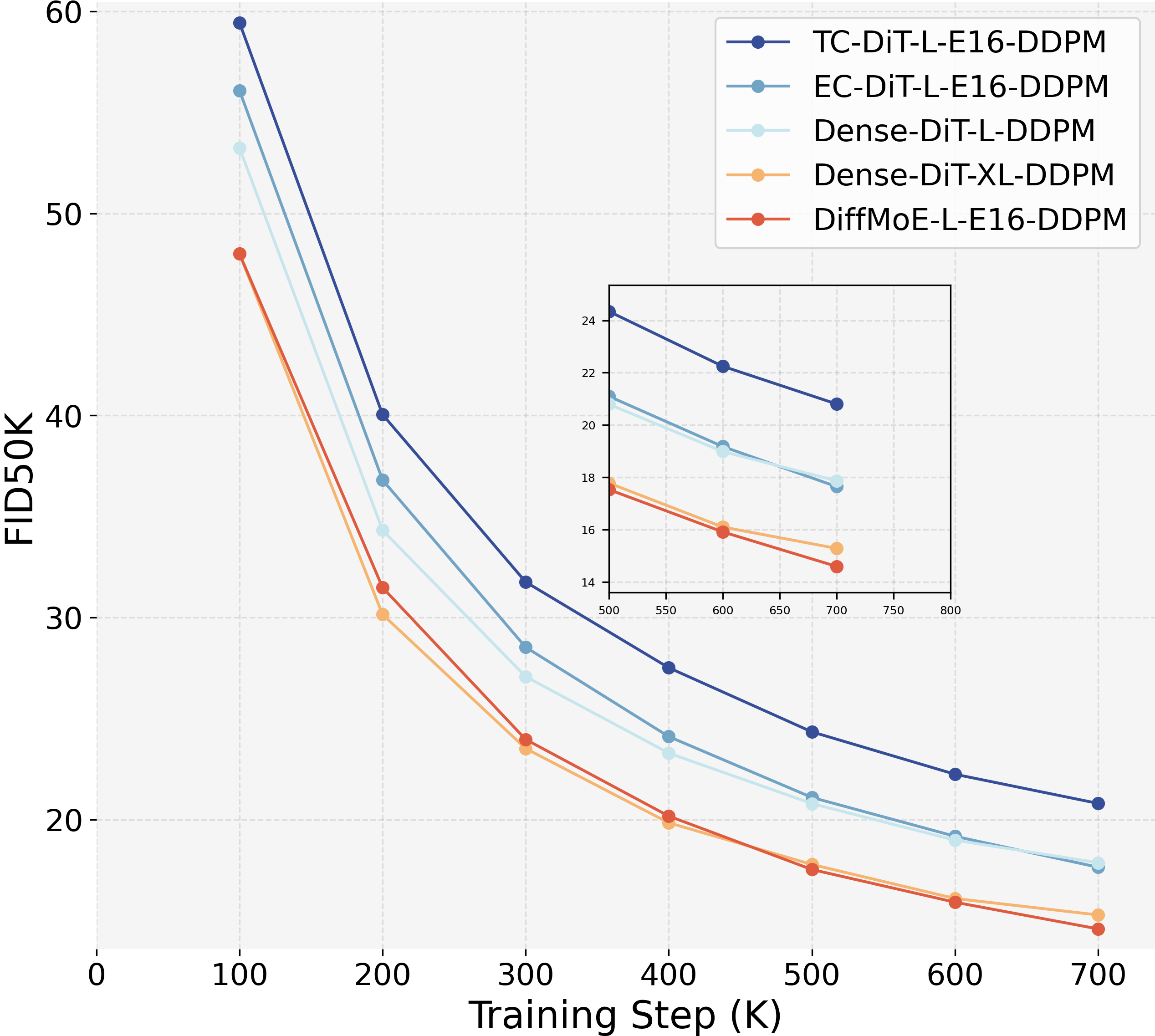}
        \par\vspace{0.1in} %
        \centering %
        \textbf{(a)} \textbf{Comparisons with the Baseline Size-L (DDPM).}
    \end{minipage}
    \hfill %
    \begin{minipage}{0.45\columnwidth} %
        \includegraphics[width=\linewidth]{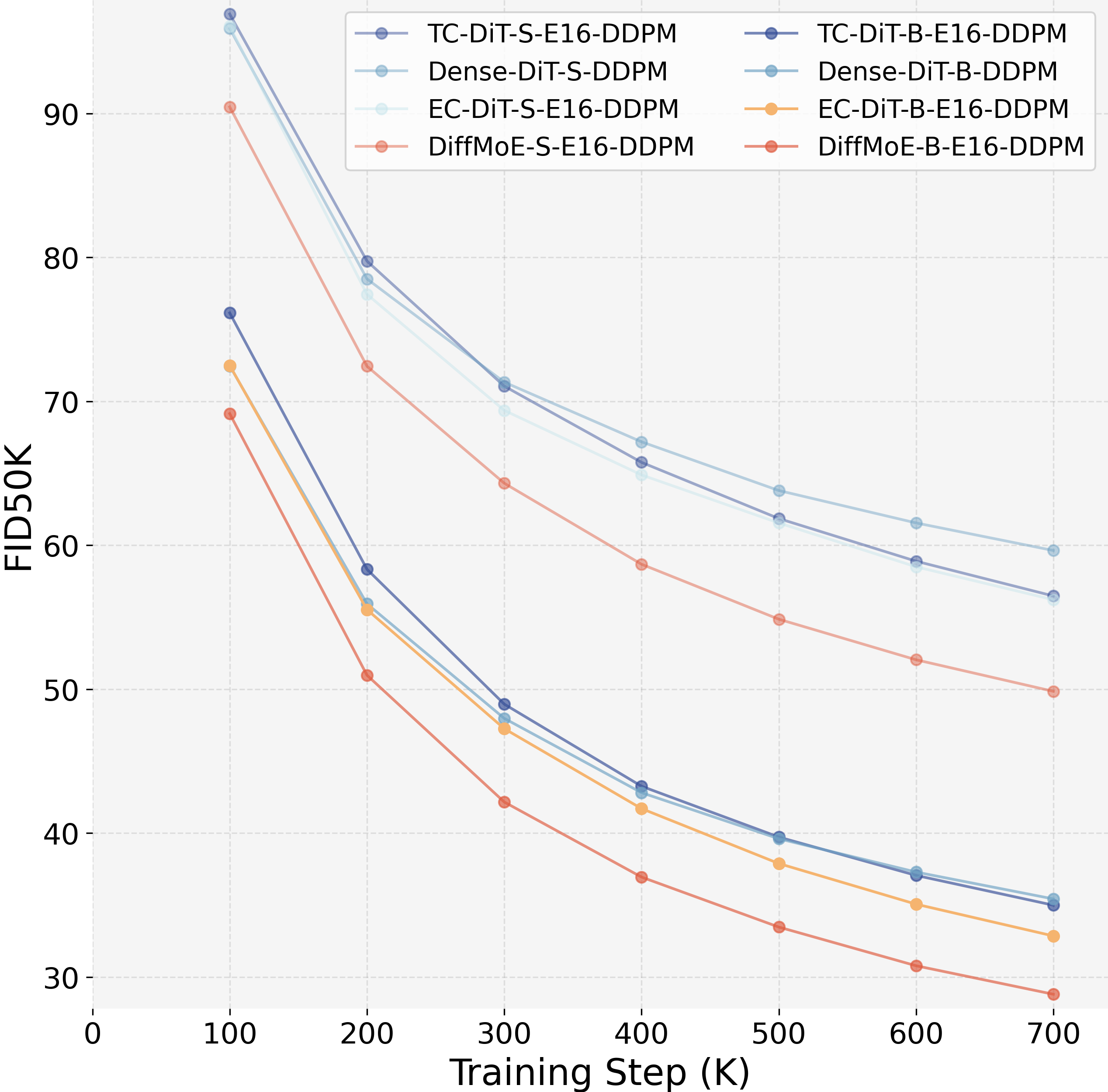}
        \par\vspace{0.1in} %
        \centering %
        \textbf{(b)} \textbf{Comparisons with the Baseline Size-S/B (DDPM).}     
    \end{minipage}
    \caption{\textbf{Comparisons with the Baseline Models.} \textbf{(a)} We compare TC, EC, and Dense Models and show the average activated parameters of all experts across all sampling steps. DiffMoE-L-E16-DDPM even surpasses DenseDiT-XL-DDPM (1.5x params). \textbf{(b)} We also examine S/B size DiffMoE models to further demonstrate the scalability.}
    \label{fig:ddpm_compare_with_baseline}
\end{figure}

\begin{figure}[ht]
    \centering
    \begin{minipage}{\linewidth} %
        \begin{minipage}{0.49\linewidth} %
            \includegraphics[width=\linewidth]{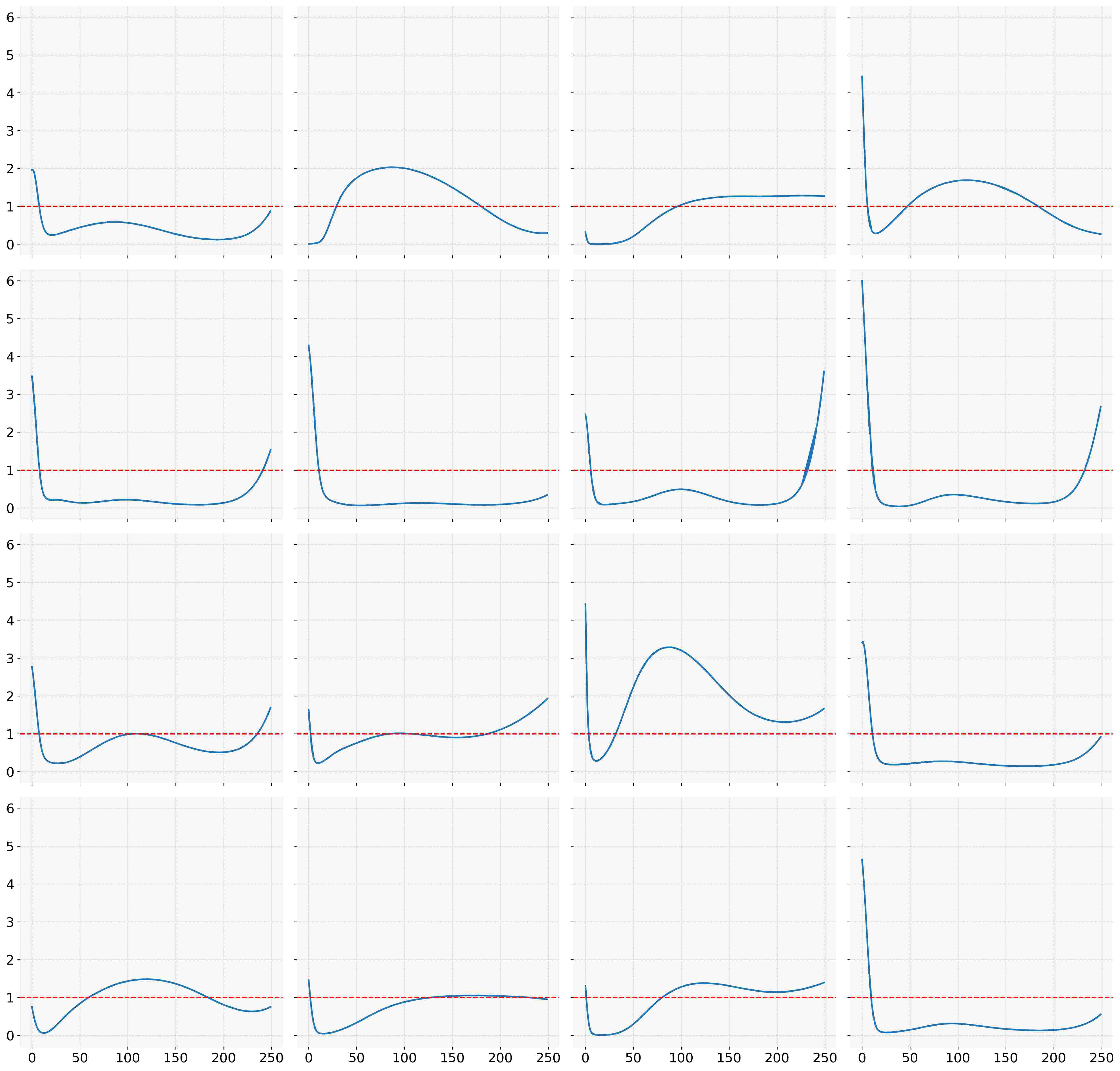}
            \centering %
            \textbf{(a)} Capacity of All Experts in Layer 1
        \end{minipage}
        \hfill
        \begin{minipage}{0.49\linewidth} %
        \includegraphics[width=\linewidth]{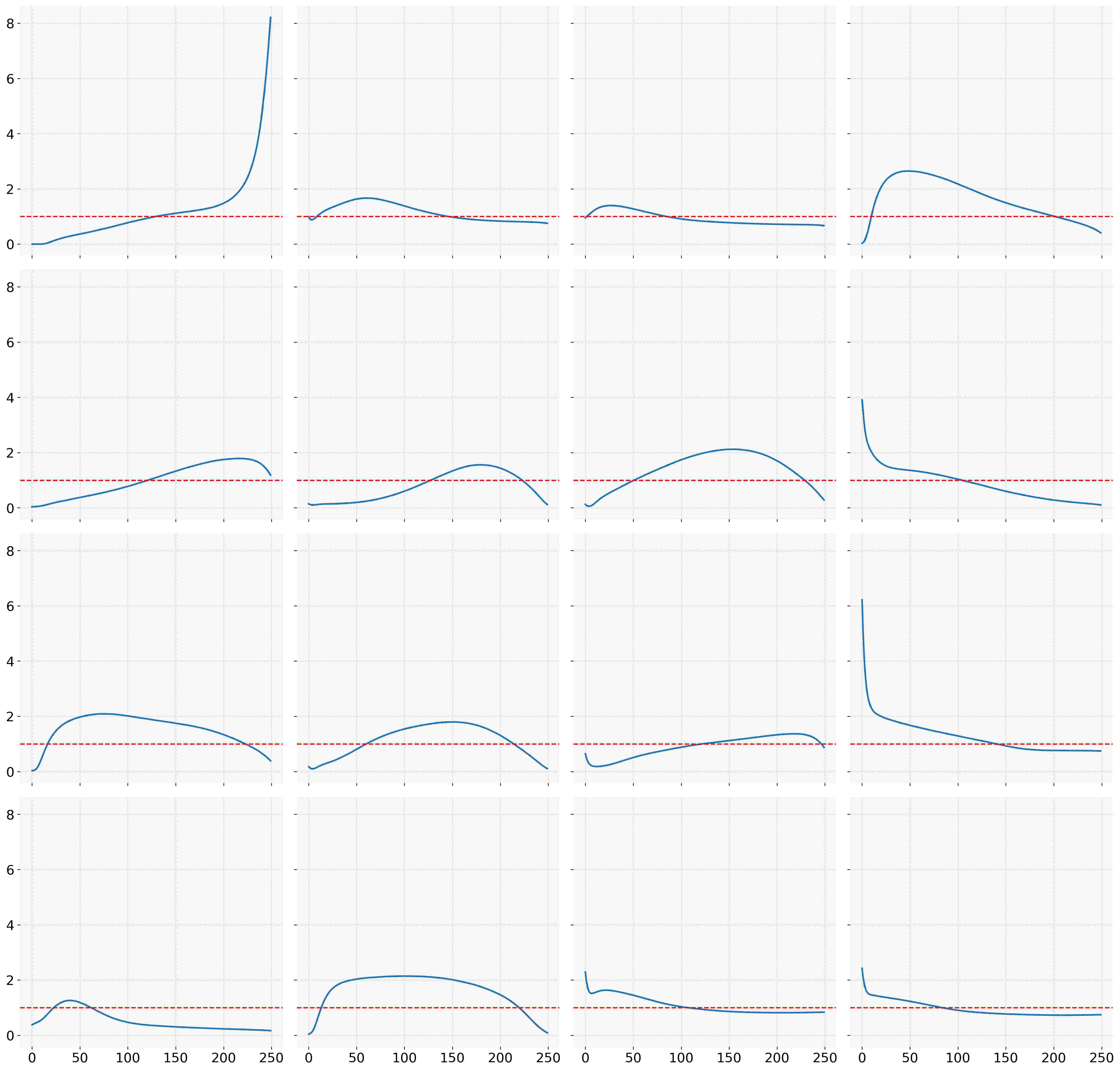}
        \centering %
        \textbf{(b)} Capacity of All Experts in Layer 7
        
        \end{minipage}
    \end{minipage}
    \vskip 0.2cm
    \begin{minipage}{\linewidth} %
        \begin{minipage}{0.49\linewidth} %
            \includegraphics[width=\linewidth]{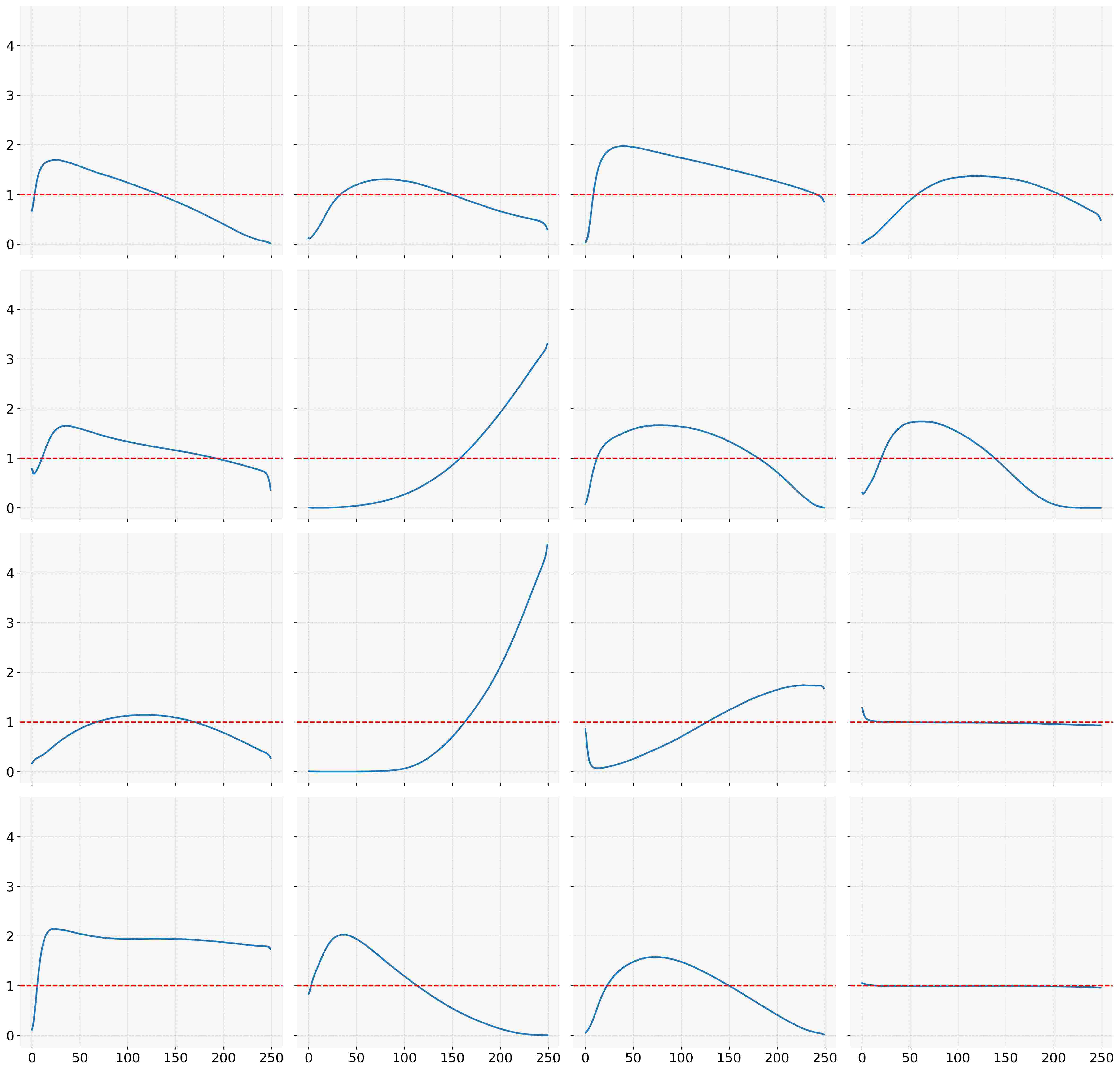}
            \centering %
            \textbf{(c)} Capacity of All Experts in Layer 13
        \end{minipage}
        \hfill
        \begin{minipage}{0.49\linewidth} %
        \includegraphics[width=\linewidth]{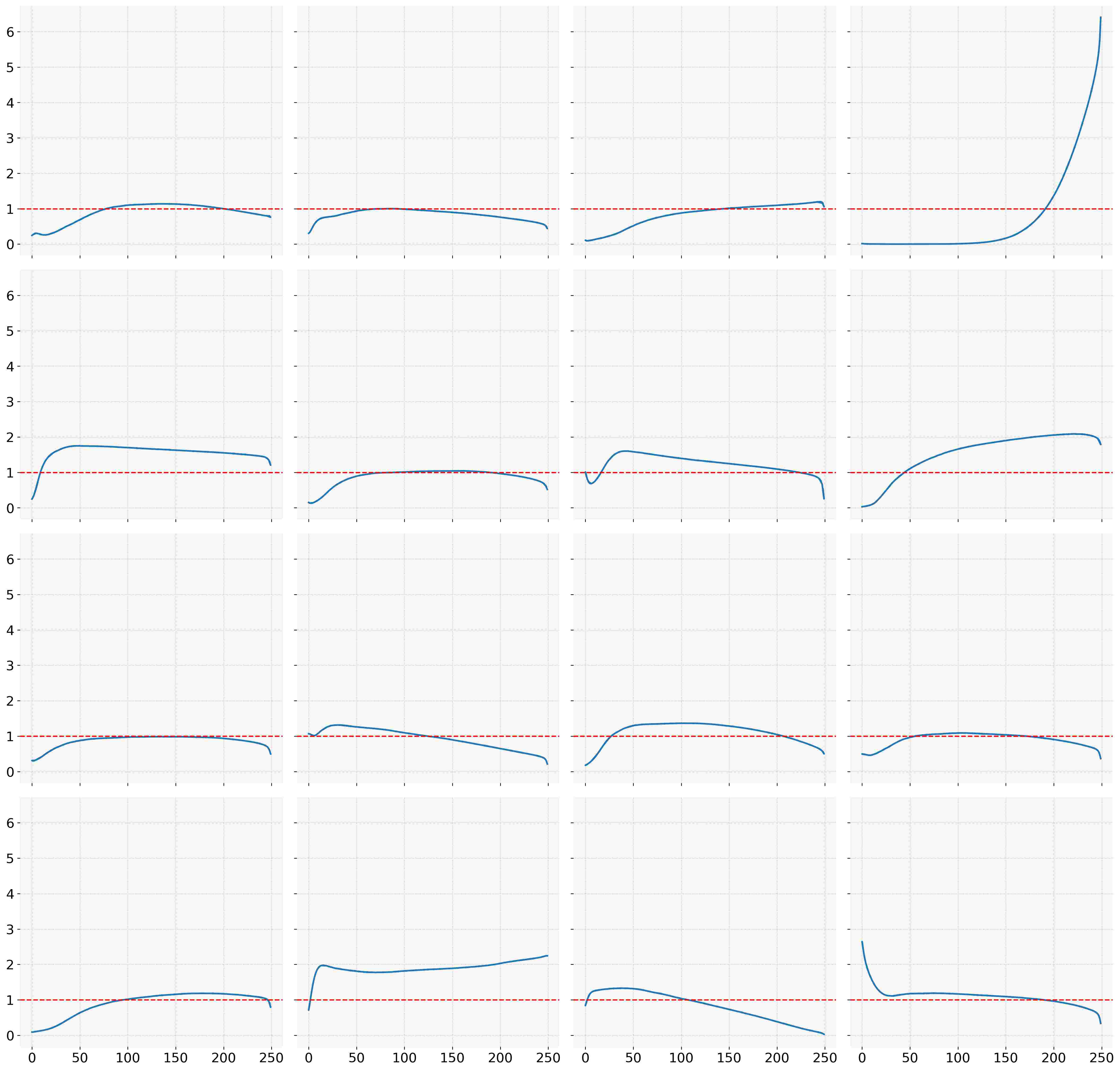}
        \centering %
        \textbf{(d)} Capacity of All Experts in Layer 19
        \end{minipage}
    \end{minipage}

    \caption{\textbf{Expert Dynamics across Network Layers.} Visualization of expert capacity patterns in network layers (1, 7, 13, 19). Early layers show high-amplitude fluctuations, while deeper layers exhibit increasingly stable utilization, demonstrating natural expert specialization throughout the diffusion process.}
    \label{fig:per_expert_dynamics}
\end{figure}

\section{More DiffMoE Analysis}

\subsection{Additional DiffMoE DDPM Class-conditional Generation Qualitative and Quantitative Results}
\label{appendix:sec_ddpm_reults}
We present comprehensive evaluations of DiffMoE-L-E16-DDPM series models. Table~\ref{tab:compare_w_baseline_ddpm} shows the experimental results, while Figure~\ref{fig:loss-comparison} illustrates the diffusion loss comparison against the baseline model, revealing substantial performance improvements. Furthermore, Figure~\ref{fig:ddpm_compare_with_baseline} demonstrates the scaling capabilities of our DiffMoE-DDPM architecture.

\subsection{Additional DiffMoE Text-to-Image Qualitative and Quantitative Results}
\label{appendix:sec_t2i_results}
\begin{figure}[t]
\begin{center}
\centerline{\includegraphics[width=\columnwidth]{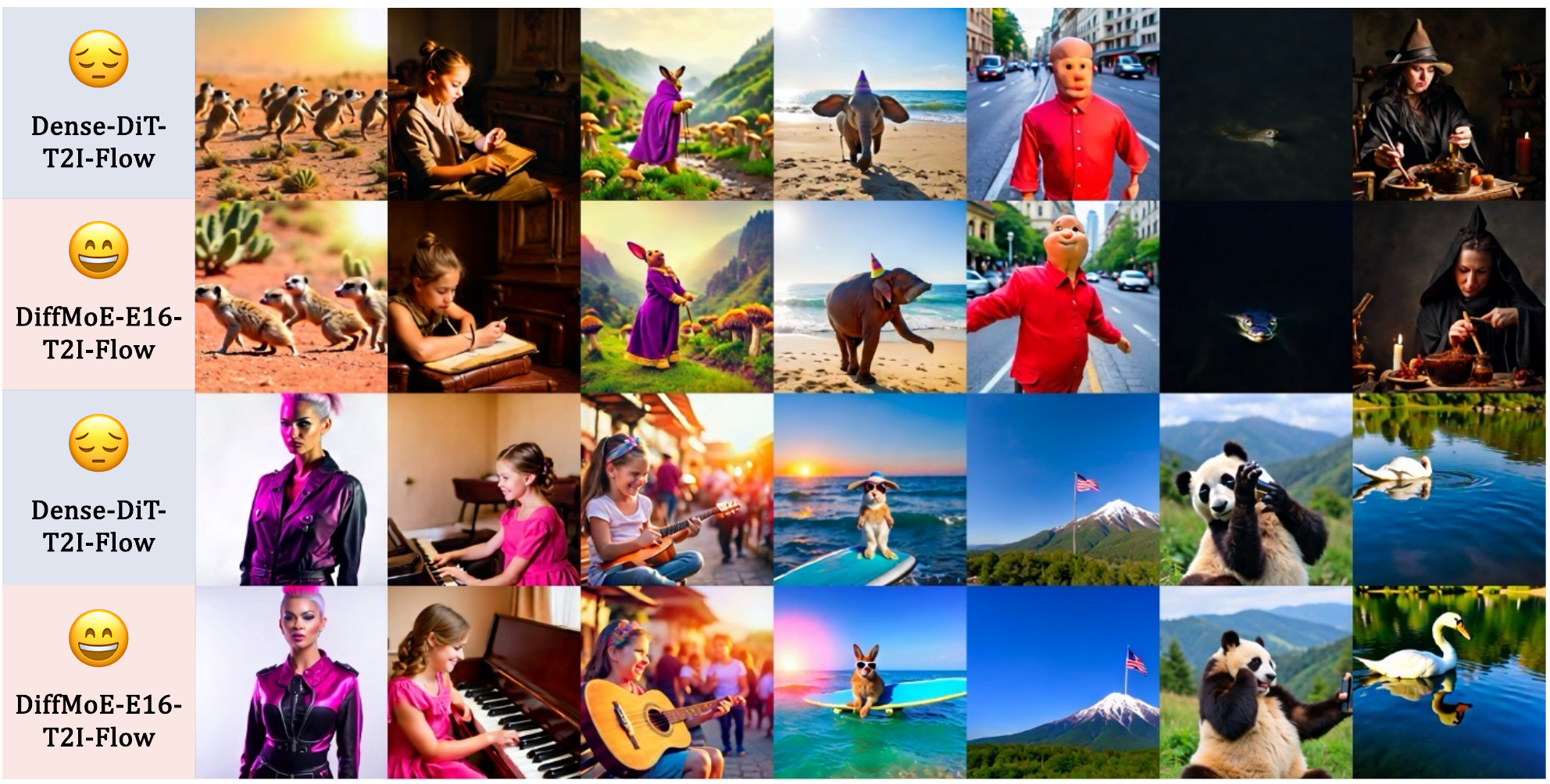}}
\caption{\textbf{Text-to-Image Generation.} Comparison of DiffMoE-E16-T2I-Flow (w/o SFT) and Dense Model (w/o SFT).}
\label{fig:flow_t2i_compare_viz}
\end{center}
\vskip -0.2in
\end{figure}

To further validate the capability of DiffMoE of more challenging task, we conducted quantitative and qualitative evaluations of Dense-DiT-T2I-Flow and DiffMoE-E16-T2I-Flow. The experimental results are presented in Table~\ref{tab:geneval_t2i}, demonstrating the comparative performance metrics of both models. Figure~\ref{fig:loss-t2i} provides a detailed visualization of the diffusion loss comparison between the dense model and DiffMoE, highlighting significant performance improvements achieved by our approach. The superior text-to-image generation capabilities of DiffMoE are further illustrated through qualitative examples in Figure~\ref{fig:flow_t2i_compare_viz} and Figure~\ref{fig:flow_c2i_viz}. It is worth noting that both text-to-image models were only trained for about 2 days, which is limited for this task, suggesting significant room for further performance improvements.

\subsection{Analysis of Token Interaction Strategies.}
As shown in Figure~\ref{fig:teaser}. The interaction levels (L1/L2/L3) represent: L1 for isolated token processing, L2 for local token routing within samples, and L3 for global token routing across sample. Table~\ref{tab:flow_matching_comparison} presents a comprehensive comparison of different token interaction strategies in diffusion models. The baseline models with L1 strategy (TC-DiT-L-Flow and Dense-DiT-L-Flow) process tokens independently, resulting in limited performance (FID: 19.06 and 17.01). The L2 strategy, implemented in EC-DiT-L-Flow, enables local token routing within samples, showing improved performance (FID: 16.12) with the same parameter count. Our proposed L3 strategy in DiffMoE-L-Flow introduces cross-sample token routing, achieving superior results (FID: 14.41) even compared to the 1.5x larger Dense-DiT-XL-Flow (675M parameters). Notably, when combined with Dynamic Global CP, our model not only achieves the best FID score but also reduces the computational capacity to 0.95x, demonstrating both effectiveness and efficiency.

\subsection{Dynamic Conditional Computation: Harder Work needs More Computation}

\begin{figure}[ht]
    \centering
    \begin{minipage}{\linewidth} %
        \includegraphics[width=\linewidth]{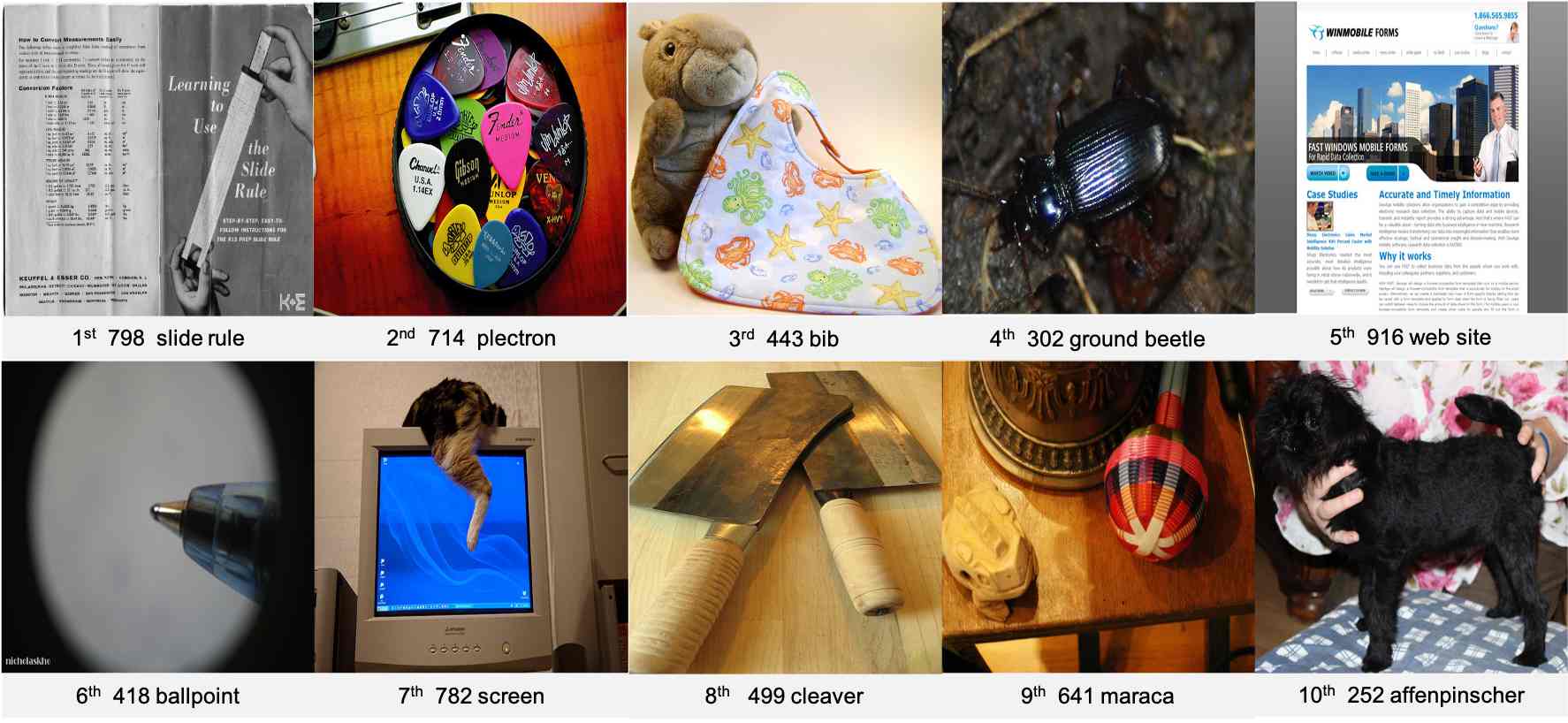}
        \par\vspace{0.1in} %
        \centering %
        \textbf{(a)} \textbf{DiffMoE-L-E16-Flow (700K)}
    \end{minipage}
    \vskip 0.5cm
    \begin{minipage}{\linewidth} %
        \includegraphics[width=\linewidth]{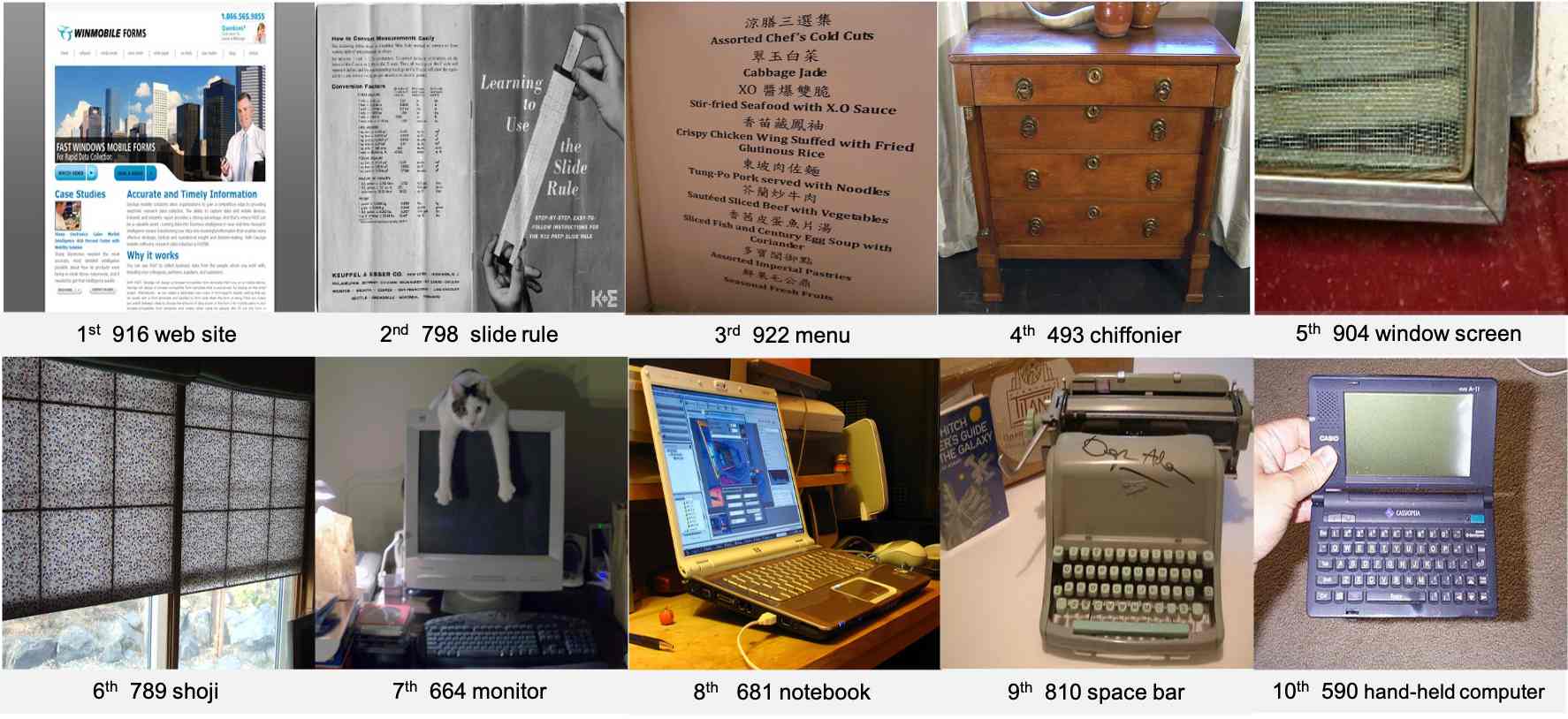}
        \centering %
        \textbf{(b)} \textbf{DiffMoE-L-E16-DDPM (700K)}
    \end{minipage}
    \caption{\textbf{Top 10 Hardest Classes.} The 10 classes with the highest computational cost, sampled from the training set.}
    \label{fig:top10_hardest}
\end{figure}

\begin{figure}[ht]
    \centering
    \begin{minipage}{\linewidth} %
        \includegraphics[width=\linewidth]{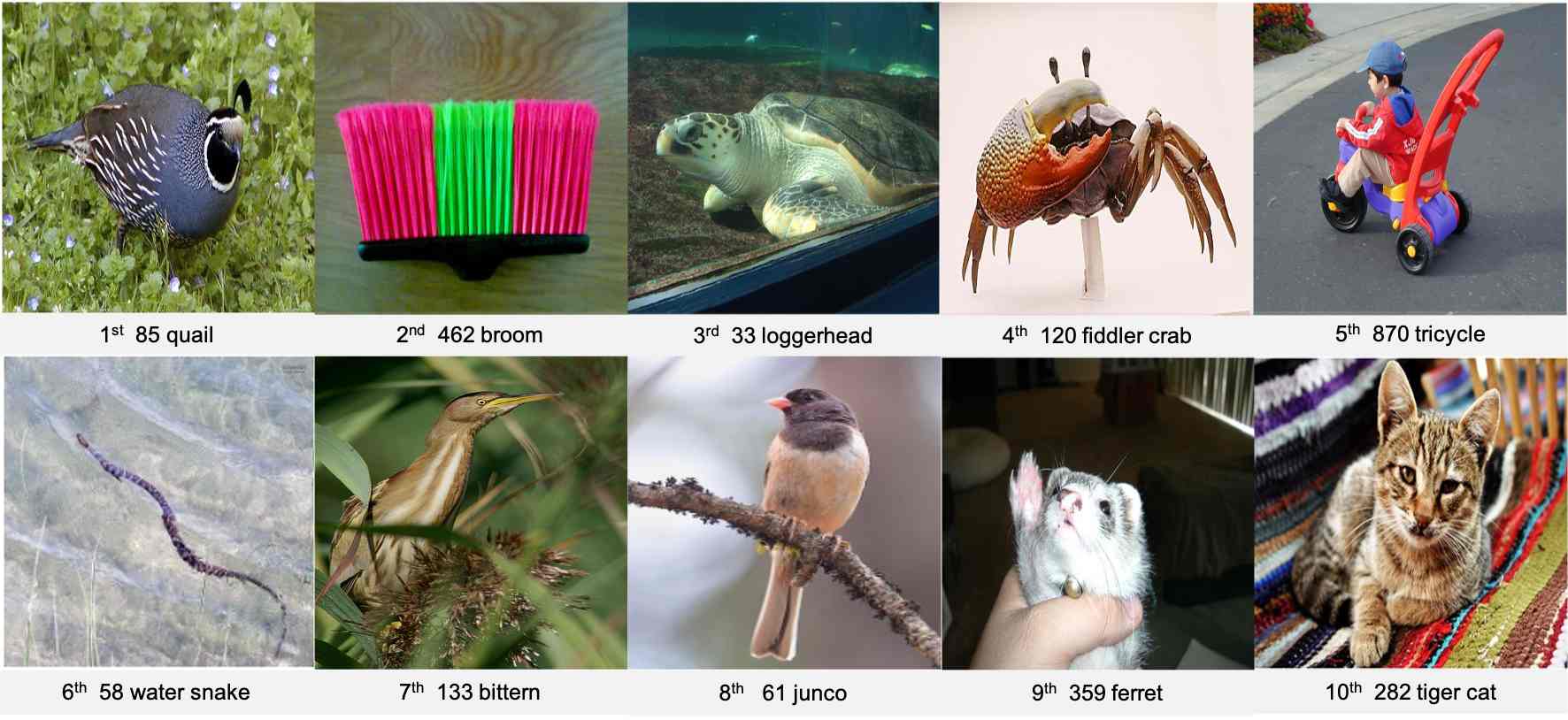}
        \centering %
        \textbf{(a)} \textbf{DiffMoE-L-E16-Flow (700K)}
    \end{minipage}
    \vskip 0.5cm 
    \begin{minipage}{\linewidth} %
        \includegraphics[width=\linewidth]{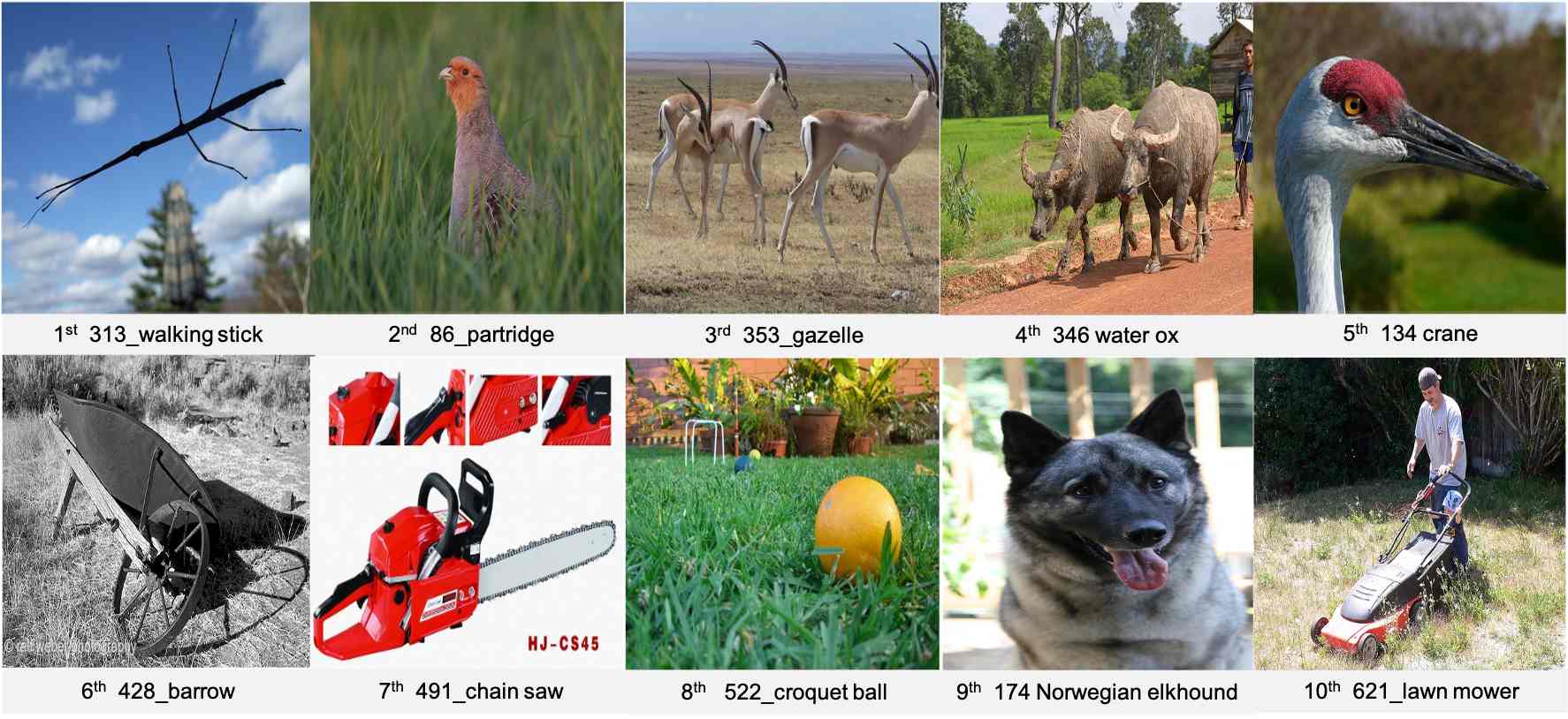}
        \centering %
        \textbf{(b)} \textbf{DiffMoE-L-E16-DDPM (700K)}
    \end{minipage}
    \caption{\textbf{Top 10 Easiest Classes.} The 10 classes with the lowest computational cost, sampled from the training set.}
    \label{fig:top10_easiest}
\end{figure}

Figure~\ref{fig:teaser} demonstrates that different classes require varying computational resources during generation. To analyze this variation, we sample 1K different class labels in a batch and rank their $C^{\rm avg}_{\rm infer}$ in descending order, revealing the computational complexity of generation across classes. The top-10 most computationally intensive classes for both flow-based and DDPM models are displayed in Figure ~\ref{fig:top10_hardest}. The top-10 least computationally intensive classes for both flow-based and DDPM models are displayed in Figure ~\ref{fig:top10_easiest}.

\subsection{Dynamic Token Selection across Network Layers}
As illustrated in Figure ~\ref{fig:per_expert_dynamics}, our analysis reveals distinctive expert utilization patterns across different network depths. The shallow Layer 1 exhibits pronounced fluctuations and sharp capacity spikes, indicating intensive early-stage feature extraction. Moving to intermediate Layer 7, we observe more stabilized capacity patterns, suggesting balanced processing of mid-level features. Layer 13 demonstrates gradual, long-term capacity transitions, while the deep Layer 19 shows notably uniform expert utilization. This systematic progression from volatile to stable expert engagement reflects the natural specialization of experts: from low-level feature detection in early layers to refined semantic processing in deeper layers. Such hierarchical organization of expert behaviors aligns with the progressive nature of diffusion-based generation.

\begin{table}[t]
  \centering
      \begin{minipage}{0.48\textwidth}
        \centering
        \caption{\textbf{Batch Size Ablation Study:} FID scores under different batch sizes using fine-tuned EMA VAE decoder and Heun sampler. \textbf{Bold} indicates best performance in its cell.}
        \adjustbox{width=\linewidth}
        {
                \begin{tabular}{lcccccc}\toprule
                \multirow{1}{*}{Model} 
                & \multirow{1}{*}{Training Steps} 
                & \multirow{1}{*}{CFG} 
                & \multirow{1}{*}{Batch Size} 
                & \multirow{1}{*}{FID50K $\downarrow$}
                \\\midrule

                DiffMoE-L-E8-Flow & 7000K & 1.0 & 10 & \textbf{9.60}\\ 
                DiffMoE-L-E8-Flow & 7000K & 1.0 & 15 & 9.62\\ 
                DiffMoE-L-E8-Flow & 7000K & 1.0 & 32 & 9.77\\ 
                DiffMoE-L-E8-Flow & 7000K & 1.0 & 50 & 9.98\\ 
                DiffMoE-L-E8-Flow & 7000K & 1.0 & 75 & 9.90\\ 
                DiffMoE-L-E8-Flow & 7000K & 1.0 & 100 & 9.76\\ 
                DiffMoE-L-E8-Flow & 7000K & 1.0 & 125 & 9.78\\\midrule

                Dense-DiT-XL-Flow$^*$ & 7000K & 1.0 & 10 &  9.57 \\ 
                Dense-DiT-XL-Flow$^*$ & 7000K & 1.0 & 15 & \textbf{9.47} \\ 
                Dense-DiT-XL-Flow$^*$ & 7000K & 1.0 & 50 & 9.84 \\ 
                Dense-DiT-XL-Flow$^*$ & 7000K & 1.0 & 75 & 9.67 \\ 
                Dense-DiT-XL-Flow$^*$ & 7000K & 1.0 & 100 & 9.65 \\ 
                Dense-DiT-XL-Flow$^*$ & 7000K & 1.0 & 125 & 9.64 \\\midrule

                DiffMoE-L-E8-Flow & 7000K & 1.5 & 10 & 2.19\\ 
                DiffMoE-L-E8-Flow & 7000K & 1.5 & 32 & 2.16\\ 
                DiffMoE-L-E8-Flow & 7000K & 1.5 & 50 & 2.16\\ 
                DiffMoE-L-E8-Flow & 7000K & 1.5 & 75 & \textbf{2.13}\\ 
                DiffMoE-L-E8-Flow & 7000K & 1.5 & 100 & 2.17\\ 
                DiffMoE-L-E8-Flow & 7000K & 1.5 & 125 & 2.18\\\midrule

                Dense-DiT-XL-Flow$^*$ & 7000K & 1.5 & 10 &  2.23 \\ 
                Dense-DiT-XL-Flow$^*$ & 7000K & 1.5 & 15 & 2.23 \\ 
                Dense-DiT-XL-Flow$^*$ & 7000K & 1.5 & 50 & 2.21 \\ 
                Dense-DiT-XL-Flow$^*$ & 7000K & 1.5 & 75 & \textbf{2.19} \\ 
                Dense-DiT-XL-Flow$^*$ & 7000K & 1.5 & 100 & 2.22\\ 
                Dense-DiT-XL-Flow$^*$ & 7000K & 1.5 & 125 & 2.21 \\\midrule
                DiffMoE-L-E8-DDPM & 7000K & 1.5 & 50 & \textbf{2.30}\\ 
                DiffMoE-L-E8-DDPM & 7000K & 1.5 & 75 & 2.33\\ 
                DiffMoE-L-E8-DDPM & 7000K & 1.5 & 100 & 2.32\\ 
                DiffMoE-L-E8-DDPM & 7000K & 1.5 & 125 & 2.32\\

                \bottomrule
                \label{tab:fid_abla_batchsize}
                \end{tabular}%
        }
        \label{tab:fid_abla_batchsize}%
        \end{minipage}
        \hfill  %
        \begin{minipage}{0.48\textwidth}
        \begin{minipage}{0.97\textwidth}
        \centering
        \caption{\textbf{CFG Scale Ablation Study:} FID scores across different CFG scales using fine-tuned EMA VAE decoder. \textbf{Bold} indicates best performance in its cell.}
        \adjustbox{width=\linewidth}
        {
                \begin{tabular}{lcccccc}\toprule
                \multirow{1}{*}{Model} 
                & \multirow{1}{*}{Training Steps} 
                & \multirow{1}{*}{Sampler} 
                & \multirow{1}{*}{CFG} 
                & \multirow{1}{*}{Batch Size} 
                & \multirow{1}{*}{FID50K $\downarrow$}
                \\\midrule
                DiffMoE-L-E8-Flow & 4900K &Heun & 1.0 & 125 & \textbf{9.21} \\ 
                 Dense-DiT-XL-Flow$^*$ & 7000K &Heun & 1.0 & 125 & 9.64 \\
                DiffMoE-L-E8-Flow & 7000K &Heun  & 1.0 & 125 & 9.78 \\\midrule
                DiffMoE-L-E8-Flow & 4900K &Heun & 1.43 & 125 & 2.14\\
                Dense-DiT-XL-Flow$^*$ & 7000K &Heun & 1.43 & 125 &  \textbf{2.08} \\ 
                DiffMoE-L-E8-Flow & 7000K &Heun  & 1.43 & 125 & 2.13 \\ \midrule
                DiffMoE-L-E8-Flow & 4900K &Heun & 1.5  & 125 & 2.28 \\ 
                Dense-DiT-XL-Flow$^*$ & 7000K &Heun & 1.5 & 125 & 2.21 \\
                DiffMoE-L-E8-Flow & 7000K &Heun & 1.5& 125 & \textbf{2.18} \\ \midrule
                DiffMoE-L-E8-DDPM & 6500K &DDPM & 1.0  & 125 & {9.39} \\ 
                Dense-DiT-XL-DDPM$^*$ & 7000K &DDPM & 1.0 & 125 & 9.63 \\
                DiffMoE-L-E8-DDPM & 7000K &DDPM & 1.0 & 125 & \textbf{9.17} \\ \midrule
                DiffMoE-L-E8-DDPM & 6500K &DDPM & 1.5  & 125 & \textbf{2.27} \\ 
                Dense-DiT-XL-DDPM$^*$ & 7000K &DDPM & 1.5 & 125 & 2.32 \\
                DiffMoE-L-E8-DDPM & 7000K &DDPM & 1.5 & 125 & 2.32 \\
                \bottomrule
                \label{tab:fid_abla_cfg}
                \end{tabular}%
        }
        \label{tab:fid_abla_cfg}%
        \end{minipage}
        \vskip 0.1in
        \begin{minipage}{0.97\textwidth}
        \centering
        \caption{\textbf{Flow ODE Sampler Ablation Study:} FID scores across different ODE samplers with CFG scale 1.0 and fine-tuned EMA VAE decoder. \textbf{Bold} indicates best performance in its cell.}
        \adjustbox{width=\linewidth}
        {
                \begin{tabular}{lcccccc}\toprule
                \multirow{1}{*}{Model} 
                & \multirow{1}{*}{Training Steps} 
                & \multirow{1}{*}{Sampler} 
                & \multirow{1}{*}{Batch Size} 
                & \multirow{1}{*}{FID50K $\downarrow$}
                \\\midrule
                DiffMoE-L-E8-Flow & 4900K  & Euler & 125 & 9.37 \\
                DiffMoE-L-E8-Flow & 4900K  & Heun & 125 & 9.21 \\ 
                DiffMoE-L-E8-Flow & 4900K  & Euler & 250 & 9.39 \\ 
                DiffMoE-L-E8-Flow & 4900K  & Dopri5 & 250 & \textbf{9.06} \\\midrule
                DiffMoE-L-E8-Flow & 7000K  & Euler & 125 & 9.86 \\
                DiffMoE-L-E8-Flow & 7000K  & Heun & 125 & 9.78 \\ 
                DiffMoE-L-E8-Flow & 7000K  & Euler & 250 & 9.94 \\ 
                DiffMoE-L-E8-Flow & 7000K  & Dopri5 & 250 & \textbf{9.56} \\ 
                
                \bottomrule
                \label{tab:fid_abla_sampler}
                \end{tabular}%
        }
        \label{tab:fid_abla_batchsize}%
        \end{minipage}
        \end{minipage}
\end{table}

\section{FID Sensibility and Ablation Study}
FID scores are sensitive to implementation details, necessitating careful ablation studies to understand the differences between various implementations. Through these studies, we aim to provide the academic community with clearer insights for fair comparisons between diffusion models.

The observed FID degradation at higher CFG scales is well-documented~\cite{ma2024sit}, primarily due to ImageNet's diverse image quality distribution. When generating high-quality samples, the deviation from ImageNet's mixed-quality dataset can lead to increased FID scores, despite improved visual quality.

For FID calculation, we follow the implementations from SiT~\cite{ma2024sit} and DiT~\cite{peebles2023scalable}. Results marked with $^\dagger$ are directly quoted from~\cite{ma2024sit} (DDPM) and~\cite{peebles2023scalable} (Flow). For results marked with $^*$, we reproduce the experiments using officially released checkpoints under identical evaluation conditions.

\subsection{VAE Decoder Ablations}
Following ~\cite{peebles2023scalable}, throughout our experiments, we employed pre-trained VAE models. Specifically, we utilized fine-tuned versions (ft-MSE and ft-EMA) of the original LDM "f8" model, where only the decoder weights were fine-tuned. For the analysis presented in Experiments section ~\ref{sec:exp} and Tables ~\ref{tab:compare_w_baseline_ddpm}  and ~\ref{tab:diff_trd_method} , we tracked metrics using the ft-MSE decoder, while the final metrics reported in Table ~\ref{tab:sota} was obtained using the ft-EMA decoder. In this section, we examine the impact of two distinct VAE decoders for our experiments - the two fine-tuned variants employed in Stable Diffusion. Since all models share identical encoders, we can interchange decoders without necessitating diffusion model retraining. As demonstrated in Table ~\ref{tab:ablation_vae}, the DiffMoE model maintains its superior performance over existing diffusion models.

\subsection{Flow ODE-Sampler Ablations}
Higher-order ODE samplers generally achieve better FID scores. As shown in Table~\ref{tab:fid_abla_sampler}, the black-box dopri5 sampler outperforms heun (NFE=250), which in turn surpasses euler (NFE). For fair comparison with baseline models, we employ the euler sampler in flow-based experiments. However, to benchmark against SiT-XL (Dense-DiT-XL-Flow)~\cite{ma2024sit}, we use the heun sampler to achieve SOTA results.

\subsection{Classifier-Free-Guidence Ablations}

We evaluate different models with varying classifier-free guidance (CFG) scales in Table ~\ref{tab:fid_abla_cfg}, and discover that the CFG scale of 1.5 adopted in DiT ~\cite{peebles2023scalable}  and SiT ~\cite{ma2024sit} studies may not be universally optimal. Our analysis reveals the best CFG scale approximates to 1.43 through comprehensive comparisons. However, different models exhibit distinct characteristics that lead to varying optimal CFG scales, suggesting that fixing a uniform scale across all models could introduce evaluation bias. To ensure relatively fair comparisons while maintaining consistency with established practices, we ultimately adopt CFG 1.5 as the default setting in our experiments. This decision aligns with the well-documented trade-off in diffusion models: higher CFG scales (\eg, 4.0) typically enhance image fidelity at the cost of increased FID scores, while lower scales (\eg, 1.5) yield better FID metrics despite reduced perceptual quality. This phenomenon primarily stems from FID's sensitivity to distributional coverage - higher guidance scales tend to produce samples with reduced diversity that more closely match the training distribution statistics, paradoxically resulting in worse FID scores despite improved individual sample quality.

\subsection{Batch Sizes Ablations}

The batch sizes ablation study reveals critical insights into the interplay between batch size and classifier-free guidance (CFG) scales for the DiffMoE-L-E8-Flow model. At CFG=1.0, FID scores remain elevated (9.60\textendash 9.98), with smaller batch sizes (\textit{e.g.}, \texttt{bs=10}) marginally outperforming larger configurations, exhibiting a U-shaped trend. However, elevating CFG to 1.5 drastically reduces FID to 2.13\textendash 2.19, achieving optimal performance at \texttt{bs=75} (\textbf{2.13}), while demonstrating remarkable robustness to batch size variations ($\Delta$=0.06 vs. $\Delta$=0.38 at CFG=1.0).

\subsection{Conclusion: A little thought about FID}
While Fréchet Inception Distance (FID) is widely adopted for evaluating generative models, particularly on ImageNet, it exhibits several notable limitations. Our analysis reveals counterintuitive behaviors, especially when evaluating models with classifier-free guidance (CFG). For example, higher CFG scales typically enhance perceptual quality but paradoxically result in worse FID scores, despite producing visually superior images. This discrepancy stems from FID's fundamental mechanism: it measures statistical similarities between generated and real distributions in the Inception network's feature space, often failing to capture perceptual quality and fine-grained details. Moreover, FID scores are susceptible to various implementation factors, including choice of ODE samplers, hardware configurations, random seeds, and sample size for estimation. These sensitivities can impact reproducibility and comparison across different studies. Furthermore, FID's focus on distributional overlap overlooks critical aspects such as mode collapse and overfitting, as it does not explicitly evaluate sample diversity or novelty. These limitations underscore the pressing need for more robust and comprehensive metrics that can better reflect the true modeling capabilities of generative models. We advocate for developing new evaluation frameworks that combine precision-recall curves, perceptual quality metrics, and human evaluation studies, which would provide a more reliable assessment of generative model performance.

\section{Visual Generation Results}

\subsection{Class-Conditional Image Generation}
\begin{figure}[t]
\begin{center}
\centerline{\includegraphics[width=\columnwidth]{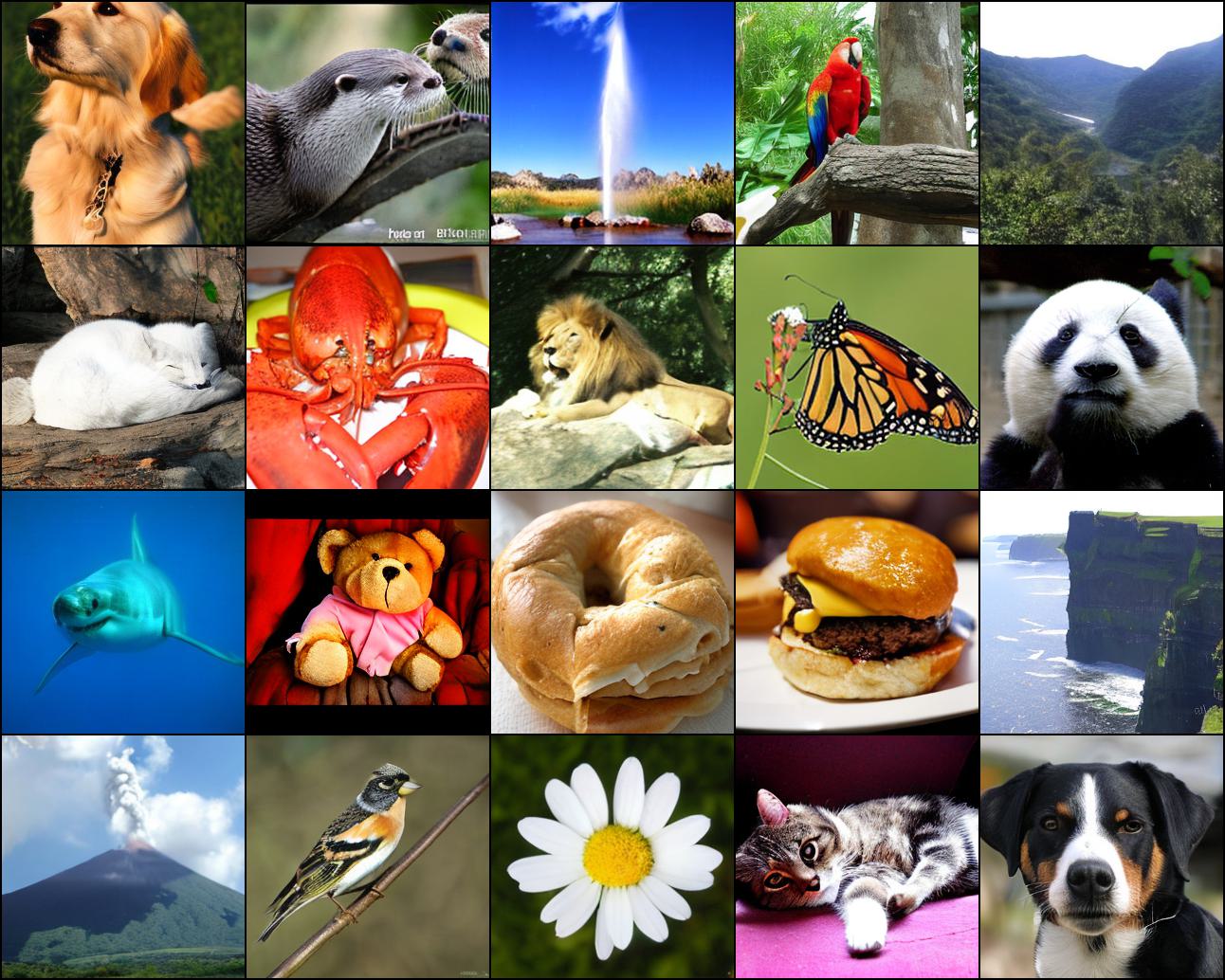}}
\caption{\textbf{Class-conditional Generation.} \textbf{Uncurated} 256$\times$256 \textbf{DiffMoE-L-E8-Flow} samples CFG scale = 4.0.}
\label{fig:flow_c2i_viz}
\end{center}
\vskip -0.2in
\end{figure}

\begin{figure}[t]
\begin{center}
\centerline{\includegraphics[width=\columnwidth]{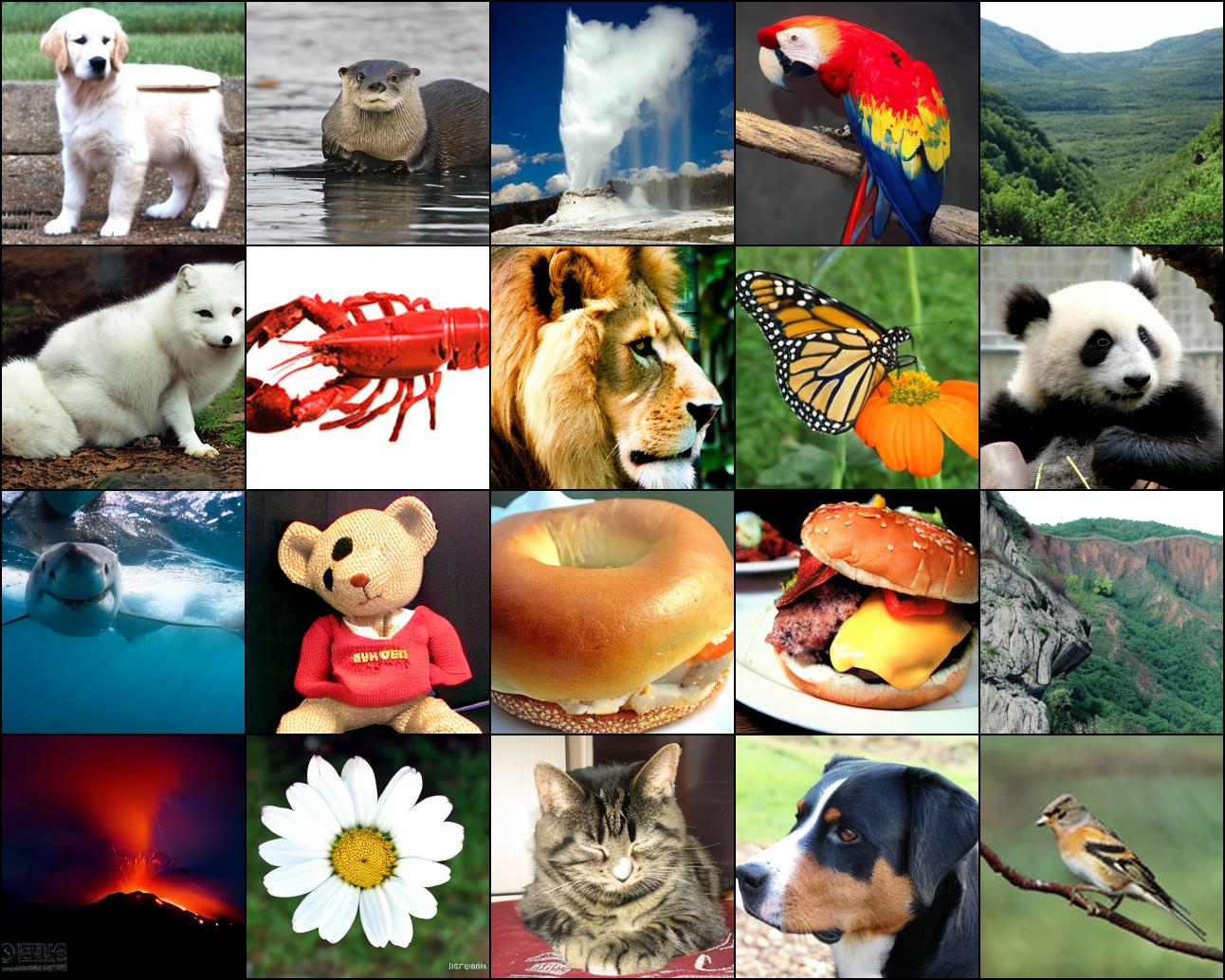}}
\caption{\textbf{Class-conditional Generation.} \textbf{Uncurated} 256$\times$256 \textbf{DiffMoE-L-E8-DDPM} samples CFG scale = 4.0.}
\label{fig:ddpm_c2i_viz}
\end{center}
\vskip -0.2in
\end{figure}

To demonstrate the generation capabilities of our model, we showcase diverse images sampled from DiffMoE-L-E8-Flow and   DiffMoE-L-E8-DDPM, conditioned on ImageNet class labels. These visualizations illustrate the model's ability to generate high-quality, class-specific images. See Figure ~\ref{fig:flow_c2i_viz} and ~\ref{fig:ddpm_c2i_viz}.

\subsection{Text-Conditional Image Generation}
\begin{figure}[t]
\begin{center}
\centerline{\includegraphics[width=\columnwidth]{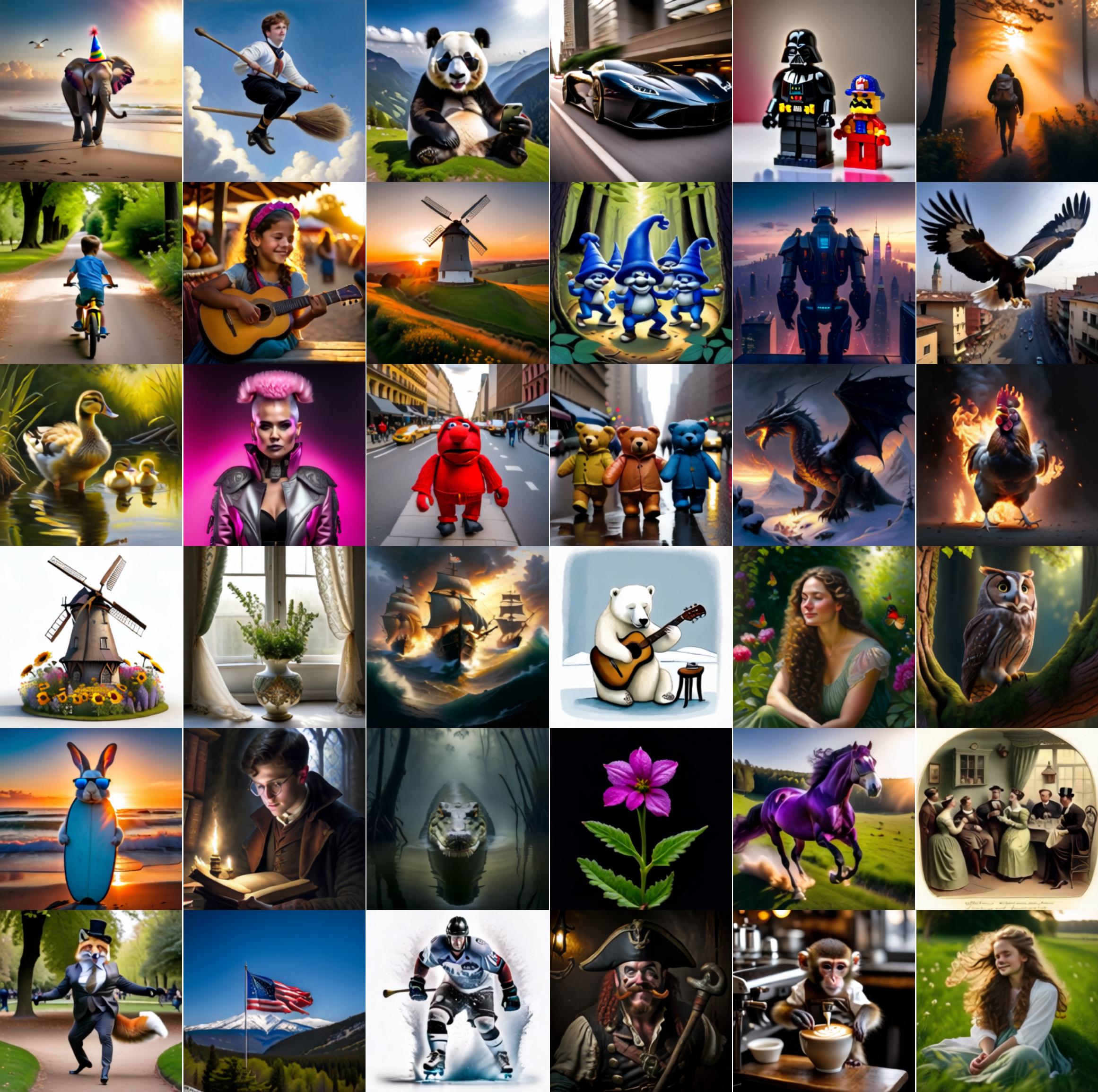}}
\caption{\textbf{Text-to-Image Generation.} \textbf{Uncurated} 256$\times$256 Images generated by \textbf{DiffMoE-E16-T2I-Flow} (w SFT)}
\label{fig:flow_t2i_viz}
\end{center}
\vskip -0.2in
\end{figure}

We present a collection of images generated by our DiffMoE-T2I-Flow model using various text prompts as conditioning inputs. These examples demonstrate the model's versatility in translating textual descriptions into corresponding visual representations. See Figure ~\ref{fig:flow_t2i_viz}.

\begin{algorithm}[b]
   \caption{EC-DiT Layer}
   \label{algo:ecdit}
\begin{algorithmic}
   \STATE {\bfseries Input:} $x$ (input tensor)
   \STATE {\bfseries Variables:} $B$ (batch size), $S$ (sequence length), $d$ (hidden dim), $W_r$ (routing weights), $experts$ (list of expert FFNs)
   \STATE \hspace{3.8em} $E$ (number of experts), $C$ (expert capacity)
   
   \STATE /* Step 1: Compute Token-Expert Affinity Matrix */

   \STATE logits $\leftarrow$ einsum($'bsd,de\rightarrow bse'$, $x$, $W_r$)
   \STATE scores $\leftarrow \texttt{softmax}(logits, \text{dim}=-1).\texttt{permute}(-1, -2)$
   
   \STATE /* Step 2: Select top-k tokens for each expert */
   \STATE gating, index $\leftarrow$ top\_k(scores, k=$C$, dim=$-1$)
   \STATE dispatch $\leftarrow$ one\_hot(index, num\_classes=$S$)
   
   \STATE /* Step 3: Process tokens through experts and combine */
   \STATE $x_{in} \leftarrow$ einsum($'becs,bsd\rightarrow becd'$, dispatch, $x$)
   \STATE $x_e \leftarrow$ [experts[$e$]($x_{in}[:,e]$) for $e$ in range($E$)]
   \STATE $x_e \leftarrow$ stack($x_e$, dim=$1$)
   \STATE $x_{out} \leftarrow$ einsum($'becs,bec,becd\rightarrow bsd'$, dispatch, gating, $x_e$)
    \STATE {\bfseries Return:} $x_{out}$
\end{algorithmic}
\end{algorithm}

\begin{algorithm}[t]
   \caption{TC-DiT layer}
   \label{algo:tcdit}
\begin{algorithmic}
   \STATE {\bfseries Input:} $x$ (input tensor), $W_r$ (routing weights), $experts$ (list of expert FFNs)

   \STATE {\bfseries Variables:} $B$ (batch size), $S$ (sequence length), $d$ (hidden dim) ,$K$ (experts per token)
   
   \STATE /* Step 1: Save original input shape */
   \STATE orig\_shape $\leftarrow \texttt{shape}(x)$
   
   \STATE /* Step 2: Compute Token-Expert Affinity Matrix */
   \STATE logits $\leftarrow$ einsum($'bsd,de\rightarrow bse'$, $x$, $W_r$)
   \STATE scores $\leftarrow \texttt{softmax}(logits, \text{dim}=-1)$
   
   \STATE /* Step 3: Select top-k tokens for each expert */
   \STATE gating, index $\leftarrow$ top\_k(scores, k=$K$, dim=$-1$)
   
   \STATE /* Step 4: Flatten $x$ and top-k indices */
   \STATE $x \leftarrow \texttt{view}(x, (-1, x.shape[-1]))$
   \STATE flat\_topk\_idx $\leftarrow \texttt{view}(topk\_idx, (-1))$
   
   \STATE /* Step 4: Process tokens through experts */
   \STATE $x \leftarrow \texttt{repeat\_interleave}(x, K, \text{dim}=0)$
   \STATE $y \leftarrow \texttt{empty\_like}(x)$
   \FOR{$i \gets 1$ {\bfseries to} $len(experts)$}
       \STATE $y[\text{flat\_topk\_idx} == i] \leftarrow \texttt{expert}_i(x[\text{flat\_topk\_idx} == i])$
   \ENDFOR
   \STATE $y \leftarrow \texttt{sum}(\texttt{view}(y,(*\text{gating}.shape, -1))\cdot \text{gating}.unsqueeze(-1), \text{dim}=1)$
   \STATE $y \leftarrow \texttt{view}(y, \text{orig\_shape})$
   \STATE {\bfseries Return:} $y$
\end{algorithmic}
\end{algorithm}

\begin{algorithm}[t]
   \caption{DiffMoE layer (Training)}
   \label{algo:diff_moe_train}
\begin{algorithmic}
   \STATE {\bfseries Input:} $x$ (input tensor)
   \STATE {\bfseries Variables:} $B$ (batch size), $S$ (flattened sequence length), $D$ (hidden dim), $N$ (number of experts) 
   \STATE \hspace{3.8em} $W_r$ (routing weights), $experts$ (list of expert FFNs), $C$ (expert capacity)
   
   \STATE /* Step 1: Batch-level token pool and compute capacity prediction */
   \STATE $x \leftarrow \texttt{view}(x, (-1, D))$
   \STATE $S \leftarrow \texttt{shape}(x)[0]$
   \STATE $capacity\_pred \leftarrow \texttt{capacity\_predictor}(\texttt{detach}(x))$
   \STATE $C_{\rm train} \leftarrow \texttt{int}((S / N) \times C)$
   
   \STATE /* Step 2: Compute token-expert affinity scores */
   \STATE logits $\leftarrow$ einsum($'sd,de\rightarrow se'$, $x$, $W_r$)
   \STATE scores $\leftarrow \texttt{softmax}(logits, \text{dim}=-1).\texttt{permute}(-1, -2)$
   \STATE gating, index $\leftarrow \texttt{top\_k}(scores, k=C_{\rm train}, \text{dim}=-1, \text{sorted}=\texttt{False})$
   
   \STATE /* Step 3: Process tokens through experts */
   \STATE $y \leftarrow \texttt{zeros\_like}(x)$
   \STATE ones $\leftarrow \texttt{zeros}(N, S)$
   \FOR{$i \gets 1$ {\bfseries to} N}
       \STATE $y[\text{index}[i], :] \leftarrow y[\text{index}[i], :] + \text{gating}[i].\texttt{unsqueeze}(-1) \times \texttt{expert}_i(x[\text{index}[i], :])$
       \STATE $\text{ones}[i][\text{index}[i]] \leftarrow 1.$
   \ENDFOR
   
   \STATE /* Step 4: Update capacity threshold */
   \STATE $\texttt{update\_threshold}(capacity\_pred)$
   
   \STATE /* Step 5: Reshape output */
   \STATE $x_{out} \leftarrow \texttt{view}(y, (B, s, D))$
   
   \STATE {\bfseries Return:} $x_{out}, ones, capacity\_pred$
\end{algorithmic}
\end{algorithm}

\begin{algorithm}[t]
   \caption{DiffMoE layer (Inference)}
   \label{algo:diff_moe_infer}
\begin{algorithmic}
   \STATE {\bfseries Input:} $x$ (input tensor)
   \STATE {\bfseries Variables:} $B$ (batch size), $S$ (flattened sequence length), $D$ (hidden dim), $N$ (number of experts) 
   \STATE \hspace{3.8em} $W_r$ (routing weights), $experts$ (list of expert FFNs), $C$ (expert capacity)
   \STATE \hspace{3.8em} $threshold$ (expert threshold)
   
   \STATE /* Step 1: Reshape input and compute capacity prediction */
   \STATE $x \leftarrow \texttt{view}(x, (-1, D))$
   \STATE $S \leftarrow \texttt{shape}(x)[0]$
   \STATE $capacity\_pred \leftarrow \texttt{sigmoid}(\texttt{capacity\_predictor}(\texttt{detach}(x)))$
   
   \STATE /* Step 2: Compute token-expert affinity scores */

   \STATE logits $\leftarrow$ einsum($'sd,de\rightarrow se'$, $x$, $W_r$)
   \STATE scores $\leftarrow \texttt{softmax}(logits, \text{dim}=-1).\texttt{permute}(-1, -2)$
   
   \STATE /* Step 3: Process tokens through experts */
   \STATE $y \leftarrow \texttt{zeros\_like}(x)$
   
   \FOR{$i \gets 1$ {\bfseries to} N}
       \STATE $k_{pred} \leftarrow \texttt{sum}(\texttt{where}(capacity\_pred[:,i] > threshold[i], 1, 0))$
       \STATE $\text{gating}, \text{index} \leftarrow \texttt{top\_k}(scores[i], k=k_{pred}, \text{dim}=-1, \text{sorted}=\texttt{False})$
       \STATE $y[\text{index}, :] \leftarrow y[\text{index}, :] + \text{gating}.\texttt{unsqueeze}(-1) \times \texttt{expert}_i(x[\text{index}, :])$
   \ENDFOR
   
   \STATE /* Step 4: Reshape output */
   \STATE $x_{out} \leftarrow \texttt{view}(y, (B, s, D))$
   
   \STATE {\bfseries Return:} $x_{out}$
\end{algorithmic}
\end{algorithm}

\end{document}